\documentclass{article}

\usepackage{microtype}
\usepackage{graphicx}
\usepackage{subcaption}
\usepackage{booktabs} 

\usepackage{hyperref}

\usepackage[preprint]{icml2026}

\usepackage{amsmath}
\usepackage{amssymb}
\usepackage{mathtools}
\usepackage{amsthm}
\usepackage{multirow}

\usepackage[capitalize,noabbrev]{cleveref}

\theoremstyle{plain}

\theoremstyle{definition}

\theoremstyle{remark}

\usepackage[textsize=tiny]{todonotes}

\icmltitlerunning{Unveiling Scaling Behaviors in Molecular Language Models: Effects of Model Size, Data, and Representation}

\begin{document}

\twocolumn[
  \icmltitle{Unveiling Scaling Behaviors in Molecular Language Models: Effects of Model Size, Data, and Representation}
    \icmlsetsymbol{equal}{*}

  \begin{icmlauthorlist}
    \icmlauthor{Dong Xu}{equal,szu}
    \icmlauthor{Qihua Pan}{equal,szu}
    \icmlauthor{Sisi Yuan}{szu}
    \icmlauthor{Jianqiang Li}{szu}
    \icmlauthor{Zexuan Zhu}{szu}
    \icmlauthor{Junkai Ji}{szu}
  \end{icmlauthorlist}

  \icmlaffiliation{szu}{School of Artificial Intelligence, Shenzhen University, Shenzhen 518060, China}

  \icmlcorrespondingauthor{Junkai Ji}{jijunkai@szu.edu.cn}

  \icmlkeywords{Molecular language models, scaling laws, pretraining, representation learning}

  \icmlkeywords{Machine Learning, ICML}

  \vskip 0.3in
]

·\printAffiliationsAndNotice{}  

\begin{abstract}
Molecular generative models, often employing GPT-style language modeling on molecular string representations, have shown promising capabilities when scaled to large datasets and model sizes. However, it remains unclear and subject to debate whether these models adhere to predictable scaling laws under fixed computational budgets, which is a crucial understanding for optimally allocating resources between model size, data volume, and molecular representation. In this study, we systematically investigate the scaling behavior of molecular language models across both pretraining and downstream tasks. We train 300 models and conduct over 10,000 experiments, rigorously controlling compute budgets while independently varying model size, number of training tokens, and molecular representation. Our results demonstrate clear scaling laws in molecular models for both pretraining and downstream transfer, reveal the substantial impact of molecular representation on performance, and explain previously observed inconsistencies in scaling behavior for molecular generation. Additionally, we publicly release the largest library of molecular language models to date to facilitate future research and development. Code and models are available at \url{https://github.com/SZU-ADDG/MLM-Scaling}.
\end{abstract}

\section{Introduction}
Traditional drug discovery pipelines rely on large-scale virtual screening and iterative validation, which remain computationally and temporally expensive. By casting molecular design as a problem of distribution modeling, generative approaches offer a more efficient alternative to exhaustive screening. A common paradigm encodes molecules as linearized sequences, such as SMILES\cite{1988-SMILES,2024-SAFE,2018-DeepSMILES}, and pretrains sequence models via next-token prediction for downstream transfer. GPT-style autoregressive models have been applied to molecular strings and shown to generate valid and diverse molecules when trained on large-scale SMILES corpora~\cite{2023-GPT4, 2021-MolGPT}.

\begin{figure}[t]
  \vskip 0.2in
  \centering
  \includegraphics[width=0.94\columnwidth]{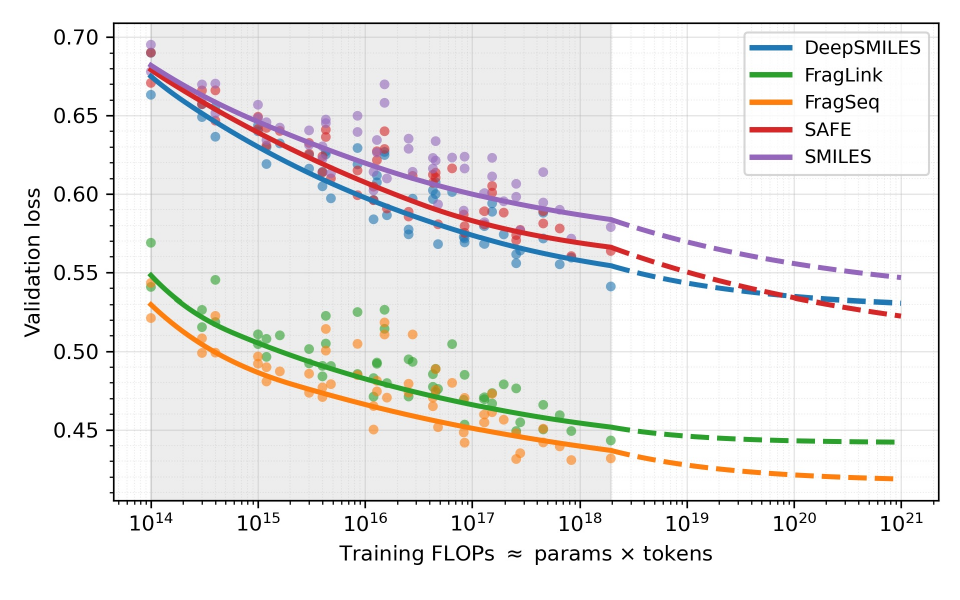}
  \vspace{-4mm}
  \caption{
  Pretraining loss scaling under compute-controlled analysis.
  Points are end-of-run validation losses from single-epoch from-scratch runs.
  The shaded region marks the compute range covered by our training grid, and dashed segments show extrapolation beyond this range.
  }
  \label{fig:frontier_loss_vs_compute}
  \vspace{-6mm}
\end{figure}

As molecular corpora continue to expand and training runs scale up, questions of scaling become unavoidable. A central issue is whether molecular language models obey predictable scaling laws under a fixed computational budget. If such laws hold, the trade-off between model capacity and data can be systematically quantified, providing a principled basis for training design. Conversely, if scaling behavior breaks down, further increasing model size may yield diminishing returns. This uncertainty directly shapes where effort should be invested, toward larger models, more data, improved representations, or more informative evaluation.

In natural language modeling, scaling analysis follows a well-established methodology based on pretraining loss and compute-controlled comparisons. Power-law trends in cross-entropy loss enable reliable extrapolation across scales~\cite{2020-ScalingLaws}, while compute-optimal training has been shown to depend critically on the parameter–token balance, with large models often undertrained under limited data~\cite{2022-Chinchilla}. Molecular modeling introduces additional complexity because molecular strings are engineered representations rather than naturally occurring text. Different representations alter sequence length and token statistics, potentially reshaping scaling behavior and the compute-optimal frontier. 

\cite{2023-09-NMI-Scaling4chemical} reported evidence of scaling behavior in molecular language models by varying model and dataset sizes. However, the experimental analysis remains incomplete: it is largely confined to the pretraining stage; does not explicitly control for compute budget that is a critical factor in scaling studies; and does not examine how different molecular representations may influence scaling behavior. In contrast, \cite{2025-08-ICMLWS-NovoMolGen} argued that chemical models do not exhibit consistent scaling behavior on \textit{de novo} generation tasks, and reported weak correlations between commonly used pretraining metrics and molecular generation performance. Nevertheless, this conclusion warrants further scrutiny, as the metrics currently used to assess performance may be insufficient to fully capture generative quality and task-relevant capabilities, which we discuss in detail in the following sections. Taken together, existing studies have yet to provide a reliable and systematic characterization of scaling behavior in molecular models across both pretraining and downstream transfer settings. To address these gaps, we present a systematic scaling study, where we train a total of 300 molecular language models and conduct over 10,000 experiments covering both pretraining and downstream transfer. Under rigorously controlled compute budgets, we study the individual scaling trends by independently varying model size, number of training tokens, and molecular representation.

To sum up, the contributions of this study are fourfold.
\vspace{-2mm}
\begin{itemize}
  \item We rigorously demonstrate that molecular language models exhibit scaling behaviors in both pretraining and downstream tasks.
  \vspace{-2mm}
  \item We reveal that molecular representations have a significant impact on model performance across different tasks.
  \vspace{-2mm}
  \item We provide an explanation for the previously observed lack of scaling in molecular generation tasks.
  \vspace{-2mm}
  \item We train and publicly release the largest library of molecular language models to date, spanning a range of model sizes, numbers of training tokens, and molecular representations.
\end{itemize}

\section{Related Work}

\subsection{Scaling in biological sequence models}
Scaling effects have been extensively studied in protein and nucleotide sequence models. Protein language models have been shown to capture increasingly rich structural and functional information as training scale increases \cite{2021-04-PANS-ESM}. Similarly, genome-scale modeling demonstrates performance gains when model capacity and context length are appropriately matched to the task \cite{2024-02-Science-Evo}. Multi-scale evaluations on genomic benchmarks further highlight scale-dependent variations across different tasks \cite{2024-10-NM-NucleotideTransformer}. However, some studies report unstable or non-monotonic scaling trends, attributing these to data redundancy and compositional shifts \cite{2025-07-DataSaturationScaling}.

\subsection{Scaling studies for molecular language models}
In chemistry, scaling studies have addressed transfer learning after pretraining, trend fitting across scales, and data selection for large-scale training. Early work on SMILES pretraining demonstrated improved transfer performance with larger pretraining corpora~\cite{2020-10-ChemBERTa}. Subsequent scaling analyses in deep chemical modeling explored how scaling curves evolve with changes in model size, dataset scale, and task-specific error floors~\cite{2023-09-NMI-Scaling4chemical}. Studies on data selection emphasize that diversity and information content can significantly influence scaling trends in chemistry~\cite{2025-12-CommunChem-ChemFM}. However, more skeptical findings have been reported for de novo generation metrics and goal-directed optimization. For example, \cite{2025-08-ICMLWS-NovoMolGen} found weak correlations between common pretraining metrics and generation outcomes, with inconsistent benefits from scaling across various settings. Similarly, \cite{2025-11-DiversityBeatsScaling} observed diminishing returns in hit-oriented optimization tasks, underscoring the importance of data diversity. Overall, conclusions regarding scaling behavior in chemistry strongly depend on the specific task and evaluation metric, with different observables suggesting varying scaling dynamics.

\begin{figure*}[ht]
  \begin{center}
  \centerline{\includegraphics[width=\linewidth]{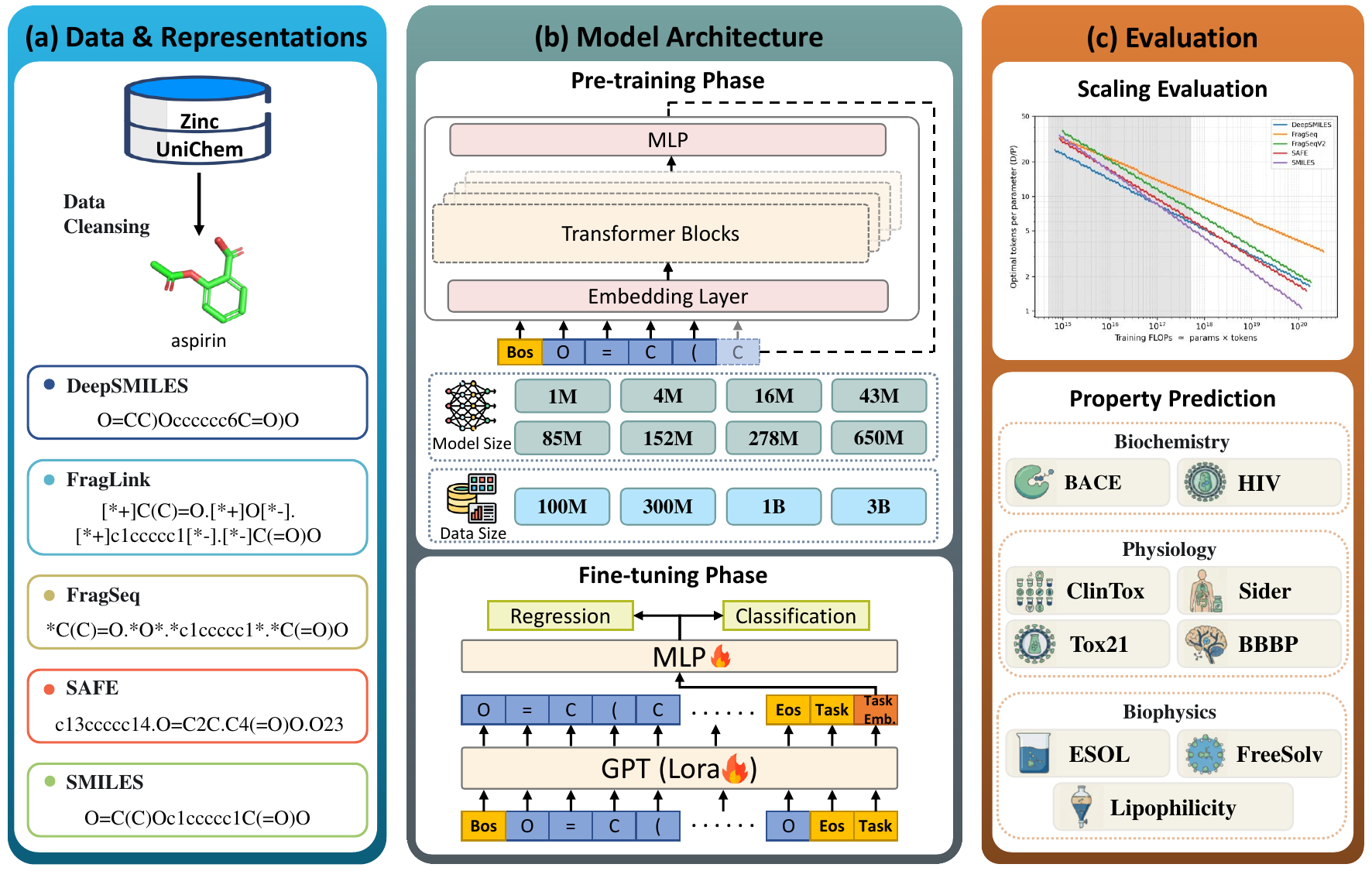}}
    \caption{An overview of our research framework. (a) Data \& Representations: Beginning with raw molecular data from the ZINC and UniChem databases, each molecule is converted into five distinct string-based representations: DeepSMILES, FragLink, FragSeq, SAFE and SMILES. (b) Model Architecture: A GPT-based model is used for all experiments. The pre-training phase utilizes an autoregressive prediction objective to train models of varying sizes (from 1M to 650M parameters) on different data scales (from 100M to 3B tokens).  The fine-tuning phase adapts the pre-trained model using LoRA for specific downstream regression or classification tasks. (c) Evaluation: The model's capabilities are assessed across a wide range of tasks, including the predicted minimal validation loss along the compute-optimal frontier and a comprehensive suite of property prediction benchmarks spanning biochemistry, physiology and biophysics.}
    \vspace{-8mm}
    \label{figure: overview}
  \end{center}
\end{figure*}

\section{Problem Setup and Experimental Design}
\label{sec:exp_design}
\subsection{Factors and Study Design}
\label{sec:exp_design_factors}

This work studies the scaling behavior of autoregressive language models on molecular representations.
A molecule is tokenized into a sequence $x_{1:T}$ using a tokenizer $\tau_r(\cdot)$ for a given representation $r$ from the set of $\mathcal{R}=\{\texttt{SMILES},\texttt{SAFE},\texttt{DeepSMILES},\texttt{FragSeq},\texttt{FragLink}\}$. A decoder-only Transformer with parameters $\theta$ is pretrained to model the sequence distribution $p_\theta(x_{1:T})=\prod_{t=1}^{T} p_\theta(x_t \mid x_{<t})$ using a token-averaged cross-entropy loss.
Scaling is analyzed at two levels.
The first level, pretraining loss scaling, quantifies how validation loss varies with model size and training tokens under a fixed compute budget. 
The second level, downstream transfer scaling, examines how performance changes as a function of the pretraining scale after lightweight adaptation.

\textbf{Notation.} $P$ denotes the number of trainable parameters in a pretrained model, and $D$ denotes the number of training tokens consumed during pretraining after tokenization under $r$, counting only tokens that contribute to the loss.
Training compute is denoted by $C$ and is approximated as proportional to $P\cdot D$ for dense Transformer pretraining, with the constant absorbed into the definition of $C$.
The final validation loss is denoted by $L(P,D)$ and is the token-averaged cross-entropy on a held-out validation set under the same representation.
Section~\ref{sec:compute_opt_scaling} models $L(P,D)$ with a bivariate power law of the form $L(P,D)=L_{\infty}+k_P P^{-\alpha}+k_D D^{-\beta}$, where $L_{\infty}$ is an empirical lower bound and $(k_P,k_D,\alpha,\beta)$ are fitted constants.
Unless stated otherwise, the dependence on $r$ is implicit to reduce notation.

\subsection{Pretraining Grids and Compute-Controlled Sweeps}
\label{sec:exp_design_pretrain}

We specify a structured pretraining grid over representation, model size, dataset token budget, and training duration.
Dataset token budget refers to the number of tokens in the training corpus after tokenization under $r$.
The effective training tokens $D$ increase with training duration because the same dataset can be re-read for multiple epochs.
The main grid trains one epoch over a full $P \times$ dataset-token-budget plane for each representation.
We use five molecular representations $r \in \mathcal{R}$.
We use eight model sizes
$P \in \{1\text{M}, 4\text{M}, 16\text{M}, 43\text{M}, 85\text{M}, 152\text{M}, 278\text{M}, 650\text{M}\}$.
For each representation and each model size, we train on four dataset token budgets
$\{100\text{M}, 300\text{M}, 1\text{B}, 3\text{B}\}$ for one epoch.
This grid provides the primary observations of $L(P,D)$ used for scaling fits and compute-optimal analysis.

To isolate the effect of additional training compute at fixed dataset size, we run duration-controlled sweeps.
For each representation, we select mid-scale models
$P \in \{4\text{M}, 16\text{M}, 43\text{M}, 85\text{M}, 152\text{M}\}$.
To study the impact of longer training, we evaluate each setting under two training duration conditions: a single-epoch run and a two-epoch run. Crucially, the two-epoch runs are trained from scratch to prevent warm-start effects from confounding the comparison. Under this setup, the dataset token budget is fixed and the effective number of training tokens, $D$, increases with the number of epochs.
We extend training for the smallest model to cover a larger range of $D$.
For each representation, the $P{=}1\text{M}$ model is trained on each dataset token budget for up to five epochs.
This sweep is used to probe early trends under extended training for small models.
An extreme-duration sweep is run to support later analysis of de novo metrics.
On \texttt{SMILES}, the $P{=}1\text{M}$ model is trained on the 100M-token dataset for up to ten epochs.
This setting is used as a diagnostic regime for saturation and sensitivity analysis of common de novo metrics \cite{2021-MolGPT}.

\subsection{Evaluation and Downstream Transfer}
\label{sec:exp_design_eval}

This section defines the evaluation outputs used in later analysis.
It covers pretraining validation loss, de novo generation evaluation, and downstream transfer evaluation.

For each pretraining run, validation loss is tracked under the same representation as training.
The loss $L(P,D)$ is the token-averaged cross-entropy on a held-out validation set.
It is evaluated at fixed points during training and at the end of training.
Checkpoints are saved to support controlled comparisons across training progress.
For the downstream transfer subset described below, five checkpoints are saved per epoch and taken at fixed intervals within the epoch.
This creates multiple observations at different effective token counts $D$ under the same $(P,r)$ and dataset token budget.

De novo evaluation samples molecules from a pretrained model without task-specific supervision.
Samples are generated with a fixed decoding setting unless stated otherwise.
We report standard de novo metrics, including validity, uniqueness, novelty, and diversity.
These metrics are used as diagnostic observables in later sections.
Details of sampling settings and metric definitions are provided in the appendix.

Downstream transfer is evaluated on nine molecular property prediction tasks.
The classification tasks are BACE, HIV, BBBP, SIDER, Tox21, and ClinTox.
The regression tasks are ESOL, FreeSolv, and Lipophilicity.
For classification, ROC-AUC is reported.
For regression, RMSE is reported.
Dataset splits and preprocessing follow the benchmark defaults.
Implementation details are provided in the Appendix \ref{sec: appendix C}.
Transfer uses lightweight adaptation with LoRA \cite{hu2022lora} on top of a pretrained checkpoint.
The base model is kept fixed and only LoRA parameters and the task head are trained.
A controlled subset of pretrained checkpoints is selected to create a balanced transfer grid.
We use four model sizes $P \in \{4\text{M}, 16\text{M}, 43\text{M}, 152\text{M}\}$ and four dataset token budgets $\{100\text{M}, 300\text{M}, 1\text{B}, 3\text{B}\}$.
We cover all five representations $r \in \mathcal{R}$.
For each $(r,P,\text{budget})$ setting, five single-epoch checkpoints are included.
This yields $5 \times 4 \times 4 \times 5 = 400$ checkpoints.
Each checkpoint is adapted to nine tasks.
This produces $400 \times 9 = 3600$ transfer training runs.
This design enables matched comparisons across $r$, $P$, dataset token budget, and training progress.

\section{Compute-Optimal Scaling Laws}
\label{sec:compute_opt_scaling}

\subsection{A bivariate scaling law for MLM}
\label{sec:bivariate_scaling}

Let $L(P,D)$ denote the validation loss measured at the end of a training run that consumes $D$ tokens.
Each $(P,D)$ point is obtained from an independent from-scratch run to avoid confounding changes in training schedules when the run length is extended.
A bivariate power-law form is used to model how loss varies with model scale and data scale:
\begin{equation}
\label{eq:bivariate_scaling}
L(P,D) = L_{\infty} + k_P P^{-\alpha} + k_D D^{-\beta}.
\end{equation}
Here $L_{\infty}$ is an empirical loss floor that captures a combination of factors such as model class mismatch,
finite context length, representation constraints, optimization imperfections, and data noise.
The terms $k_P P^{-\alpha}$ and $k_D D^{-\beta}$ describe the marginal loss reductions associated with increasing model capacity and training data, respectively.
The parameters $(L_{\infty},k_P,k_D,\alpha,\beta)$ are fit separately for each molecular representation.

\subsection{Compute constraint and optimal frontier}
\label{sec:compute_opt_frontier}

Training is constrained by a finite compute budget $C$ measured in FLOPs.
For a dense Transformer, training compute is approximately proportional to the product of parameter count and training tokens:
\begin{equation}
\label{eq:compute_constraint}
C = \kappa\, P D,
\end{equation}
where $\kappa$ is a constant that depends on architectural details.
For derivations, $\kappa$ can be absorbed into the definition of $C$, yielding the iso-FLOP constraint $PD=C$.

Under a fixed compute budget $C$, the compute-optimal allocation is obtained by minimizing $L(P,D)$ subject to $PD=C$.
Substituting $D=C/P$ into Eq.~\eqref{eq:bivariate_scaling} gives a one-variable form:
\begin{equation}
\label{eq:loss_under_constraint}
\begin{aligned}
L(P;C)
&= L_{\infty} + k_P P^{-\alpha} + k_D \left(\frac{C}{P}\right)^{-\beta} \\
&= L_{\infty} + k_P P^{-\alpha} + k_D C^{-\beta} P^{\beta}.
\end{aligned}
\end{equation}
Assume $\alpha,\beta>0$ and $k_P,k_D>0$, and optimize over $P>0$.
Setting $\frac{d}{dP}L(P;C)=0$ yields a unique compute-optimal model size:
\begin{equation}
\label{eq:P_opt}
P_{\mathrm{opt}}(C) =
\left(\frac{\alpha k_P}{\beta k_D}\right)^{\frac{1}{\alpha+\beta}}
C^{\frac{\beta}{\alpha+\beta}}.
\end{equation}
The corresponding compute-optimal number of training tokens is
\begin{equation}
\label{eq:D_opt}
D_{\mathrm{opt}}(C) =
\frac{C}{P_{\mathrm{opt}}(C)} =
\left(\frac{\beta k_D}{\alpha k_P}\right)^{\frac{1}{\alpha+\beta}}
C^{\frac{\alpha}{\alpha+\beta}}.
\end{equation}
An equivalent summary is the compute-optimal tokens-per-parameter ratio,
\begin{equation}
\label{eq:rho_opt}
\rho_{\mathrm{opt}}(C) \equiv \frac{D_{\mathrm{opt}}(C)}{P_{\mathrm{opt}}(C)}
=
\left(\frac{\beta k_D}{\alpha k_P}\right)^{\frac{2}{\alpha+\beta}}
C^{\frac{\alpha-\beta}{\alpha+\beta}}.
\end{equation}
These closed-form expressions define a compute-optimal frontier in $(P,D)$ space.
The fitted exponents and their implications for different representations are analyzed empirically in Sec.~\ref{sec:results_analysis}.

\subsection{A power-law summary of $\rho_{\mathrm{opt}}(C)$}
\label{sec:rho_1d_approx}

Within a finite compute range, $\rho_{\mathrm{opt}}(C)$ can be summarized by a representation-specific one-dimensional power law:
\begin{equation}
\label{eq:rho_1d}
\rho_{\mathrm{opt}}(C) \approx a_{\mathrm{repr}}\, C^{s_{\mathrm{repr}}},
\end{equation}
where $a_{\mathrm{repr}}>0$ and $s_{\mathrm{repr}}$ are scalars associated with a given representation $\mathrm{repr}$.
Eq.~\eqref{eq:rho_1d} is fit via a log-linear regression in base 10.
Let $(C_k,\rho_k)$ be the compute-optimal points obtained from Eq.~\eqref{eq:rho_opt} across compute levels.
Eq.~\eqref{eq:rho_opt} implies a power-law dependence of $\rho_{\mathrm{opt}}$ on $C$.
We report $(b_{\mathrm{repr}}, s_{\mathrm{repr}})$ via a log-linear fit over a common set of compute levels to summarize this dependence for each representation.
Define
\begin{equation}
x_k = \log_{10} C_k,\qquad y_k = \log_{10}\rho_k,
\end{equation}
and fit
\begin{equation}
\label{eq:loglin_fit}
y_k \approx b_{\mathrm{repr}} + s_{\mathrm{repr}} x_k.
\end{equation}
The least-squares solution is
\begin{equation}
\label{eq:ls_slope}
s_{\mathrm{repr}} =
\frac{\sum_{k=1}^{K}(x_k-\bar{x})(y_k-\bar{y})}{\sum_{k=1}^{K}(x_k-\bar{x})^2},
\qquad
b_{\mathrm{repr}} = \bar{y} - s_{\mathrm{repr}}\bar{x},
\end{equation}
where $\bar{x}=\frac{1}{K}\sum_{k=1}^{K}x_k$ and $\bar{y}=\frac{1}{K}\sum_{k=1}^{K}y_k$.
This yields
\begin{equation}
\label{eq:a_from_b}
a_{\mathrm{repr}} = 10^{b_{\mathrm{repr}}},\qquad
\rho_{\mathrm{opt}}(C) \approx 10^{b_{\mathrm{repr}}}\, C^{s_{\mathrm{repr}}}.
\end{equation}
The factor $10^{s_{\mathrm{repr}}}$ has a direct interpretation: when compute increases by one order of magnitude, the ratio $\rho_{\mathrm{opt}}$ is multiplied by $10^{s_{\mathrm{repr}}}$.
Estimated $(s_{\mathrm{repr}}, b_{\mathrm{repr}})$ values are reported together with empirical results.
The specific derivation process is detailed in Appendix \ref{sec: appendix A}.

\section{Results and Analysis}
\label{sec:results_analysis}
\subsection{Loss Scaling}
\label{sec:loss_scaling}

For each molecular representation $r$, models are trained on a fixed grid of parameter counts $P$ and consumed training tokens $D$ (single-epoch from-scratch runs).
Each $(P,D)$ point is obtained from an independent from-scratch run that consumes $D$ training tokens.
All representations share the same vocabulary, so the cross-entropy losses are directly comparable across representations.
Figure~\ref{fig:frontier_loss_vs_compute} plots end-of-run loss against compute $C\propto P D$ and overlays the predicted compute-optimal frontier from Sec.~\ref{sec:compute_opt_scaling}.
Across the covered compute range, the frontier predicts a consistent reduction in validation loss as compute increases.
The offset between representations indicates that representation choice shifts the compute-optimal frontier under matched compute.

Across the covered compute range, validation loss decreases as compute increases along the predicted frontier.
Within this range, the frontier follows the lower envelope of observed runs, indicating that the fitted bivariate law is adequate for compute-controlled analysis.
The frontier is consistently shifted across representations under matched compute, so representation choice changes the compute-optimal loss level.
The extrapolated segments suggest continued but diminishing loss reductions at larger compute, which motivates forecasting scaling trends beyond the current grid.

\subsection{IsoFLOP and IsoLoss}
\label{sec:isoflop_isoloss}

Compute-controlled scaling is visualized from two complementary views.
All curves and contours are generated from the fitted bivariate law in Sec.~\ref{sec:compute_opt_scaling},
where compute is approximated as $C \propto PD$ and the proportionality constant is absorbed into $C$.
Table~\ref{tab:fit_summary} summarizes the fitted scaling exponents and in-grid errors, indicating stable fits across representations on the single-epoch grid.

\begin{figure}[!ht]
  \centering
  \begin{subfigure}[t]{0.49\columnwidth}\centering
    \includegraphics[width=\linewidth]{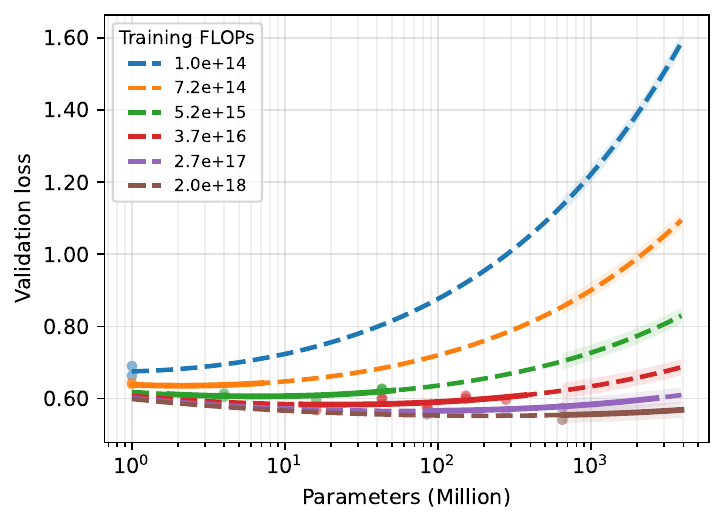}
    \vspace{-6mm}
    \caption{DeepSMILES}
  \end{subfigure}\hfill
  \begin{subfigure}[t]{0.49\columnwidth}\centering
    \includegraphics[width=\linewidth]{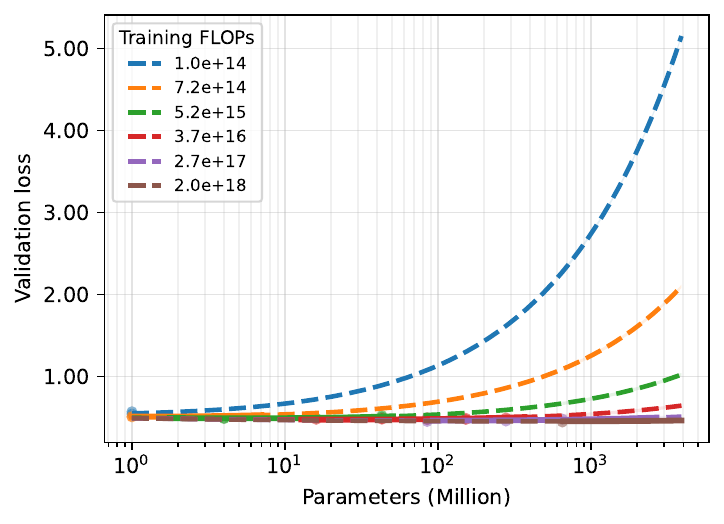}
    \vspace{-6mm}
    \caption{FragLink}
  \end{subfigure}
  \begin{subfigure}[t]{0.49\columnwidth}\centering
    \includegraphics[width=\linewidth]{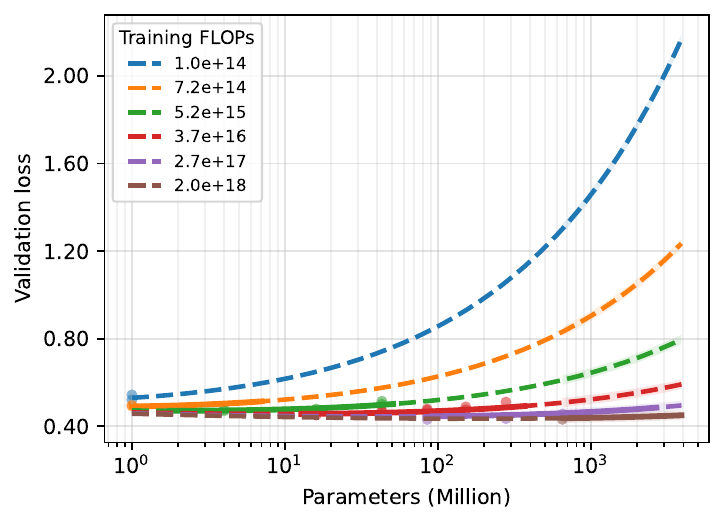}
    \vspace{-6mm}
    \caption{FragSeq}
  \end{subfigure}\hfill
  \begin{subfigure}[t]{0.49\columnwidth}\centering
    \includegraphics[width=\linewidth]{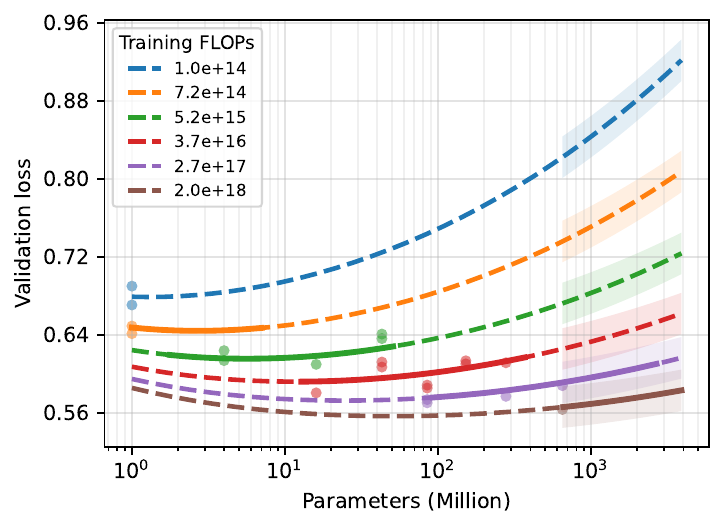}
    \vspace{-6mm}
    \caption{SAFE}
  \end{subfigure}
  \begin{subfigure}[t]{0.49\columnwidth}\centering
    \includegraphics[width=\linewidth]{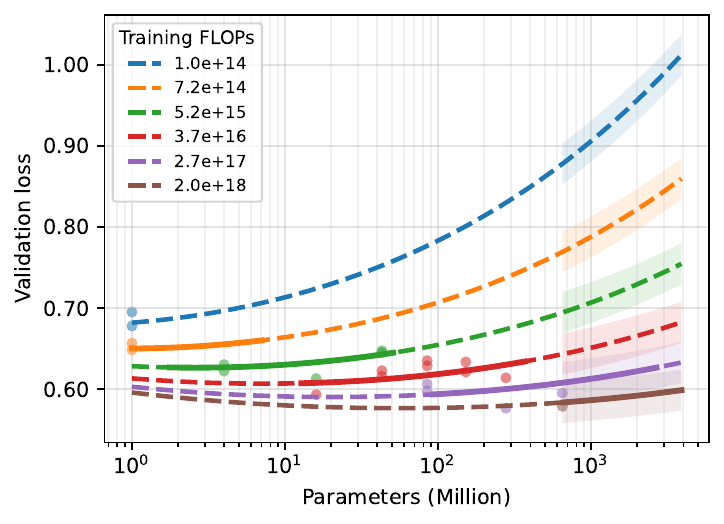}
    \vspace{-6mm}
    \caption{SMILES}
  \end{subfigure}\hfill
  \begin{subfigure}[t]{0.49\columnwidth}\centering
    \includegraphics[width=\linewidth]{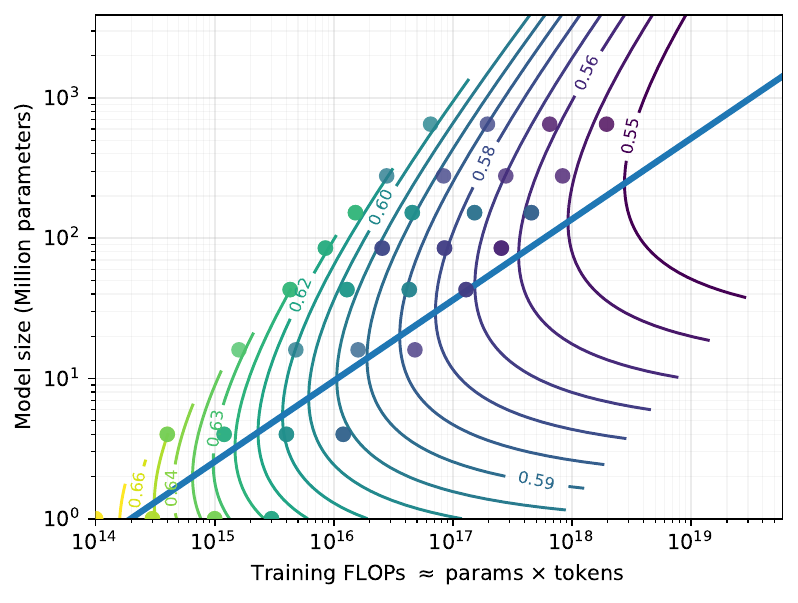}
    \vspace{-6mm}
    \caption{DeepSMILES}
  \end{subfigure}
  \begin{subfigure}[t]{0.49\columnwidth}\centering
    \includegraphics[width=\linewidth]{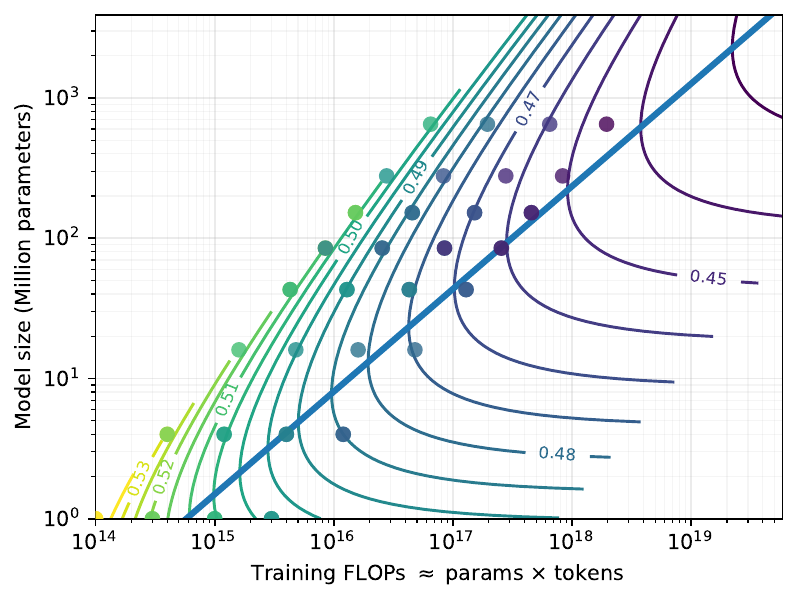}
    \vspace{-6mm}
    \caption{FragLink}
  \end{subfigure}\hfill
  \begin{subfigure}[t]{0.49\columnwidth}\centering
    \includegraphics[width=\linewidth]{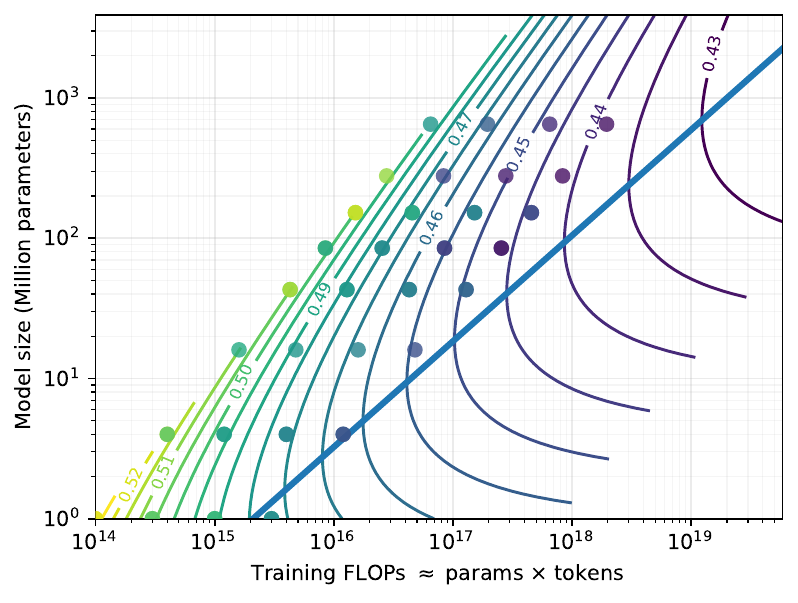}
    \vspace{-6mm}
    \caption{FragSeq}
  \end{subfigure}
  \begin{subfigure}[t]{0.49\columnwidth}\centering
    \includegraphics[width=\linewidth]{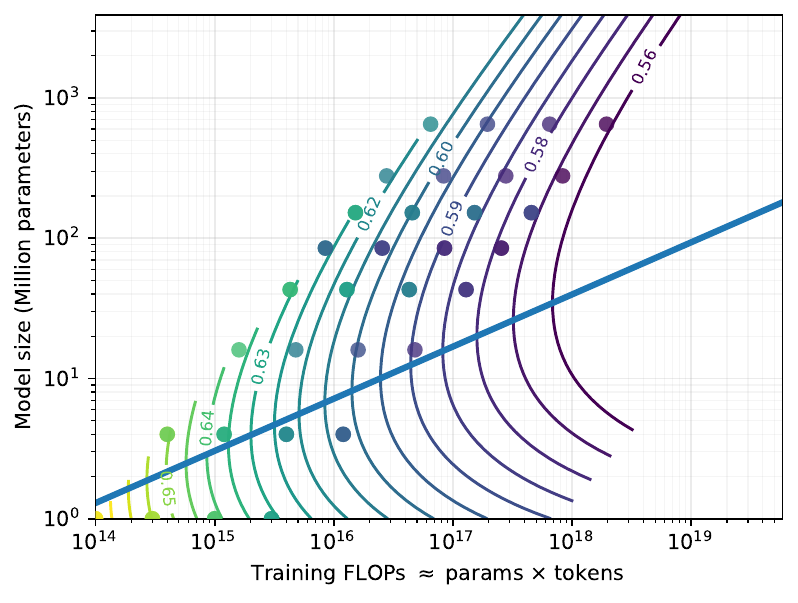}
    \vspace{-6mm}
    \caption{SAFE}
  \end{subfigure}\hfill
  \begin{subfigure}[t]{0.49\columnwidth}\centering
    \includegraphics[width=\linewidth]{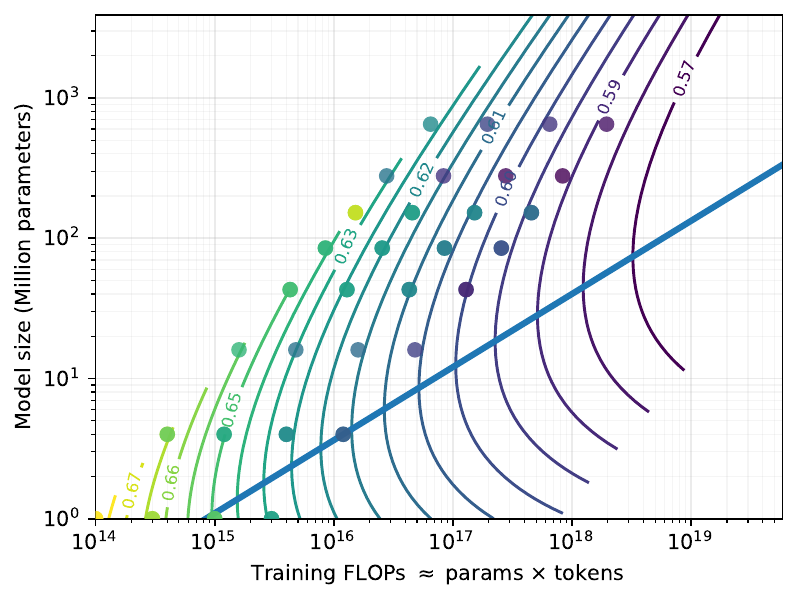}
    \vspace{-6mm}
    \caption{SMILES}
  \end{subfigure}
  \caption{
  Compute-controlled views from the fitted bivariate law.
  (a--e) IsoFLOP curves under fixed compute budgets, with observed single-epoch runs overlaid.
  (f--j) IsoLoss contours in the $(C,P)$ plane, with the compute-optimal frontier highlighted.
  }
  \label{fig:isoflop_isoloss}
  \vspace{-8mm}
\end{figure}

\textbf{IsoFLOP curves.}
Panels (a--e) of Fig.~\ref{fig:isoflop_isoloss} report isoFLOP curves.
For each fixed $C$, the constraint $PD=C$ implies $D=C/P$.
Substituting into Eq.~\eqref{eq:bivariate_scaling} yields the predicted loss $L(P;C)$.
Observed single-epoch end-of-run losses are overlaid as colored markers.
In each panel, solid segments indicate the range where $D$ falls within the token range covered by the single-epoch grid for that representation.
Dashed segments indicate extrapolation beyond that covered range.
The translucent bands on the right visualize an empirical uncertainty scale estimated from grid residuals.

\textbf{IsoLoss curves.}
Panels (f--j) of Fig.~\ref{fig:isoflop_isoloss} report isoLoss contours in the $(C,P)$ plane.
Each contour corresponds to a fixed target loss level $\tilde{L}$ under the fitted law, and therefore represents a trade-off between model size and consumed tokens.
For a target $\tilde{L} > L_{\infty}$, Eq.~\eqref{eq:bivariate_scaling} defines a level set in $(P,D)$ space.
Solving for $D$ yields
\begin{equation}
\label{eq:isoloss_curve_PD}
D(P;\tilde{L}) =
\left(
\frac{k_D}{\tilde{L}-L_{\infty}-k_P P^{-\alpha}}
\right)^{\frac{1}{\beta}},
\end{equation}
when the denominator is positive.
To plot isoLoss curves against compute, the level set is mapped to $(C,P)$ space by
\begin{equation}
\label{eq:isoloss_curve_CP}
C(P;\tilde{L}) = P \cdot D(P;\tilde{L}),
\end{equation}
with $D(P;\tilde{L})$ defined in Eq.~\eqref{eq:isoloss_curve_PD}.

\begin{table}[!t]
\centering
\small
\setlength{\tabcolsep}{6pt}
\caption{
Bivariate scaling fits per representation on the single-epoch grid.
$n$ is the number of grid points used in fitting.
MAE and RMSE are computed on the same grid points.
}
\label{tab:fit_summary}
\begin{tabular}{lcccc}
\toprule
Representation & $\alpha$ & $\beta$ & MAE & RMSE \\
\midrule
DeepSMILES & 0.0588 & 0.3624 & 0.0063 & 0.0076 \\
FragLink   & 0.0282 & 0.5214 & 0.0056 & 0.0074 \\
FragSeq    & 0.0189 & 0.5207 & 0.0062 & 0.0081 \\
SAFE       & 0.0200 & 0.2001 & 0.0053 & 0.0066 \\
SMILES     & 0.0171 & 0.4299 & 0.0084 & 0.0103 \\
\bottomrule
\end{tabular}
\vspace{-4mm}
\end{table}

\textbf{Compute-optimal allocation trends.}
The compute-optimal allocation is summarized by the tokens-per-parameter ratio
$\rho_{\mathrm{opt}}(C) = D_{\mathrm{opt}}(C) / P_{\mathrm{opt}}(C)$, where $(P_{\mathrm{opt}}(C),D_{\mathrm{opt}}(C))$
minimizes loss under the constraint $P D = C$.
Table~\ref{tab:rho_opt_summary} reports, for each representation, the compute range covered by the single-epoch grid,
the endpoint values of $\rho_{\mathrm{opt}}(C)$ and $L_{\mathrm{opt}}(C)$ over that range, and the corresponding correlations.
Across DeepSMILES, FragLink, FragSeq, and SMILES, $\rho_{\mathrm{opt}}(C)$ decreases with compute in log space.
In contrast, SAFE exhibits an opposite trend, where $\rho_{\mathrm{opt}}(C)$ increases with compute.
For all representations, $L_{\mathrm{opt}}(C)$ decreases as compute increases within the studied range.

\textbf{Representation dependence.}
Both views vary substantially across representations.
First, the absolute loss level under matched compute differs, shifting the full set of contours and isoFLOP curves.
Second, the compute-optimal token-to-parameter ratio $\rho_{\mathrm{opt}}(C)=D_{\mathrm{opt}}(C)/P_{\mathrm{opt}}(C)$ changes with compute in a representation-specific manner (Figure \ref{figure: overview}(c)).
For DeepSMILES, FragLink, FragSeq, and SMILES, $\rho_{\mathrm{opt}}(C)$ decreases as $C$ increases, indicating that the compute-optimal allocation becomes progressively more parameter-heavy at larger compute.
In contrast, SAFE exhibits the opposite trend in the current fit, where $\rho_{\mathrm{opt}}(C)$ increases with $C$, implying a progressively more token-heavy optimum at larger compute.
These differences indicate that compute-optimal scaling in molecular language modeling is not representation-invariant.

\textbf{Implications.}
The isoFLOP view shows that, under fixed compute, loss is not monotonic in model size and admits an interior optimum.
The isoLoss view further indicates that small models require substantially larger compute to reach the same loss.

\subsection{Longer Training on a Fixed Corpus}
\label{sec:longer_training}

This subsection evaluates longer training on a fixed corpus.
The same tokenized corpus is replayed for multiple epochs.
Therefore, training tokens $D$ and compute $C \propto P D$ increase with epochs, while corpus support is unchanged.
This analysis is reported as an auxiliary study and is not used as the primary basis for scaling conclusions.

For a fixed representation and a fixed dataset token budget
$B \in \{100\mathrm{M}, 300\mathrm{M}, 1\mathrm{B}, 3\mathrm{B}\}$,
an epoch-$e$ run consumes $D = eB$ tokens.
All runs are trained from scratch and evaluated at the end of each epoch.
Unless otherwise noted, $e \in \{1,\ldots,5\}$ is used.
An additional long run is performed for SMILES with $P{=}1\mathrm{M}$ and $B{=}100\mathrm{M}$ up to $e{=}10$.

\begin{figure}[t]
  \centering
  \begin{subfigure}[t]{0.49\linewidth}
    \centering
    \includegraphics[width=\linewidth]{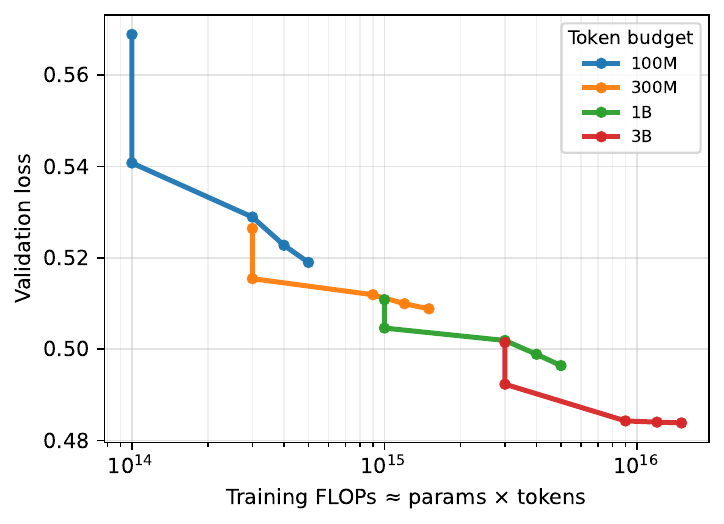}
    \vspace{-4mm}
    \caption{FragLink}
    \label{fig:longer_fraglink}
  \end{subfigure}
  \hfill
  \begin{subfigure}[t]{0.49\linewidth}
    \centering
    \includegraphics[width=\linewidth]{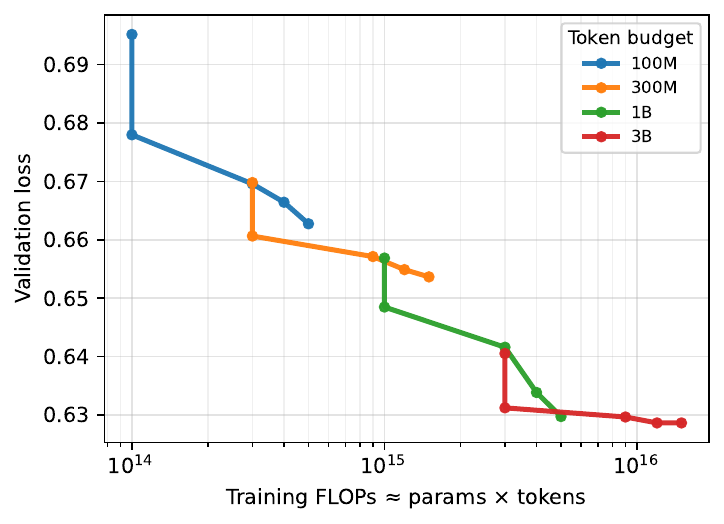}
    \vspace{-4mm}
    \caption{SMILES}
    \label{fig:longer_smiles}
  \end{subfigure}
  \caption{Longer training on a fixed corpus. Each panel reports end-of-epoch validation loss versus compute for repeated passes. See Appendix Figure \ref{fig:longer_more} for details.}
  \label{fig:longer_training}
  \vspace{-8mm}
\end{figure}

Figure~\ref{fig:longer_fraglink} and Figure~\ref{fig:longer_smiles} show that validation loss is generally reduced when training is extended to multiple epochs under a fixed corpus.
However, the marginal improvement decreases as epochs increase.
Across token budgets, most of the loss reduction is concentrated in early passes, while later passes yield small gains and can show mild fluctuations.

\begin{figure}[t]
  \centering\includegraphics[width=0.8\columnwidth]{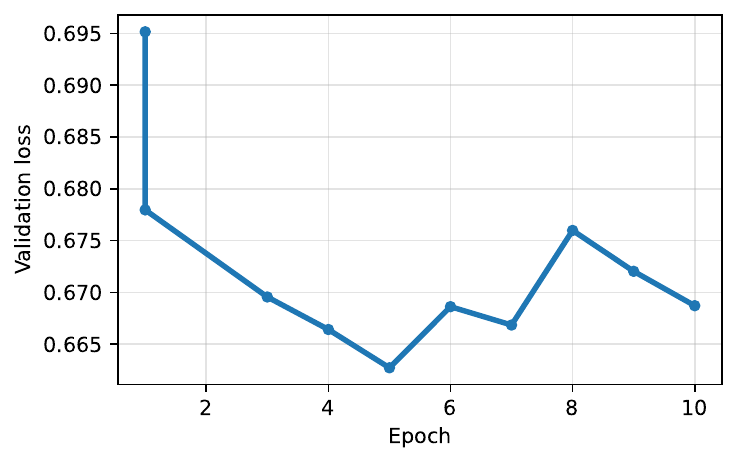}
  \vspace{-2mm}
  \caption{SMILES, $P{=}1$M, $B{=}100$M: end-of-epoch validation loss from epoch 1 to 10.}
  \label{fig:smiles_epoch10}
  \vspace{-6mm}
\end{figure}

The long run in Figure~\ref{fig:smiles_epoch10} further illustrates this behavior.
After several epochs, the loss trajectory becomes less smooth and improvements become limited.
This pattern is consistent with the fact that additional compute is spent on repeated tokens rather than on new tokens.

\textbf{Implication for scaling analysis.}
Longer training on a fixed corpus increases $D$ and $C$ without increasing corpus support.
Therefore, repeated passes are not equivalent to acquiring new tokens when interpreting scaling trends.
For this reason, the primary scaling analyses in Sec.~\ref{sec:loss_scaling}--Sec.~\ref{sec:isoflop_isoloss} are based on epoch-1 runs, where $D$ corresponds to a single pass over the constructed dataset token budget.

\subsection{Optimization Caveats}
\label{sec:opt_caveats}

This subsection clarifies which observables should not be used to draw scaling conclusions.
Two settings are discussed: \textit{de novo} generation metrics and goal-directed optimization benchmarks.
Both settings can be useful for application-oriented evaluation.
However, both can be weak or misleading as primary evidence for scaling laws.

\subsubsection{Metric Saturation}
\label{sec:metric_saturation}

Many commonly reported \textit{de novo} metrics saturate early.
Validity, uniqueness, novelty, and diversity often reach high values after short training.
They are also sensitive to sampling choices, including temperature and top-$k$.
These metrics can be adjusted without changing the pretrained model.
Therefore, they provide limited resolution for separating model capacity.

To make this concrete, we conduct a small-model long-duration sweep.
A $P{=}1$M model is trained on SMILES with a fixed dataset token budget at two scales.
For each scale, the model is trained from scratch for up to ten epochs.
For each checkpoint, \textit{de novo} metrics are evaluated under multiple sampling settings.
The results show that high validity can be achieved within a very small compute budget.
For example, SMILES-$1$M with $D{=}100$M tokens already reaches high validity within a short run on a single H100, as illustrated in Table \ref{table: Ten Epoch Train}.
Meanwhile, changes in temperature and top-$k$ produce large shifts in uniqueness and diversity.
To isolate sampling sensitivity, we additionally fix a single checkpoint and sweep $(T,k)$.
Large variations in uniqueness and diversity are observed under identical model weights, so these metrics cannot serve as the primary basis for scaling conclusions.
Taken together, these observations support a clear conclusion.
Saturated and sampling-sensitive \textit{de novo} metrics are not suitable as primary evidence for or against scaling laws.

\begin{table}[]
  \caption{The de novo molecules metrics (validity, uniqueness, diversity, and novelty) of the 1M parameters model with 100M training tokens in SMILES string.}
  \vspace{-2mm}
  \label{table: Ten Epoch Train}
  \begin{center}
    \begin{small}
      \begin{sc}
      \resizebox{\columnwidth}{!} {
        \begin{tabular}{lcccc}
          \toprule
          Epochs   & Validity  & Uniqueness  & Diversity  & Novelty  \\
          \midrule
          1        & 0.9370    & 1.0000  & 0.8732     & 0.9989   \\
          2        & 0.9530    & 1.0000  & 0.8727     & 0.9990   \\
          3        & 0.9590    & 1.0000  & 0.8754     & 1.0000   \\
          4        & 0.9610    & 1.0000  & 0.8718     & 0.9990   \\
          5        & 0.9510    & 1.0000  & 0.8723     & 0.9958   \\
          6        & 0.9680    & 1.0000  & 0.8743     & 0.9990   \\
          7        & 0.9520    & 1.0000  & 0.8738     & 1.0000   \\
          8        & 0.9540    & 1.0000  & 0.8706     & 1.0000   \\
          9        & 0.9670    & 1.0000  & 0.8734     & 1.0000   \\
          10       & 0.9620    & 1.0000  & 0.8708     & 0.9990   \\
          \bottomrule
        \end{tabular}
      }
      \end{sc}
    \end{small}
  \end{center}
  \vspace{-6mm}
\end{table}

\subsubsection{Optimization Dominance}
\label{sec:optimization_dominance}

Goal-directed optimization benchmarks add another confounder.
In these benchmarks, the reported score is often driven by reward shaping and search strategy.
The bottleneck is not necessarily the pretrained model.
Rather, the performance bottleneck often lies within the objective definition and the optimization procedure itself.

This issue is illustrated by the \textsc{Trio} framework~\cite{2025-12-Trio}.
\textsc{Trio} combines a GPT generator with tree search and a property-weighted objective.
In particular, QED and SA are used as weighted terms in the scoring function.
Under this design, the reported hit rate can approach $100\%$.
This outcome is mainly explained by the objective and the search procedure.
It does not imply that the pretrained model alone has solved a harder distribution modeling problem.
This leads to the second conclusion.
Optimization-dominated hit rates are not suitable as a basis for scaling claims about molecular language models.
They may reflect algorithmic choices rather than representational scaling.

\subsubsection{Implications}
\label{sec:caveat_implications}

The implications are straightforward.
Scaling claims should be grounded in compute-controlled pretraining loss trends and in downstream evaluations that remain discriminative.
In this work, scaling is therefore assessed primarily through validation loss $L(P,D)$ under compute-controlled sweeps.
\textit{De novo} metrics and optimization benchmarks are treated as secondary diagnostics.
They are used to explain why prior studies can reach negative conclusions, but not to define the existence of scaling.

\subsection{Downstream Task Evaluation: Property Prediction}
To validate the practical utility of the representations learned during pre-training, we fine-tuned our models on a comprehensive set of nine benchmark tasks from MoleculeNet. These tasks were grouped into three scientific domains to provide a holistic view of the model's predictive power: Biochemistry (BACE, HIV), Physiology (BBBP, ClinTox, Sider, Tox21), and Biophysics (ESOL, FreeSolv, Lipophilicity). The Comparisons against state-of-the-art (SOTA) models are provided in Table \ref{table: MoleculeNet-cla} and \ref{table: MoleculeNet-reg}. For each task, we evaluated models pre-trained across different scales, and the performance is summarized in Appendix \ref{sec: appendix E} Figure \ref{figure: Biochemistry MoleculeNet Scaling}, \ref{figure: Physiology MoleculeNet Scaling} and \ref{figure: Biophysics MoleculeNet Scaling}. 

\textbf{Biochemistry Tasks.}
On the BACE benchmark, we observe a clear and positive scaling trend for most representations. Performance, measured in ROC-AUC, consistently improves with increasing model size, particularly for the token-based representations SMILES and DeepSMILES. Notably, the FragLink representation demonstrates the best overall performance, consistently outperforming other methods across all model scales. This suggests that its fragment-based approach, which encapsulates key structural motifs, is highly effective for predicting the binding affinity targeted in this task. For the HIV task, a similar positive scaling trend is evident, where larger models generally yield better results. Although SAFE achieved the best results in this case, due to SAFE's own encoding problems, the number of samples it tested was only 83\% of the original test set. Therefore, we ignore the results of SAFE here. In this case, SMILES and DeepSMILES achieve the top performance tiers, indicating that atom-level representations, which preserve the complete and detailed topological structure of complex molecules, are particularly well-suited for modeling HIV inhibitors.

\textbf{Physiology Tasks.}
The physiology benchmarks reveal more diverse scaling behaviors. For BBBP, which is strongly linked to physicochemical properties, DeepSMILES shows a clear advantage and strong positive scaling, establishing its superiority for this task. The Tox21 and Sider tasks present more complex, non-monotonic scaling patterns. For instance, on Tox21, performance for several representations peaks at the 16M model size before declining. This phenomenon suggests a complex interplay between model capacity and generalization for these tasks \cite{schaeffer2023double}. On the Sider task, FragSeq demonstrates the most robust and superior performance, whereas other representations show more erratic behavior. On the ClinTox task, SMILES and DeepSMILES representations achieve very high performance, converging to near-perfect ROC-AUC scores ($>99.0$) with larger models, suggesting the task may have a performance ceiling that is easily reached.

\textbf{Biophysics Tasks.}
The biophysics tasks are all regression problems, where lower RMSE indicates better performance. On all three tasks, we observe a strong and consistent scaling trend: as model size increases, the prediction error (RMSE) steadily decreases. These results indicate that larger models are better able to capture the subtle physicochemical features that govern these properties. Across these three tasks, no single representation is universally dominant, but FragLink consistently demonstrates exceptional performance, achieving or closely approaching the lowest error rates in all cases. This highlights its robustness and strong capability for predicting continuous physical properties, likely due to its effective balance of structural abstraction and detailed information.

The fragment-based approach of FragLink demonstrates exceptional strength in select, high-impact tasks, such as BACE classification and the full suite of biophysics regression benchmarks. Its ability to encapsulate structural motifs appears to provide a powerful inductive bias for predicting properties governed by local chemical features and continuous physicochemical values. Conversely, atom-level representations show remarkable performance on a broader range of tasks. They achieve top-tier results in HIV, BBBP and ClinTox. Their strength lies in preserving the complete, fine-grained topological information of molecules, which proves critical for tasks involving complex global structures or those with easily learnable patterns. Additionally, our downstream evaluations reveal a highly nuanced performance landscape where the optimal molecular representation is strongly task-dependent.

\section{Conclusion}
In this work, we provide the first large-scale, systematic evidence that molecular language models follow predictable, compute-optimal scaling laws analogous to those in natural language processing. We show that the choice of molecular representation fundamentally shapes these laws and that scaling trends observed during pretraining reliably transfer to downstream tasks. These results offer a clear, data-driven path toward the efficient design and training of more powerful foundation models for molecular discovery.

\clearpage
\section*{Impact Statement}
This paper presents work whose goal is to advance the field of Machine
Learning. There are many potential societal consequences of our work, none
which we feel must be specifically highlighted here.

\bibliography{example_paper}

@article{2020-ScalingLaws,
  title={Scaling laws for neural language models},
  author={Kaplan, Jared and McCandlish, Sam and Henighan, Tom and Brown, Tom B and Chess, Benjamin and Child, Rewon and Gray, Scott and Radford, Alec and Wu, Jeffrey and Amodei, Dario},
  journal={arXiv preprint arXiv:2001.08361},
  year={2020}
}

@article{2022-Chinchilla,
  title={Training compute-optimal large language models},
  author={Hoffmann, Jordan and Borgeaud, Sebastian and Mensch, Arthur and Buchatskaya, Elena and Cai, Trevor and Rutherford, Eliza and Casas, Diego de Las and Hendricks, Lisa Anne and Welbl, Johannes and Clark, Aidan and others},
  journal={arXiv preprint arXiv:2203.15556},
  year={2022}
}

@article{2023-GPT4,
  title={Gpt-4 technical report},
  author={Achiam, Josh and Adler, Steven and Agarwal, Sandhini and Ahmad, Lama and Akkaya, Ilge and Aleman, Florencia Leoni and Almeida, Diogo and Altenschmidt, Janko and Altman, Sam and Anadkat, Shyamal and others},
  journal={arXiv preprint arXiv:2303.08774},
  year={2023}
}

@article{2021-MolGPT,
  title={MolGPT: molecular generation using a transformer-decoder model},
  author={Bagal, Viraj and Aggarwal, Rishal and Vinod, PK and Priyakumar, U Deva},
  journal={Journal of chemical information and modeling},
  volume={62},
  number={9},
  pages={2064--2076},
  year={2021},
  publisher={ACS Publications}
}

@article{2023-09-NMI-Scaling4chemical,
  title={Neural scaling of deep chemical models},
  author={Frey, Nathan C and Soklaski, Ryan and Axelrod, Simon and Samsi, Siddharth and Gomez-Bombarelli, Rafael and Coley, Connor W and Gadepally, Vijay},
  journal={Nature Machine Intelligence},
  volume={5},
  number={11},
  pages={1297--1305},
  year={2023},
  publisher={Nature Publishing Group UK London}
}

@article{2025-08-ICMLWS-NovoMolGen,
  title={NovoMolGen: Rethinking Molecular Language Model Pretraining},
  author={Chitsaz, Kamran and Balaji, Roshan and Fournier, Quentin and Bhatt, Nirav Pravinbhai and Chandar, Sarath},
  journal={arXiv preprint arXiv:2508.13408},
  year={2025}
}

@article{2021-04-PANS-ESM,
  title={Biological structure and function emerge from scaling unsupervised learning to 250 million protein sequences},
  author={Rives, Alexander and Meier, Joshua and Sercu, Tom and Goyal, Siddharth and Lin, Zeming and Liu, Jason and Guo, Demi and Ott, Myle and Zitnick, C Lawrence and Ma, Jerry and others},
  journal={Proceedings of the National Academy of Sciences},
  volume={118},
  number={15},
  pages={e2016239118},
  year={2021},
  publisher={National Academy of Sciences}
}

@article{ross2022large,
  title={Large-scale chemical language representations capture molecular structure and properties},
  author={Ross, Jerret and Belgodere, Brian and Chenthamarakshan, Vijil and Padhi, Inkit and Mroueh, Youssef and Das, Payel},
  journal={Nature Machine Intelligence},
  volume={4},
  number={12},
  pages={1256--1264},
  year={2022},
  publisher={Nature Publishing Group UK London}
}

@article{1988-SMILES,
  title={SMILES, a chemical language and information system. 1. Introduction to methodology and encoding rules},
  author={Weininger, David},
  journal={Journal of chemical information and computer sciences},
  volume={28},
  number={1},
  pages={31--36},
  year={1988},
  publisher={ACS Publications}
}

@article{2018-DeepSMILES,
  title={DeepSMILES: an adaptation of SMILES for use in machine-learning of chemical structures},
  author={O'Boyle, Noel and Dalke, Andrew},
  year={2018}
}

@article{2024-SAFE,
  title={Gotta be SAFE: a new framework for molecular design},
  author={Noutahi, Emmanuel and Gabellini, Cristian and Craig, Michael and Lim, Jonathan SC and Tossou, Prudencio},
  journal={Digital Discovery},
  volume={3},
  number={4},
  pages={796--804},
  year={2024},
  publisher={Royal Society of Chemistry}
}

@article{2024-02-Science-Evo,
  title={Sequence modeling and design from molecular to genome scale with Evo},
  author={Nguyen, Eric and Poli, Michael and Durrant, Matthew G and Kang, Brian and Katrekar, Dhruva and Li, David B and Bartie, Liam J and Thomas, Armin W and King, Samuel H and Brixi, Garyk and others},
  journal={Science},
  volume={386},
  number={6723},
  pages={eado9336},
  year={2024},
  publisher={American Association for the Advancement of Science}
}

@article{2024-10-NM-NucleotideTransformer,
  title={Nucleotide Transformer: building and evaluating robust foundation models for human genomics},
  author={Dalla-Torre, Hugo and Gonzalez, Liam and Mendoza-Revilla, Javier and Lopez Carranza, Nicolas and Grzywaczewski, Adam Henryk and Oteri, Francesco and Dallago, Christian and Trop, Evan and de Almeida, Bernardo P and Sirelkhatim, Hassan and others},
  journal={Nature Methods},
  volume={22},
  number={2},
  pages={287--297},
  year={2025},
  publisher={Nature Publishing Group US New York}
}

@article{2025-07-DataSaturationScaling,
  title={Scaling and Data Saturation in Protein Language Models},
  author={Spinner, Aviv and DeBenedictis, Erika and Hudson, Corey M},
  journal={arXiv preprint arXiv:2507.22210},
  year={2025}
}

@article{2020-10-ChemBERTa,
  title={ChemBERTa: large-scale self-supervised pretraining for molecular property prediction},
  author={Chithrananda, Seyone and Grand, Gabriel and Ramsundar, Bharath},
  journal={arXiv preprint arXiv:2010.09885},
  year={2020}
}

@article{2025-12-CommunChem-ChemFM,
  title={ChemFM as a scaling law guided foundation model pre-trained on informative chemicals},
  author={Cai, Feiyang and Zacour, Katelin and Zhu, Tianyu and Tzeng, Tzuen-Rong and Duan, Yongping and Liu, Ling and Pilla, Srikanth and Li, Gang and Luo, Feng},
  journal={Communications Chemistry},
  year={2025},
  publisher={Nature Publishing Group UK London}
}

@article{2025-11-DiversityBeatsScaling,
  title={Diversity Beats Size Scaling for Chemical Language Models},
  author={Medina, Borja and Tibo, Alessandro and He, Jiazhen and Janet, Jon Paul and {\"O}sterbacka, Nicklas},
  year={2025},
  journal={ChemRxiv}
}

@article{2025-12-Trio,
  title={Toward Closed-loop Molecular Discovery via Language Model, Property Alignment and Strategic Search},
  author={Ji, Junkai and Yang, Zhangfan and Xu, Dong and Bai, Ruibin and Li, Jianqiang and Hou, Tingjun and Zhu, Zexuan},
  journal={arXiv preprint arXiv:2512.09566},
  year={2025}
}

@article{hu2022lora,
  title={Lora: Low-rank adaptation of large language models.},
  author={Hu, Edward J and Shen, Yelong and Wallis, Phillip and Allen-Zhu, Zeyuan and Li, Yuanzhi and Wang, Shean and Wang, Lu and Chen, Weizhu and others},
  journal={ICLR},
  volume={1},
  number={2},
  pages={3},
  year={2022}
}

@article{wu2018moleculenet,
  title={MoleculeNet: a benchmark for molecular machine learning},
  author={Wu, Zhenqin and Ramsundar, Bharath and Feinberg, Evan N and Gomes, Joseph and Geniesse, Caleb and Pappu, Aneesh S and Leswing, Karl and Pande, Vijay},
  journal={Chemical science},
  volume={9},
  number={2},
  pages={513--530},
  year={2018},
  publisher={Royal Society of Chemistry}
}

@article{kipf2016semi,
  title={Semi-supervised classification with graph convolutional networks},
  author={Kipf, TN},
  journal={arXiv preprint arXiv:1609.02907},
  year={2016}
}

@article{xu2018powerful,
  title={How powerful are graph neural networks?},
  author={Xu, Keyulu and Hu, Weihua and Leskovec, Jure and Jegelka, Stefanie},
  journal={arXiv preprint arXiv:1810.00826},
  year={2018}
}

@article{schutt2018schnet,
  title={Schnet--a deep learning architecture for molecules and materials},
  author={Sch{\"u}tt, Kristof T and Sauceda, Huziel E and Kindermans, P-J and Tkatchenko, Alexandre and M{\"u}ller, K-R},
  journal={The Journal of chemical physics},
  volume={148},
  number={24},
  year={2018},
  publisher={AIP Publishing}
}

@inproceedings{lu2019molecular,
  title={Molecular property prediction: A multilevel quantum interactions modeling perspective},
  author={Lu, Chengqiang and Liu, Qi and Wang, Chao and Huang, Zhenya and Lin, Peize and He, Lixin},
  booktitle={Proceedings of the AAAI conference on artificial intelligence},
  volume={33},
  number={01},
  pages={1052--1060},
  year={2019}
}

@article{yang2019analyzing,
  title={Analyzing learned molecular representations for property prediction},
  author={Yang, Kevin and Swanson, Kyle and Jin, Wengong and Coley, Connor and Eiden, Philipp and Gao, Hua and Guzman-Perez, Angel and Hopper, Timothy and Kelley, Brian and Mathea, Miriam and others},
  journal={Journal of chemical information and modeling},
  volume={59},
  number={8},
  pages={3370--3388},
  year={2019},
  publisher={ACS Publications}
}

@article{hu2019strategies,
  title={Strategies for pre-training graph neural networks},
  author={Hu, Weihua and Liu, Bowen and Gomes, Joseph and Zitnik, Marinka and Liang, Percy and Pande, Vijay and Leskovec, Jure},
  journal={arXiv preprint arXiv:1905.12265},
  year={2019}
}

@inproceedings{hu2020gpt,
  title={Gpt-gnn: Generative pre-training of graph neural networks},
  author={Hu, Ziniu and Dong, Yuxiao and Wang, Kuansan and Chang, Kai-Wei and Sun, Yizhou},
  booktitle={Proceedings of the 26th ACM SIGKDD international conference on knowledge discovery \& data mining},
  pages={1857--1867},
  year={2020}
}

@article{wang2022molecular,
  title={Molecular contrastive learning of representations via graph neural networks},
  author={Wang, Yuyang and Wang, Jianren and Cao, Zhonglin and Barati Farimani, Amir},
  journal={Nature Machine Intelligence},
  volume={4},
  number={3},
  pages={279--287},
  year={2022},
  publisher={Nature Publishing Group UK London}
}

@article{liu2023rethinking,
  title={Rethinking tokenizer and decoder in masked graph modeling for molecules},
  author={Liu, Zhiyuan and Shi, Yaorui and Zhang, An and Zhang, Enzhi and Kawaguchi, Kenji and Wang, Xiang and Chua, Tat-Seng},
  journal={Advances in Neural Information Processing Systems},
  volume={36},
  pages={25854--25875},
  year={2023}
}

@article{fabian2020molecular,
  title={Molecular representation learning with language models and domain-relevant auxiliary tasks},
  author={Fabian, Benedek and Edlich, Thomas and Gaspar, H{\'e}l{\'e}na and Segler, Marwin and Meyers, Joshua and Fiscato, Marco and Ahmed, Mohamed},
  journal={arXiv preprint arXiv:2011.13230},
  year={2020}
}

@article{chithrananda2020chemberta,
  title={ChemBERTa: large-scale self-supervised pretraining for molecular property prediction},
  author={Chithrananda, Seyone and Grand, Gabriel and Ramsundar, Bharath},
  journal={arXiv preprint arXiv:2010.09885},
  year={2020}
}

@article{ahmad2022chemberta,
  title={Chemberta-2: Towards chemical foundation models},
  author={Ahmad, Walid and Simon, Elana and Chithrananda, Seyone and Grand, Gabriel and Ramsundar, Bharath},
  journal={arXiv preprint arXiv:2209.01712},
  year={2022}
}

@article{zhou2023uni,
  title={Uni-mol: A universal 3d molecular representation learning framework},
  author={Zhou, Gengmo and Gao, Zhifeng and Ding, Qiankun and Zheng, Hang and Xu, Hongteng and Wei, Zhewei and Zhang, Linfeng and Ke, Guolin},
  year={2023}
}

@article{yuksel2023selformer,
  title={SELFormer: molecular representation learning via SELFIES language models},
  author={Y{\"u}ksel, Atakan and Ulusoy, Erva and {\"U}nl{\"u}, Atabey and Do{\u{g}}an, Tunca},
  journal={Machine Learning: Science and Technology},
  volume={4},
  number={2},
  pages={025035},
  year={2023},
  publisher={IOP Publishing}
}

@article{priyadarsini2024self,
  title={Self-bart: A transformer-based molecular representation model using selfies},
  author={Priyadarsini, Indra and Takeda, Seiji and Hamada, Lisa and Brazil, Emilio Vital and Soares, Eduardo and Shinohara, Hajime},
  journal={arXiv preprint arXiv:2410.12348},
  year={2024}
}

@article{zeng2022deep,
  title={A deep-learning system bridging molecule structure and biomedical text with comprehension comparable to human professionals},
  author={Zeng, Zheni and Yao, Yuan and Liu, Zhiyuan and Sun, Maosong},
  journal={Nature communications},
  volume={13},
  number={1},
  pages={862},
  year={2022},
  publisher={Nature Publishing Group UK London}
}

@article{taylor2022galactica,
  title={Galactica: A large language model for science},
  author={Taylor, Ross and Kardas, Marcin and Cucurull, Guillem and Scialom, Thomas and Hartshorn, Anthony and Saravia, Elvis and Poulton, Andrew and Kerkez, Viktor and Stojnic, Robert},
  journal={arXiv preprint arXiv:2211.09085},
  year={2022}
}

@article{su2022molecular,
  title={A molecular multimodal foundation model associating molecule graphs with natural language},
  author={Su, Bing and Du, Dazhao and Yang, Zhao and Zhou, Yujie and Li, Jiangmeng and Rao, Anyi and Sun, Hao and Lu, Zhiwu and Wen, Ji-Rong},
  journal={arXiv preprint arXiv:2209.05481},
  year={2022}
}

@article{liu2023molxpt,
  title={Molxpt: Wrapping molecules with text for generative pre-training},
  author={Liu, Zequn and Zhang, Wei and Xia, Yingce and Wu, Lijun and Xie, Shufang and Qin, Tao and Zhang, Ming and Liu, Tie-Yan},
  journal={arXiv preprint arXiv:2305.10688},
  year={2023}
}

@article{liu2023multi,
  title={Multi-modal molecule structure--text model for text-based retrieval and editing},
  author={Liu, Shengchao and Nie, Weili and Wang, Chengpeng and Lu, Jiarui and Qiao, Zhuoran and Liu, Ling and Tang, Jian and Xiao, Chaowei and Anandkumar, Animashree},
  journal={Nature Machine Intelligence},
  volume={5},
  number={12},
  pages={1447--1457},
  year={2023},
  publisher={Nature Publishing Group UK London}
}

@article{luo2023molfm,
  title={Molfm: A multimodal molecular foundation model},
  author={Luo, Yizhen and Yang, Kai and Hong, Massimo and Liu, Xing Yi and Nie, Zaiqing},
  journal={arXiv preprint arXiv:2307.09484},
  year={2023}
}

@article{liu2023molca,
  title={Molca: Molecular graph-language modeling with cross-modal projector and uni-modal adapter},
  author={Liu, Zhiyuan and Li, Sihang and Luo, Yanchen and Fei, Hao and Cao, Yixin and Kawaguchi, Kenji and Wang, Xiang and Chua, Tat-Seng},
  journal={arXiv preprint arXiv:2310.12798},
  year={2023}
}

@article{zhang2024atomas,
  title={Atomas: Hierarchical alignment on molecule-text for unified molecule understanding and generation},
  author={Zhang, Yikun and Ye, Geyan and Yuan, Chaohao and Han, Bo and Huang, Long-Kai and Yao, Jianhua and Liu, Wei and Rong, Yu},
  journal={arXiv preprint arXiv:2404.16880},
  year={2024}
}

@article{balaji2023gpt,
  title={Gpt-molberta: Gpt molecular features language model for molecular property prediction},
  author={Balaji, Suryanarayanan and Magar, Rishikesh and Jadhav, Yayati and Farimani, Amir Barati},
  journal={arXiv preprint arXiv:2310.03030},
  year={2023}
}

@article{subramanian2016computational,
  title={Computational modeling of $\beta$-secretase 1 (BACE-1) inhibitors using ligand based approaches},
  author={Subramanian, Govindan and Ramsundar, Bharath and Pande, Vijay and Denny, Rajiah Aldrin},
  journal={Journal of chemical information and modeling},
  volume={56},
  number={10},
  pages={1936--1949},
  year={2016},
  publisher={ACS Publications}
}

@article{martins2012bayesian,
  title={A Bayesian approach to in silico blood-brain barrier penetration modeling},
  author={Martins, Ines Filipa and Teixeira, Ana L and Pinheiro, Luis and Falcao, Andre O},
  journal={Journal of chemical information and modeling},
  volume={52},
  number={6},
  pages={1686--1697},
  year={2012},
  publisher={ACS Publications}
}

@article{schaeffer2023double,
  title={Double descent demystified: Identifying, interpreting \& ablating the sources of a deep learning puzzle},
  author={Schaeffer, Rylan and Khona, Mikail and Robertson, Zachary and Boopathy, Akhilan and Pistunova, Kateryna and Rocks, Jason W and Fiete, Ila Rani and Koyejo, Oluwasanmi},
  journal={arXiv preprint arXiv:2303.14151},
  year={2023}
}
\bibliographystyle{icml2026}

\newpage
\appendix
\onecolumn
\section{Additional Preliminaries}
\label{sec: appendix A}

\subsection{Derivation of compute-optimal allocation}
\label{app:compute_opt_derivation}

This appendix derives the compute-optimal allocation $(P_{\mathrm{opt}},D_{\mathrm{opt}})$ under a fixed compute budget $C$.

\subsubsection{Setup}
\label{app:setup}

Assume the bivariate scaling law
\begin{equation}
\label{eq:app_bivariate_scaling}
L(P,D) = L_{\infty} + k_P P^{-\alpha} + k_D D^{-\beta},
\end{equation}
and the compute constraint for dense Transformers
\begin{equation}
\label{eq:app_compute_constraint}
C = \kappa P D.
\end{equation}
Absorb $\kappa$ into $C$ and use the simplified constraint $PD=C$, i.e.,
\begin{equation}
\label{eq:app_PD}
PD=C \quad\Longrightarrow\quad D=\frac{C}{P}.
\end{equation}

\subsubsection{Reduction to one variable}
\label{app:one_var}

Substituting Eq.~\eqref{eq:app_PD} into Eq.~\eqref{eq:app_bivariate_scaling} yields
\begin{equation}
\label{eq:app_LP}
L(P;C) = L_{\infty} + k_P P^{-\alpha} + k_D \left(\frac{C}{P}\right)^{-\beta}
       = L_{\infty} + k_P P^{-\alpha} + k_D C^{-\beta} P^{\beta}.
\end{equation}

\subsubsection{Optimality condition}
\label{app:opt_cond}

Differentiate Eq.~\eqref{eq:app_LP} with respect to $P$:
\begin{equation}
\label{eq:app_derivative}
\frac{d}{dP}L(P;C) = -\alpha k_P P^{-\alpha-1} + \beta k_D C^{-\beta} P^{\beta-1}.
\end{equation}
Setting $\frac{d}{dP}L(P;C)=0$ gives
\begin{equation}
\label{eq:app_stationary}
\beta k_D C^{-\beta} P^{\beta-1} = \alpha k_P P^{-\alpha-1}.
\end{equation}
Multiplying both sides by $P^{\alpha+1}$ yields
\begin{equation}
\label{eq:app_power_eq}
\beta k_D C^{-\beta} P^{\alpha+\beta} = \alpha k_P,
\end{equation}
hence
\begin{equation}
\label{eq:app_P_power}
P^{\alpha+\beta} = \frac{\alpha k_P}{\beta k_D} C^{\beta}.
\end{equation}
Therefore,
\begin{equation}
\label{eq:app_Popt}
P_{\mathrm{opt}}(C) =
\left(\frac{\alpha k_P}{\beta k_D}\right)^{\frac{1}{\alpha+\beta}}
C^{\frac{\beta}{\alpha+\beta}},
\end{equation}
and by the compute constraint,
\begin{equation}
\label{eq:app_Dopt}
D_{\mathrm{opt}}(C) =
\frac{C}{P_{\mathrm{opt}}(C)} =
\left(\frac{\beta k_D}{\alpha k_P}\right)^{\frac{1}{\alpha+\beta}}
C^{\frac{\alpha}{\alpha+\beta}}.
\end{equation}
Finally,
\begin{equation}
\label{eq:app_rhoopt}
\rho_{\mathrm{opt}}(C) \equiv \frac{D_{\mathrm{opt}}(C)}{P_{\mathrm{opt}}(C)}
=
\left(\frac{\beta k_D}{\alpha k_P}\right)^{\frac{2}{\alpha+\beta}}
C^{\frac{\alpha-\beta}{\alpha+\beta}}.
\end{equation}

\subsection{Log-linear regression for the one-dimensional approximation}
\label{app:loglin_regression}

Within the compute range covered in experiments, $\rho_{\mathrm{opt}}(C)$ is summarized by
\begin{equation}
\label{eq:app_rho1d}
\rho_{\mathrm{opt}}(C) \approx a_{\mathrm{repr}} C^{s_{\mathrm{repr}}}.
\end{equation}
Using base-10 logs,
\begin{equation}
\log_{10}\rho_{\mathrm{opt}}(C) \approx b_{\mathrm{repr}} + s_{\mathrm{repr}}\log_{10}C,
\qquad a_{\mathrm{repr}} = 10^{b_{\mathrm{repr}}}.
\end{equation}
Given $K$ compute levels $(C_k,\rho_k)$, define $x_k=\log_{10}C_k$ and $y_k=\log_{10}\rho_k$.
The least-squares estimates are
\begin{equation}
s_{\mathrm{repr}} =
\frac{\sum_{k=1}^{K}(x_k-\bar{x})(y_k-\bar{y})}{\sum_{k=1}^{K}(x_k-\bar{x})^2},
\qquad
b_{\mathrm{repr}} = \bar{y} - s_{\mathrm{repr}}\bar{x},
\qquad
a_{\mathrm{repr}}=10^{b_{\mathrm{repr}}}.
\end{equation}

\begin{table}[t]
  \centering
  \vspace{-1mm}
  \caption{
  Compute-optimal allocation and loss trends over the compute range covered by the single-epoch training grid.
  Compute is approximated as $C \propto P D$.
  }
  \label{tab:rho_opt_summary}
  \vspace{-2mm}
  \small
  \setlength{\tabcolsep}{4pt}
  \renewcommand{\arraystretch}{1.15}
  \resizebox{\columnwidth}{!}{
  \begin{tabular}{lcccccccc}
    \toprule
    Representation &
    $C_{\min}$ &
    $C_{\max}$ &
    $\rho_{\mathrm{opt}}(C_{\min})$ &
    $\rho_{\mathrm{opt}}(C_{\max})$ &
    $\mathrm{corr}(\log C,\log \rho_{\mathrm{opt}})$ &
    $L_{\mathrm{opt}}(C_{\min})$ &
    $L_{\mathrm{opt}}(C_{\max})$ &
    $\mathrm{corr}(\log C, L_{\mathrm{opt}})$ \\
    \midrule
    DeepSMILES & $1.00\times 10^{14}$ & $1.95\times 10^{18}$ & $2.186\times 10^{2}$ & $4.838\times 10^{1}$ & $-1.000$ & $0.674047$ & $0.551937$ & $-0.985$ \\
    FragLink   & $1.00\times 10^{14}$ & $1.95\times 10^{18}$ & $1.302\times 10^{3}$ & $1.337\times 10^{1}$ & $-1.000$ & $0.536930$ & $0.451283$ & $-0.984$ \\
    FragSeq    & $1.00\times 10^{14}$ & $1.95\times 10^{18}$ & $9.842\times 10^{3}$ & $6.532\times 10^{1}$ & $-1.000$ & $0.510008$ & $0.434508$ & $-0.993$ \\
    SAFE       & $1.00\times 10^{14}$ & $1.95\times 10^{18}$ & $8.218\times 10^{1}$ & $7.056\times 10^{2}$ & $+1.000$ & $0.679088$ & $0.554917$ & $-0.994$ \\
    SMILES     & $1.00\times 10^{14}$ & $1.95\times 10^{18}$ & $1.397\times 10^{5}$ & $1.593\times 10^{2}$ & $-1.000$ & $0.663775$ & $0.575229$ & $-0.999$ \\
    \bottomrule
  \end{tabular}
  }
  \vspace{-3mm}
\end{table}

\subsection{Derivation of Compute-Optimal Allocation}
This appendix provides a detailed mathematical derivation for the compute-optimal model size, $P_{opt}$, and number of training tokens, $D_{opt}$, under a fixed computational budget, $C$. The objective is to find the allocation of $P$ and $D$ that minimizes the validation loss as predicted by our bivariate scaling law.

The primary constraint is the total computational budget, C, measured in FLOPs. For a dense transformer model, the training compute is approximately proportional to the product of the number of parameters and the number of training tokens. We can express this iso-FLOP constraint as:
\begin{equation}
    C = k \cdot P \cdot D 
\end{equation}
For the purpose of this derivation, we can absorb the constant k into the definition of the compute budget, simplifying the constraint to:
\begin{equation}
    PD=C \Longrightarrow D=\frac{C}{P}
\end{equation}
Our goal is to minimize $L(P, D)$ subject to this constraint. We can achieve this by substituting the expression for $D$ from the constraint into the loss function, thereby creating a new loss function, L(P), which depends only on the single variable $P$ for a fixed budget $C$:
\begin{equation}
    L(P)=L_{\infty} + k_PP^{-\alpha} + k_DC^{-\beta}{P}^{\beta}
\end{equation}
To find the value of P that minimizes this loss function, we take the partial derivative of $L(P)$ with respect to $P$ and set it to zero. The term $L_{\infty}$ is a constant and its derivative is zero.
\begin{equation}
\begin{aligned}
    \frac{\partial L}{\partial P}=\frac{\partial (L_{\infty}+k_PP^{-\alpha}+k_DC^{-\beta}p^{\beta})}{\partial P}&=0 \\
    \frac{\alpha k_P}{P^{\alpha+1}}+\beta k_DC^{-\beta}P^{\beta-1}&=0
\end{aligned}
\end{equation}
We can now solve for $P$. Rearranging the terms to isolate the components related to $P$:
\begin{equation}
\begin{aligned}
    \beta k_D C^{-\beta} P^{\beta-1}&=-\frac{\alpha k_P}{P^{\alpha+1}} \\
    P^{\alpha+\beta}&=\frac{\alpha k_P}{\beta k_D}C^{\beta}
\end{aligned}
\end{equation}
Solving for $P$ gives us the optimal model size, $P_{opt}$, as a function of the compute budget $C$:
\begin{equation}
    P_{opt}(C)=(\frac{\alpha k_P}{\beta k_D})^{\frac{1}{\alpha+\beta}}C^{\frac{\beta}{\alpha+\beta}}
\end{equation}
Having found the optimal model size, we can use the compute constraint to find the corresponding optimal number of training tokens, $D_{opt}$:
\begin{equation}
\begin{aligned}
    D_{opt}(C) &=\frac{C}{P_{opt}(C)} \\
               &=(\frac{\beta k_D}{\alpha k_P})^{\frac{1}{\alpha+\beta}}C^{\frac{\alpha}{\alpha+\beta}}
\end{aligned}
\end{equation}
From these results, we can derive the scaling law for the optimal tokens-per-parameter ratio, $\rho_{opt}$:
\begin{equation}
\begin{aligned}
    \rho_{opt}(C) &=\frac{D_{opt}(C)}{P_{opt}(C)} \\
                  &=\frac{(\frac{\beta k_D}{\alpha k_P})^{\frac{1}{\alpha+\beta}}C^{\frac{\alpha}{\alpha+\beta}}}{(\frac{\alpha k_P}{\beta k_D})^{\frac{1}{\alpha+\beta}}C^{\frac{\beta}{\alpha+\beta}}} \\
                  &=(\frac{\beta k_D}{\alpha k_P})^{\frac{2}{\alpha+\beta}}C^\frac{\alpha-\beta}{\alpha+\beta}
\end{aligned}
\end{equation}
This final expression is particularly insightful. Our empirical results consistently show that for all molecular representations, the data scaling exponent $\beta$ is greater than the parameter scaling exponent $\alpha$ (i.e., $\beta > \alpha$). Consequently, the exponent on $C$ in the expression for $\rho_{opt}(C)$ is negative:
\begin{equation}
    \frac{\alpha-\beta}{\alpha+\beta} < 0
\end{equation}
This leads to the important conclusion that as the total computational budget $C$ increases, the optimal strategy involves training with a proportionally smaller number of tokens per parameter. In other words, while both model size and data size should be increased with more compute, the model size should be increased at a faster rate than the data size to remain on the compute-optimal frontier.

\subsection{One-dimensional Power-Law Approximation for Optimal Ratio}
While the bivariate scaling law provides a complete description, the behavior of the optimal tokens-per-parameter ratio, $\rho_{opt}(C)$, can be further understood through a simplified, one-dimensional power-law approximation. Within the compute range covered by our experiments (approximately $10^{15}$ to $10^{18}$ FLOPs), we observe that the optimal ratio exhibits a stable, log-linear decrease as a function of the compute budget. This allows us to model the relationship as:
\begin{equation}
    \rho_{opt}(C) \approx a_{repr}C^{s_{repr}}
\end{equation}
where $a_{repr}$ and $s_{repr}$ are two scalar parameters specific to each representation. These parameters can be determined via a simple log-log linear regression on the derived optimal points. The exponent, $s_{repr}$, is particularly insightful as it quantifies the rate of contraction for the optimal $D/P$ ratio. It can be directly interpreted as a ``contraction factor", $10^{s_{repr}}$, which describes how much the optimal tokens-per-parameter ratio shrinks for every order-of-magnitude increase in the compute budget. A linear least-squares fit on the unified compute points yields the parameters shown in Table \ref{table: One-dimensional-Power-Law}. The negative values for $s_{repr}$ confirm that for all representations, the optimal ratio is a monotonically decreasing function of compute, exhibiting stable power-law contraction.

\begin{table*}[]
  \caption{Optimal Tokens per parameter power law slope $s_{repr}$ and scaling factor $10^{s_{repr}}$ expressed in five molecule languages}
  \label{table: One-dimensional-Power-Law}
  \begin{center}
    \begin{small}
      \begin{sc}
        \begin{tabular}{lccccc}
          \toprule
          Parameter         & SMILES    & DeepSMILES  & SAFE     & FragSeq  & FragLink  \\
          \midrule
          $s_{repr}$        & -0.2841   & -0.2447     & -0.2128  & -0.1724  & -0.2417   \\
          $10^{s_{repr}}$   &  0.5198   &  0.5693     &  0.6126  &  0.6723  & 0.5723    \\
          $b_{repr}$        &  5.8078   &  5.1786     &  4.5852  &  4.1096  & 512142    \\
          \bottomrule
        \end{tabular}
      \end{sc}
    \end{small}
  \end{center}
  \vskip -0.1in
\end{table*}

\section{Molecular String Representations}
\label{sec: appendix B}

In this work, we evaluate five distinct string-based molecular representations. These can be broadly categorized into atom-level representations, which linearize the molecular graph atom-by-atom, and fragment-level representations, which treat molecular sub-structures as the fundamental units of tokenization. 

\subsection{Atom-Level Representations}

\subsubsection{SMILES}
SMILES represents a canonical and widely adopted method for linearizing molecular graphs via a depth-first traversal. Atoms are denoted by their elemental symbols, with aromaticity indicated by lowercase letters, while structural features such as branches and ring closures are encoded using parentheses and numerical digits, respectively. This representation is valued for its compactness and direct compatibility with the vast majority of existing cheminformatics toolchains. However, its grammatical structure is notoriously fragile; minor perturbations to the string can easily lead to syntactically invalid representations due to unmatched parentheses or ring closures. Furthermore, a single molecule can often be described by numerous equivalent SMILES strings, introducing a high degree of redundancy.

\subsubsection{DeepSMILES}
The DeepSMILES formalism was developed to address the syntactic fragility inherent in SMILES by systematically rewriting the grammar for branching and ring closures. Core to its design is the replacement of paired parentheses with a stack depth mechanism, where the number of consecutive closing parentheses indicates the depth of traversal return. Similarly, it encodes ring closures using local positional information rather than long-range numerical matching. This transformation renders the syntax more regular and locally decidable, significantly reducing the probability of generating invalid strings during autoregressive sampling and making it better suited for character-level language models.

\subsection{Fragment-Level Representations}

\subsubsection{SAFE}
SAFE is a structurally-aware representation that builds upon the SMILES foundation. It operates by partitioning the molecular graph into chemically meaningful substructures, such as functional groups and ring systems, based on a set of predefined rules. These larger local environments are then compressed into single tokens. Compared to SMILES, SAFE explicitly encodes higher-level structural units, but this comes at the cost of a larger vocabulary and a more symbolic appearance. This representation is particularly advantageous for pre-training scenarios where the goal is to direct the model's focus toward functional motifs and scaffold patterns, thereby mitigating noise from low-level syntactic rules.

\subsubsection{FragSeq}
FragSeq introduces a fragment-based serialization designed to increase the information density per token. The molecular graph is first decomposed into a series of structural fragments, each corresponding to a local chemical motif, which are then concatenated into a single sequence using a ``[SEP]" separator token. Within each fragment, attachment points are denoted by a generic ``*" placeholder. This design shifts complex topological information, such as ring structures and side chains, from the high-level sequence grammar into the internal structure of the fragment tokens. Consequently, the model operates on sequences of ``fragment-level" scaffold patterns rather than repeatedly learning the complex syntax of parentheses and ring closures at the character level.

\subsubsection{FragLink}
FragLink introduces systematic improvements to the original FragSeq by resolving ambiguities in fragment connectivity and stabilizing training dynamics. It presents two key innovations. First, it introduces directional connection markers (``[*+]" for start and ``[*-]" for end) to eliminate the ambiguity of the generic ``*" token, dramatically increasing the success rate of reconstruction to over 99.95\%. Second, it constrains the connection topology to a logical chain structure, where each fragment connects only to its immediate neighbors. This simplification of the grammar suppresses complex non-local dependencies, transforming the graph generation problem into a more strictly sequential process and making the representation more stable across different data and model scales.

\section{More Experimental Settings}
\label{sec: appendix C}

\subsection{Details of Datasets}
\subsubsection{Pre-training Dataset}
The pre-training dataset includes over 1 billion unlabeled molecules drawn from the ZINC and UniChem database, used for node-level self-supervised pre-training. 

\subsubsection{Downstram Dataset}
To evaluate model performance, we utilized nine datasets from MoleculeNet \cite{wu2018moleculenet}, described as follows:

\begin{itemize}
    \item[$\bullet$] \textbf{BACE} \cite{subramanian2016computational}: The qualitative binding data for inhibitors targeting human $\beta$-secretase 1.
    \item[$\bullet$] \textbf{HIV}: Experimental data for inhibiting HIV replication abilities.
    \item[$\bullet$] \textbf{BBBP} \cite{martins2012bayesian}: Evaluates blood-brain barrier penetration based on membrane permeability.
    \item[$\bullet$] \textbf{SIDER}: Marketed drugs and 27 system organ classes adverse drug reactions (ADR).
    \item[$\bullet$] \textbf{Tox21}: The toxicity data for 12 biological targets, including nuclear receptors and stress response pathways.
    \item[$\bullet$] \textbf{ClinTox}: Drugs approved by the FDA and drugs that have failed clinical trials for toxicity reasons.
    \item[$\bullet$] \textbf{ESOL}: Contains water solubility data.
    \item[$\bullet$] \textbf{FreeSolv}: The experimental and calculated hydration free energy of small molecules in water.
    \item[$\bullet$] \textbf{Lipophlicity}: The experimental results of octanol/water distribution coefficient (logD at pH 7.4).
\end{itemize}

\begin{table*}[]
  \caption{MoleculeNet dataset's Train/Valid/Test samples in different representations.}
  \label{table: MoleculeNet-TVT}
  \begin{center}
    \begin{small}
      \begin{sc}
        \begin{tabular}{lcccc}
          \toprule
            Dataset                         & Representation Type    & Train   & Valid    & Test  \\
            \midrule
            \multirow{5}{*}{BACE}           & SMILES                 & 1209    & 151      & 152   \\
                                            & DeepSMILES             & 1209    & 151      & 152   \\
                                            & SAFE                   & 1206    & 149      & 152   \\
                                            & FragSeq                & 1209    & 151      & 152   \\
                                            & FragLink               & 1205    & 151      & 152   \\
            \midrule
            \multirow{5}{*}{HIV}            & SMILES                 & 32900   & 4113     & 4113  \\
                                            & DeepSMILES             & 32900   & 4113     & 4113  \\
                                            & SAFE                   & 27195   & 3401     & 3409  \\
                                            & FragSeq                & 30831   & 3853     & 3839  \\
                                            & FragLink               & 30831   & 3853     & 3839  \\
            \midrule
            \multirow{5}{*}{BBBP}           & SMILES                 & 1639    & 205      & 194   \\
                                            & DeepSMILES             & 1639    & 205      & 194   \\
                                            & SAFE                   & 1415    & 186      & 172   \\
                                            & FragSeq                & 1557    & 198      & 180   \\
                                            & FragLink               & 1538    & 195      & 177   \\
            \midrule
            \multirow{5}{*}{SIDER}          & SMILES                 & 1140    & 143      & 143   \\
                                            & DeepSMILES             & 1140    & 143      & 143   \\
                                            & SAFE                   & 971     & 123      & 126   \\
                                            & FragSeq                & 1017    & 133      & 127   \\
                                            & FragLink               & 1012    & 133      & 126   \\
            \midrule
            \multirow{5}{*}{Tox21}          & SMILES                 & 6264    & 783      & 783   \\
                                            & DeepSMILES             & 6264    & 783      & 783   \\
                                            & SAFE                   & 4587    & 618      & 603   \\
                                            & FragSeq                & 6132    & 757      & 763   \\
                                            & FragLink               & 6042    & 741      & 749   \\
            \midrule
            \multirow{5}{*}{ClinTox}        & SMILES                 & 1185    & 148      & 143   \\
                                            & DeepSMILES             & 1185    & 148      & 143   \\
                                            & SAFE                   & 1035    & 130      & 127   \\
                                            & FragSeq                & 1170    & 148      & 141   \\
                                            & FragLink               & 1151    & 140      & 138   \\
            \midrule
            \multirow{5}{*}{ESOL}           & SMILES                 & 901     & 113      & 113   \\
                                            & DeepSMILES             & 901     & 113      & 113   \\
                                            & SAFE                   & 442     & 52       & 62    \\
                                            & FragSeq                & 901     & 113      & 113   \\
                                            & FragLink               & 900     & 113      & 113   \\
            \midrule
            \multirow{5}{*}{FreeSolv}       & SMILES                 & 512     & 64       & 65    \\
                                            & DeepSMILES             & 512     & 64       & 65    \\
                                            & SAFE                   & 185     & 23       & 27    \\
                                            & FragSeq                & 512     & 64       & 65    \\
                                            & FragLink               & 512     & 63       & 65    \\
            \midrule
            \multirow{5}{*}{Lipophilicity}  & SMILES                 & 3360    & 420      & 420   \\
                                            & DeepSMILES             & 3360    & 420      & 420   \\
                                            & SAFE                   & 3238    & 399      & 403   \\
                                            & FragSeq                & 3359    & 420      & 420   \\
                                            & FragLink               & 3340    & 414      & 414   \\
          \bottomrule
        \end{tabular}
      \end{sc}
    \end{small}
  \end{center}
  \vskip -0.1in
\end{table*}

\subsection{Baselines}
\label{sec: appendix C: baseline}
\begin{itemize}
    \item[$\bullet$] \textbf{GCN} \cite{kipf2016semi} is a foundational Graph Neural Network architecture that learns node representations by aggregating information from their local graph neighborhoods. It adapts the principles of convolutional neural networks to graph-structured data, enabling the learning of features directly from molecular graphs. GCN serves as a fundamental baseline for graph-based molecular property prediction.
    \item[$\bullet$] \textbf{GIN} \cite{xu2018powerful} is an advanced GNN architecture designed to be as powerful as the Weisfeiler-Lehman (WL) test in distinguishing non-isomorphic graphs. By employing a specific aggregation function, GIN provides a theoretically grounded framework for capturing complex graph structures, making it a powerful baseline for molecular representation learning.
    \item[$\bullet$] \textbf{SchNet} \cite{schutt2018schnet} is a deep learning architecture specifically designed for modeling quantum-chemical properties of molecules. It operates on atomistic systems by representing them as graphs and uses continuous-filter convolutions to learn rich, chemically-aware representations of atomic environments. SchNet is particularly effective for predicting properties that depend on fine-grained geometric and elemental information.
    \item[$\bullet$] \textbf{MGCN} \cite{lu2019molecular} is a GNN model that learns multi-level representations of atoms by capturing features from varying neighborhood sizes. It is designed to model the hierarchical nature of molecular structures, from individual atoms to larger functional groups, providing a more comprehensive view of the molecular graph for property prediction.
    \item[$\bullet$] \textbf{D-MPNN} \cite{yang2019analyzing} is a GNN framework that operates on directed molecular graphs, where messages are passed along bonds rather than between atoms. This bond-centric approach avoids redundant message passing cycles in rings and has been shown to be highly effective and efficient for learning molecular representations, establishing it as a strong baseline in cheminformatics.
    \item[$\bullet$] \textbf{AttrMask} \cite{hu2019strategies} is a self-supervised learning strategy for pre-training Graph Neural Networks. It learns representations by masking atom or bond attributes and training the model to predict the masked information from its context. This pre-training task forces the GNN to learn a deep understanding of local chemical environments and graph topology.
    \item[$\bullet$] \textbf{GPT-GNN} \cite{hu2020gpt} is a generative pre-training framework for GNNs. It pre-trains a GNN by learning to generate molecular graphs, including both their structure and node/edge attributes, in an autoregressive manner. This generative pre-training allows the model to learn complex structural dependencies, which can then be transferred to downstream property prediction tasks.
    \item[$\bullet$] \textbf{MolCLR-GCN} \cite{wang2022molecular} represents a specific implementation of the MolCLR framework that uses a Graph Convolutional Network (GCN) as its underlying encoder. It leverages the principles of contrastive learning, where the GCN is trained to maximize agreement between different augmented views of the same molecule, thereby learning robust and transferable graph-level representations.
    \item[$\bullet$] \textbf{MolCLR-GIN} \cite{wang2022molecular} is another variant of the MolCLR framework, this time using a Graph Isomorphism Network (GIN) as the graph encoder. By combining the powerful discriminative capabilities of GIN with a contrastive self-supervised learning objective, this model is designed to learn highly effective representations for a wide range of downstream tasks.
    \item[$\bullet$] \textbf{SimSGT} \cite{liu2023molca} is a self-supervised learning framework for Graph Transformers. It introduces a novel contrastive learning strategy tailored for transformers, where the model learns by distinguishing between positive and negative pairs of subgraphs. SimSGT is designed to leverage the global receptive field of transformers to capture long-range interactions within molecules.
    \item[$\bullet$] \textbf{MolBERT} \cite{fabian2020molecular} is a transformer-based model that adapts the BERT architecture for the chemical domain. It is pre-trained on a large corpus of SMILES strings using a masked language modeling objective, where it learns to predict masked tokens (atoms or sub-structures). MolBERT was one of the pioneering works demonstrating the power of large-scale pre-training for molecular property prediction.
    \item[$\bullet$] \textbf{ChemBERTa} \cite{chithrananda2020chemberta} is another transformer model based on the RoBERTa architecture, pre-trained on a massive dataset of chemical molecules. It is designed to learn a general-purpose representation of chemical space that can be effectively fine-tuned for a wide variety of downstream tasks, showcasing the transferability of knowledge from large-scale, unsupervised pre-training.
    \item[$\bullet$] \textbf{ChemBERTa2} \cite{ahmad2022chemberta} is an improved iteration of the ChemBERTa model. It is pre-trained on an even larger and more diverse dataset of molecules and incorporates optimizations to the training procedure. The goal of ChemBERTa2 is to provide a more powerful and robust foundation model for chemistry, capable of achieving state-of-the-art performance on a broader set of benchmarks.
    \item[$\bullet$] \textbf{Uni-Mol} \cite{zhou2023uni} is a versatile pre-trained model for universal 3D molecular representation learning. It uniquely combines a transformer architecture with a focus on capturing 3D spatial information. Pre-trained on a massive dataset of molecular conformations, Uni-Mol learns a representation that is sensitive to both the 2D graph topology and the 3D geometry of molecules.
    \item[$\bullet$] \textbf{SELFormer} \cite{yuksel2023selformer} is a transformer-based architecture that introduces a novel self-supervised learning strategy for molecular representation. It is designed to learn chemically meaningful features by focusing on specific structural or electronic properties during its pre-training phase, aiming to create representations that are more aligned with downstream quantum mechanical or physicochemical tasks.
    \item[$\bullet$] \textbf{MolFormer-XL} \cite{ross2022large} is a large-scale, transformer-based masked language model pre-trained on 1.1 billion SMILES strings. It adapts the RoBERTa architecture and introduces a novel Rotary Position Embedding (RoPE) suitable for SMILES. Its strong performance has established it as a leading method for transfer learning in cheminformatics.
    \item[$\bullet$] \textbf{SELF-BART} \cite{priyadarsini2024self} is a generative pre-trained transformer based on the BART architecture, specifically adapted for chemical language. It is pre-trained on a denoising auto-encoding objective, where it learns to reconstruct corrupted SMILES strings, forcing the model to learn a robust and holistic understanding of chemical syntax and structure.
    \item[$\bullet$] \textbf{KV-PLM} \cite{zeng2022deep} is a pre-trained language model that introduces a key-value prediction mechanism. Instead of only predicting masked tokens, it also learns to predict associated ``values", allowing it to store and retrieve information in a structured manner. This framework is designed to enhance the model's ability to reason about and predict functional properties.
    \item[$\bullet$] \textbf{Galactica} \cite{taylor2022galactica} is a large-scale, general-purpose scientific language model trained on a vast corpus of scientific text, including papers, reference material, and chemical data like SMILES strings. It is designed to store, combine, and reason about scientific knowledge, making it a powerful baseline for tasks that require a broad scientific understanding.
    \item[$\bullet$] \textbf{MoMu} \cite{su2022molecular} is a multi-modal model that learns to connect molecular structures with their corresponding natural language descriptions. By pre-training on a dataset of molecules paired with text, MoMu learns a joint representation space, enabling it to perform tasks like text-based molecule retrieval and captioning, as well as property prediction.
    \item[$\bullet$] \textbf{MolXPT} \cite{liu2023molxpt} is a transformer model designed for explainable molecular property prediction. It is trained to not only predict a property but also to identify the key sub-structures or atoms that are responsible for that prediction. This focus on interpretability makes it a valuable tool for understanding structure-activity relationships.
    \item[$\bullet$] \textbf{MoleculeSTM} \cite{liu2023multi} is a model based on the principles of self-taught learning, where a transformer is pre-trained on a large unlabeled dataset of SMILES strings and then fine-tuned on specific downstream tasks. It represents a strong baseline for standard transformer-based transfer learning in chemistry.
    \item[$\bullet$] \textbf{MolFM} \cite{luo2023molfm} is a foundational model for chemistry that is pre-trained on a massive and diverse dataset of molecules. It is designed to serve as a general-purpose "molecular foundation model" that can be adapted to a wide range of tasks, from property prediction to generative chemistry, embodying the trend towards large, unified models for science.
    \item[$\bullet$] \textbf{MolCA} \cite{liu2023molca} is a model that incorporates a contrastive learning approach with a SMILES-based transformer. It learns by maximizing the agreement between representations of similar molecules while distinguishing them from dissimilar ones, a technique designed to create a well-structured and semantically meaningful embedding space.
    \item[$\bullet$] \textbf{Atomas} \cite{zhang2024atomas} is an advanced model that aims to capture multi-scale information within molecules. It is designed to learn features ranging from the atomic level to the level of functional groups and entire scaffolds, integrating this hierarchical information to make more accurate predictions.
    \item[$\bullet$] \textbf{GPT-MolBERTa} \cite{balaji2023gpt} is a hybrid model that combines the strengths of both GPT-style generative pre-training and BERT-style masked language modeling. It is pre-trained on a dual objective to learn both generative and discriminative capabilities, aiming to create a more versatile and powerful representation for a wide range of chemical tasks.
\end{itemize}

\section{Additional Pre-training Analysis}
\label{sec: appendix D}

This appendix contains supplementary figures and detailed analyses from the pre-training evaluation stage, providing robust empirical support for the conclusions presented in the main text.

\begin{figure*}[]
  \vskip 0.2in
    \begin{center}
        \begin{subfigure}[]{0.48\linewidth}
            \centering
            \includegraphics[width=\linewidth]{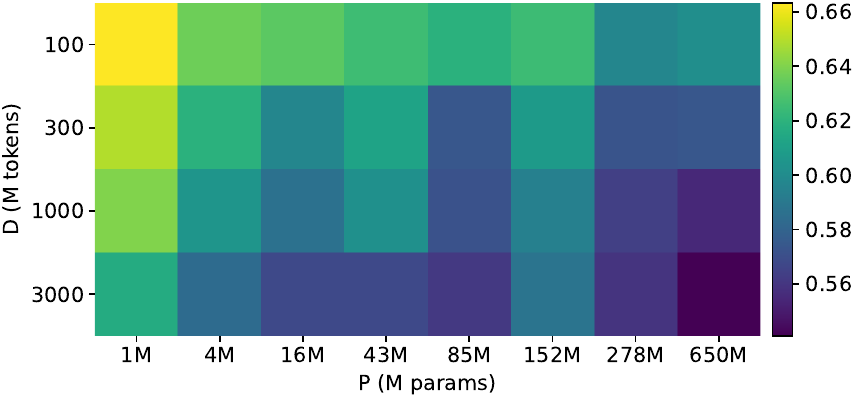}
            \caption{DeepSMILES}
        \end{subfigure}
        \begin{subfigure}[]{0.48\linewidth}
            \centering
            \includegraphics[width=\linewidth]{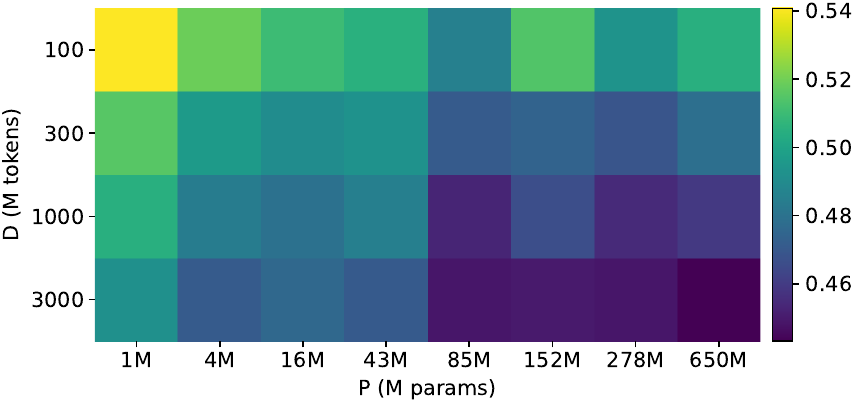}
            \caption{FragLink}
        \end{subfigure}
        \begin{subfigure}[]{0.48\linewidth}
            \centering
            \includegraphics[width=\linewidth]{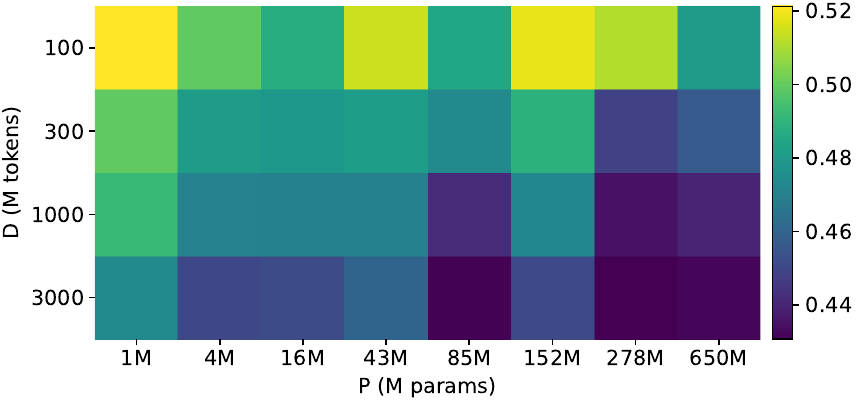}
            \caption{FragSeq}
        \end{subfigure}
        \begin{subfigure}[]{0.48\linewidth}
            \centering
            \includegraphics[width=\linewidth]{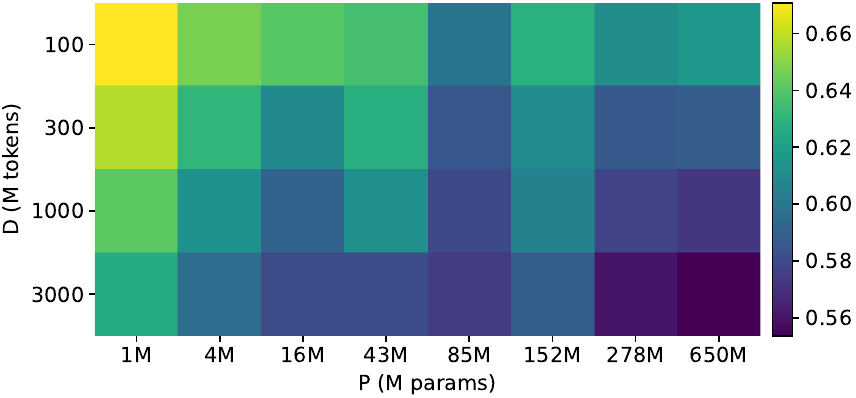}
            \caption{SAFE}
        \end{subfigure}
        \begin{subfigure}[]{0.48\linewidth}
            \centering
            \includegraphics[width=\linewidth]{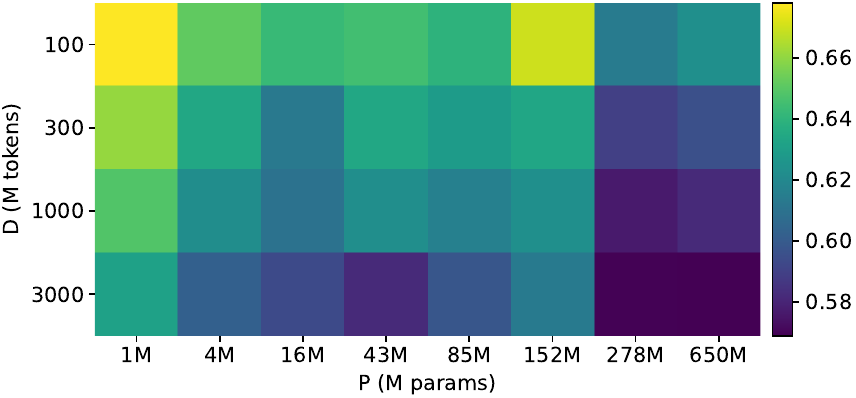}
            \caption{SMILES}
        \end{subfigure}    
    \caption{Heatmaps of validation loss for the five molecular representations as a function of model size and data size.}
    \label{figure: valid_loss_heatmap}
  \end{center}
\end{figure*}

\begin{figure*}[]
  \vskip 0.2in
    \begin{center}
        \begin{subfigure}[]{0.32\linewidth}
            \centering
            \includegraphics[width=\linewidth]{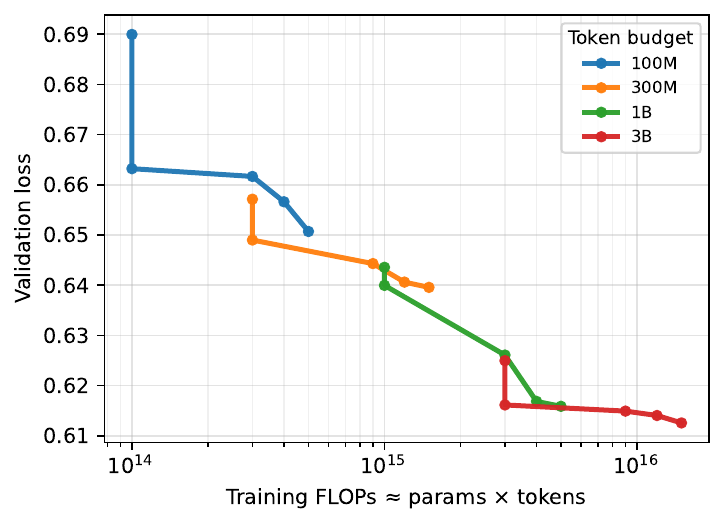}
            \caption{DeepSMILES}
        \end{subfigure}
        \begin{subfigure}[]{0.32\linewidth}
            \centering
            \includegraphics[width=\linewidth]{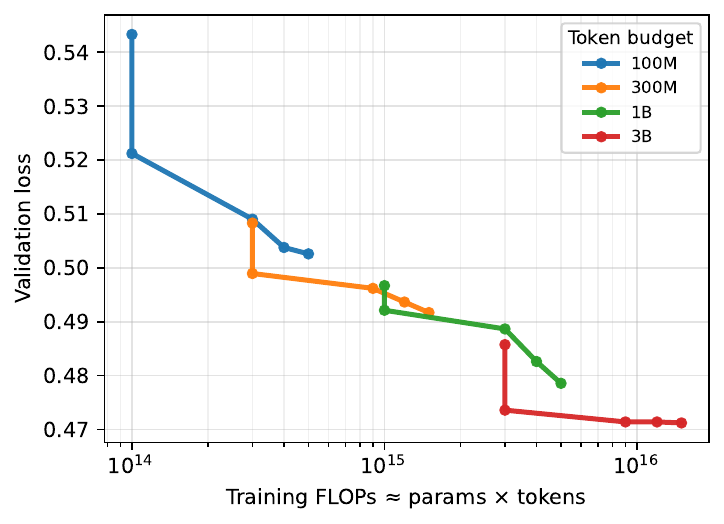}
            \caption{FragSeq}
        \end{subfigure}
        \begin{subfigure}[]{0.32\linewidth}
            \centering
            \includegraphics[width=\linewidth]{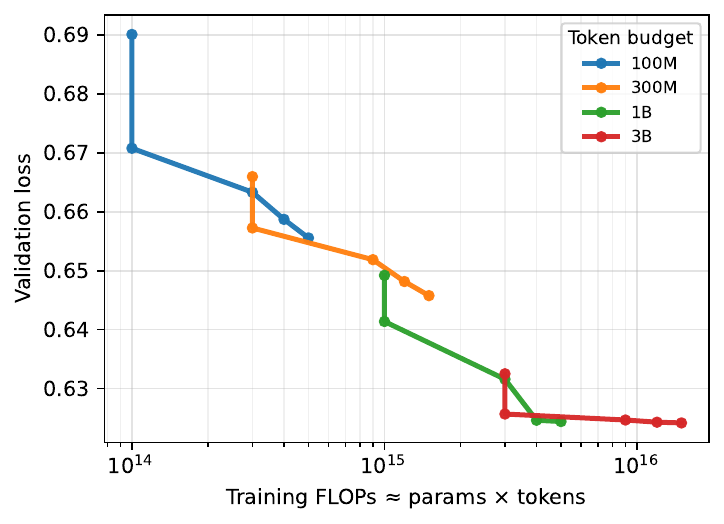}
            \caption{SAFE}
        \end{subfigure}
    \caption{Longer training on a fixed corpus. Each panel reports end-of-epoch validation loss versus compute for repeated passes.}
    \label{fig:longer_more}
  \end{center}
\end{figure*}

\begin{figure*}[]
  \vskip 0.2in
    \begin{center}
        \begin{subfigure}[]{0.48\linewidth}
            \centering
            \includegraphics[width=\linewidth]{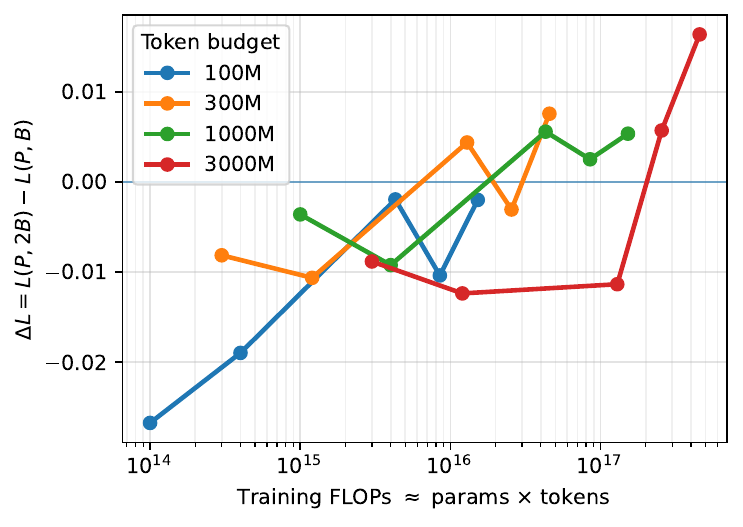}
            \caption{DeepSMILES}
        \end{subfigure}
        \begin{subfigure}[]{0.48\linewidth}
            \centering
            \includegraphics[width=\linewidth]{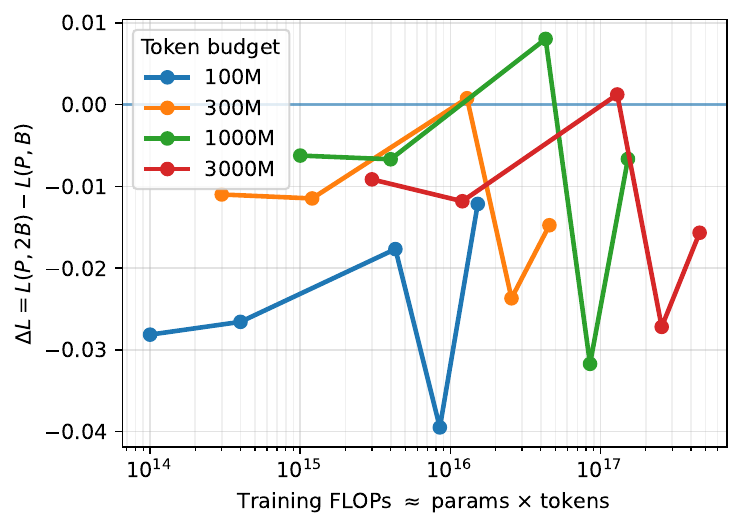}
            \caption{FragLink}
        \end{subfigure}
        \begin{subfigure}[]{0.48\linewidth}
            \centering
            \includegraphics[width=\linewidth]{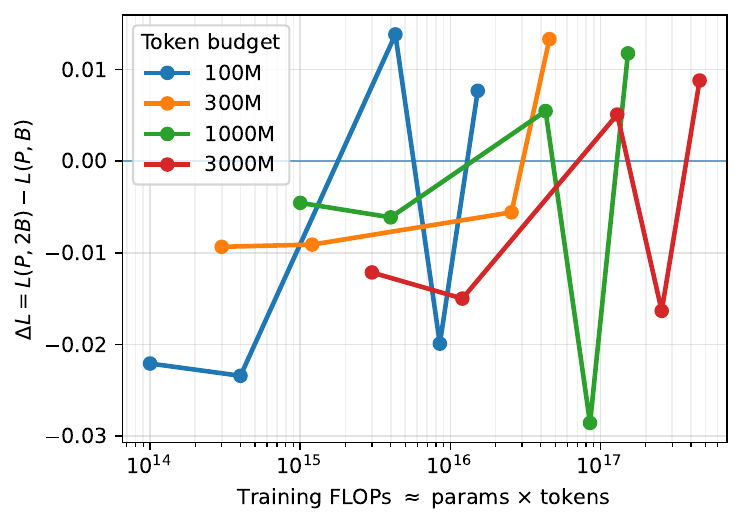}
            \caption{FragSeq}
        \end{subfigure}
        \begin{subfigure}[]{0.48\linewidth}
            \centering
            \includegraphics[width=\linewidth]{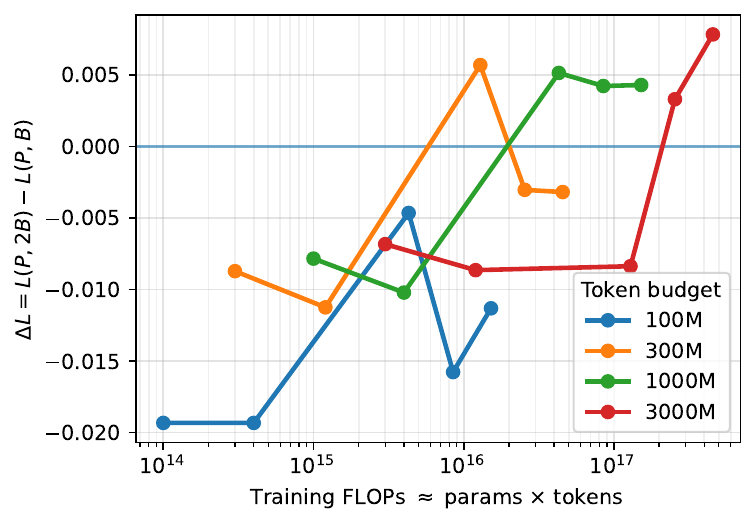}
            \caption{SAFE}
        \end{subfigure}
        \begin{subfigure}[]{0.48\linewidth}
            \centering
            \includegraphics[width=\linewidth]{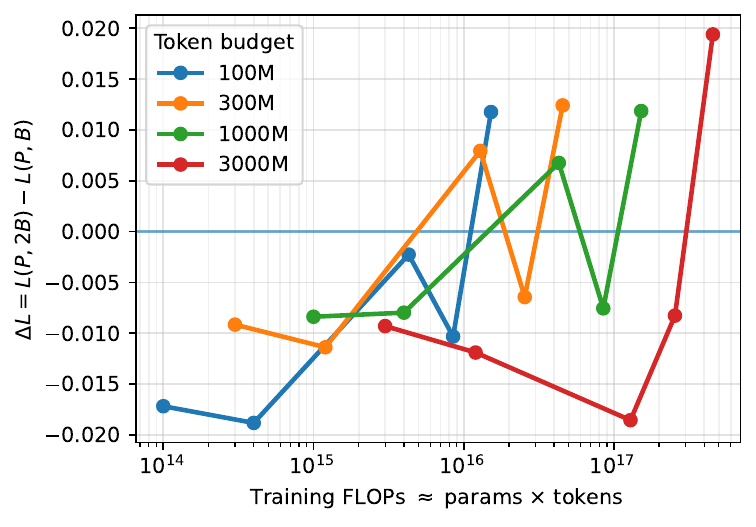}
            \caption{SMILE}
        \end{subfigure}    
    \caption{Longer training on a fixed corpus with a small model. Each panel corresponds to one molecular representation. A $P{=}1$M model is trained under each dataset token budget $B \in \{100\text{M},300\text{M},1\text{B},3\text{B}\}$. Each point is the end-of-run validation loss from an independent from-scratch run that consumes $D=eB$ tokens at epoch count $e \in \{1,\dots,5\}$, and the computation is summarized by $C \propto P D$. Curves connect runs with the same $B$ to show the trend under repeated passes over the same corpus.}
    \label{fig：delta}
  \end{center}
\end{figure*}

To validate the bivariate power-law model, we visualize the validation loss as a function of model size ($P$) and data size ($D$) in Figure \ref{figure: valid_loss_heatmap}. The heatmaps for all five representations show smooth, well-behaved loss landscapes. The consistent color gradients confirm that performance scales predictably, decreasing as either model size or data volume increases. The models achieved a high quality of fit, indicating no significant systematic deviation from the power-law assumption. This provides a strong foundation for our subsequent compute-optimal analysis. These results also highlight that the fragment-based representations (FragSeq and FragLink) converge to a significantly lower irreducible loss ($L_{\infty}$), suggesting a superior theoretical performance limit.

\begin{figure*}[]
  \vskip 0.2in
    \begin{center}
        \begin{subfigure}[]{0.32\linewidth}
            \centering
            \includegraphics[width=\linewidth]{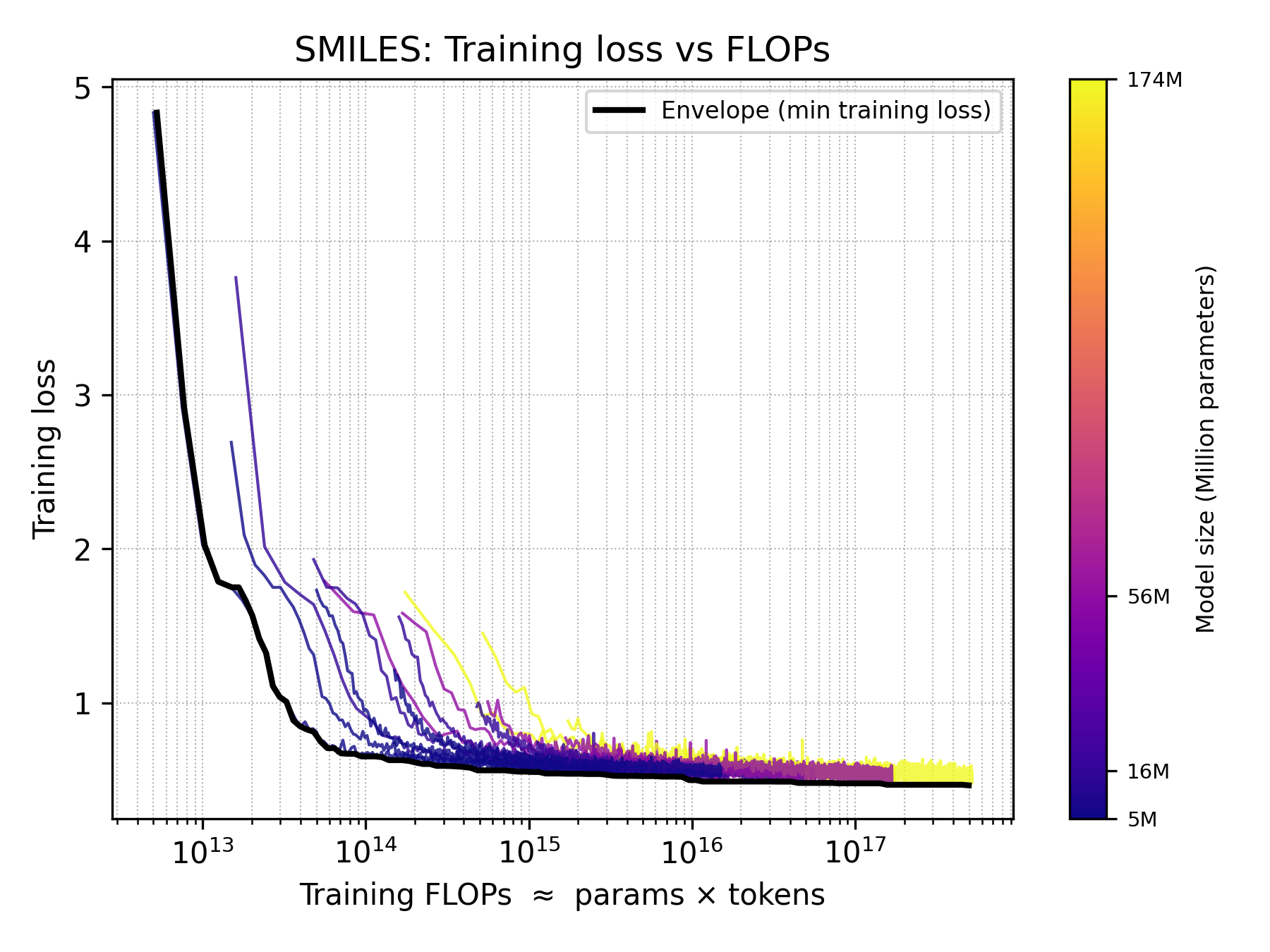}
            \caption{SMILES}
        \end{subfigure}
        \begin{subfigure}[]{0.32\linewidth}
            \centering
            \includegraphics[width=\linewidth]{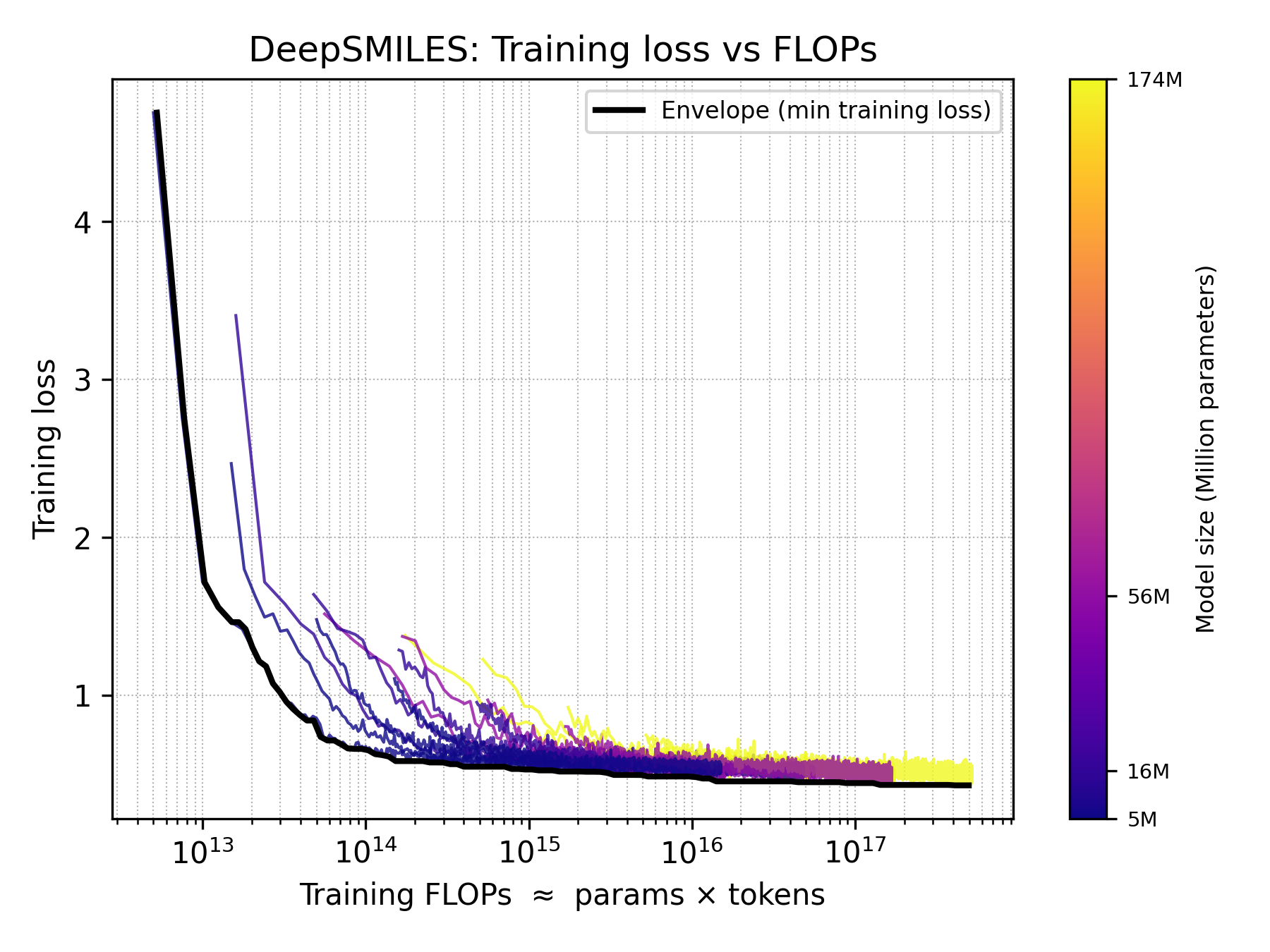}
            \caption{DeepSMILES}
        \end{subfigure}
        \begin{subfigure}[]{0.32\linewidth}
            \centering
            \includegraphics[width=\linewidth]{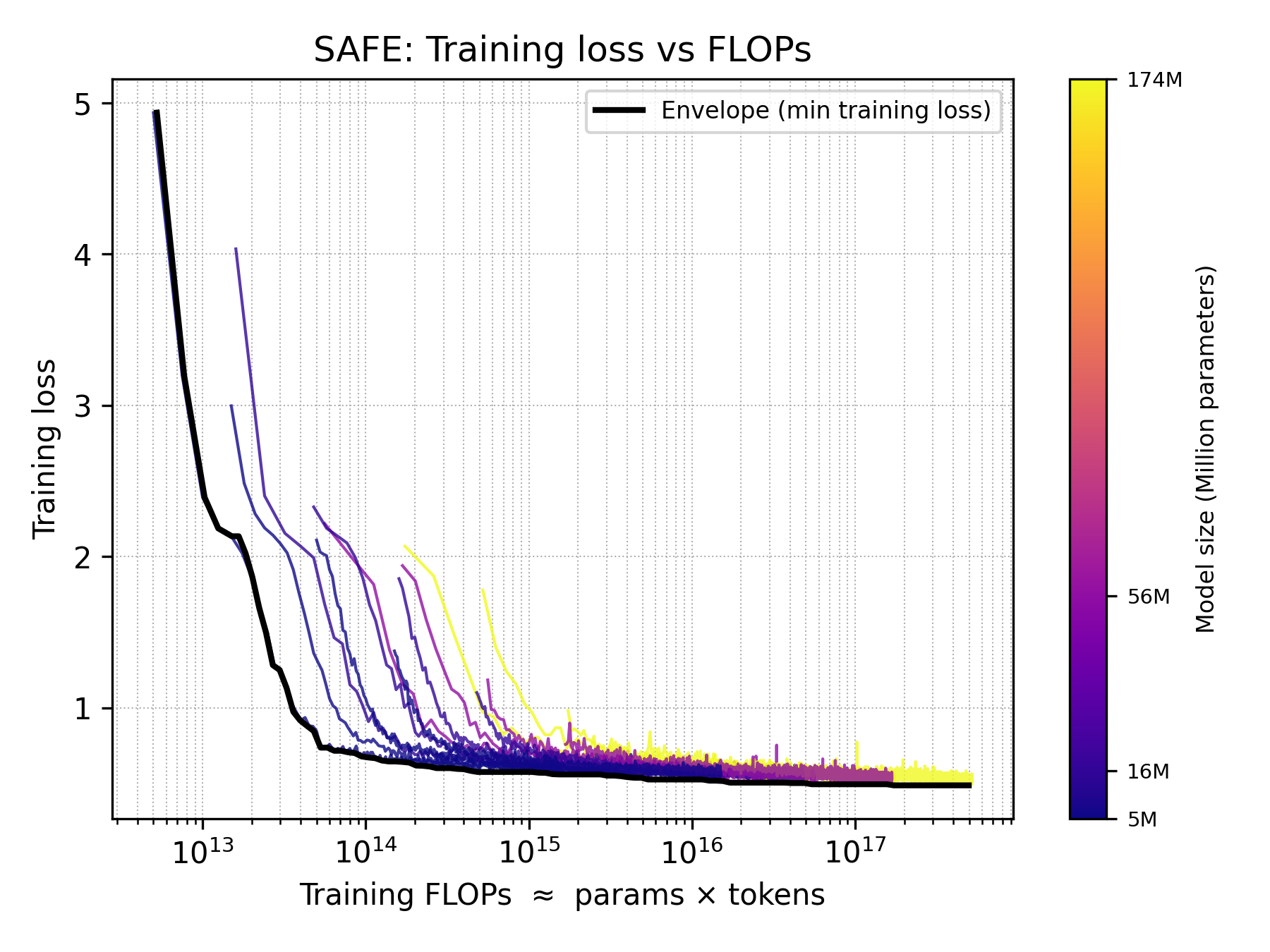}
            \caption{SAFE}
        \end{subfigure}
        \begin{subfigure}[]{0.32\linewidth}
            \centering
            \includegraphics[width=\linewidth]{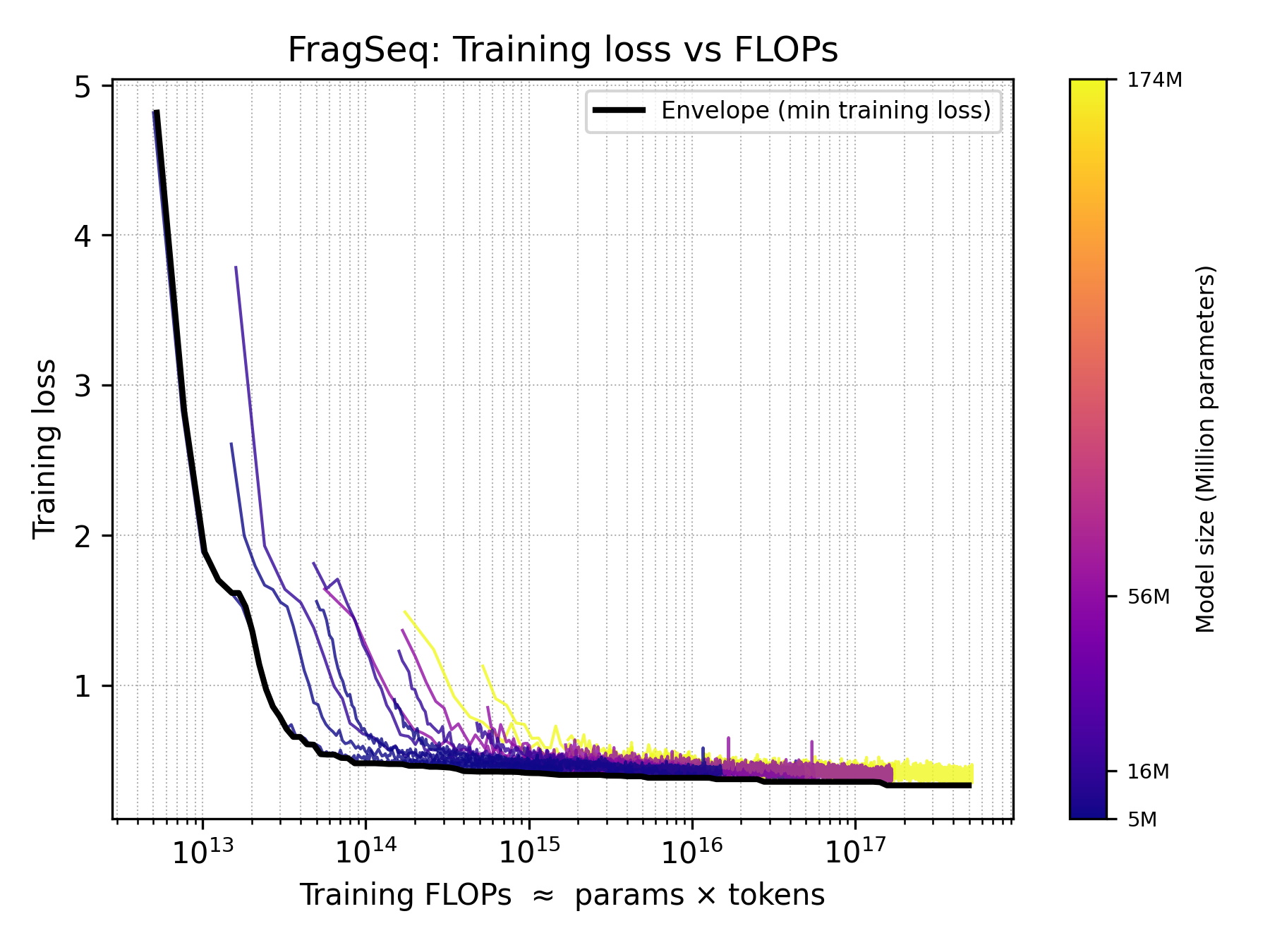}
            \caption{FragSeq}
        \end{subfigure}
        \begin{subfigure}[]{0.32\linewidth}
            \centering
            \includegraphics[width=\linewidth]{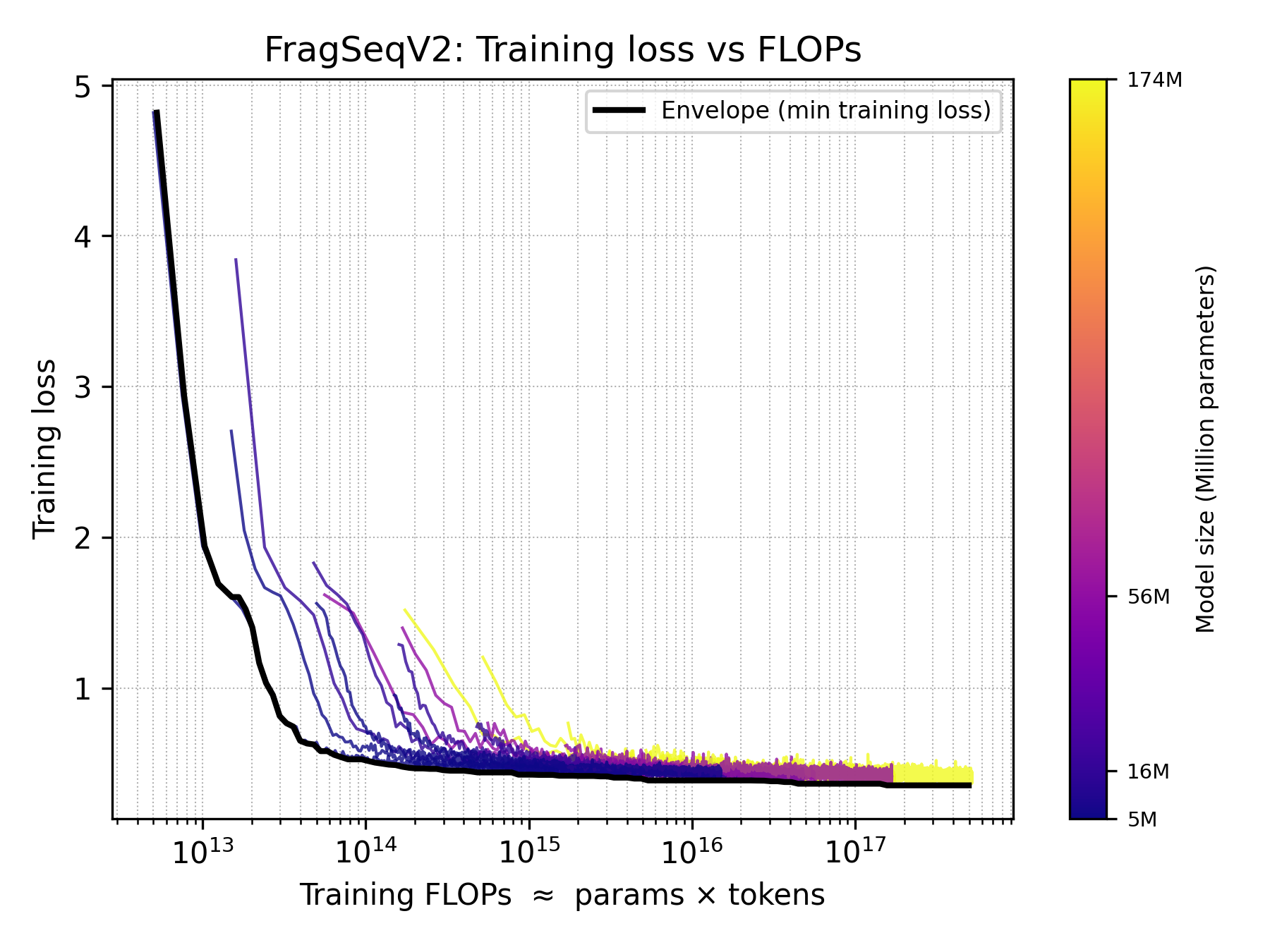}
            \caption{FragLink}
        \end{subfigure}    
    \caption{Training loss versus training FLOPs. The bold black line represents the minimum loss envelope.}
    \label{figure: train_loss_curves}
  \end{center}
\end{figure*}

Figure \ref{figure: train_loss_curves} illustrates the training loss as a function of the total computational budget (FLOPs). The thin colored lines represent the training trajectories of individual models, while the bold black line is the minimum loss envelope, which traces the best performance achievable at any given compute budget. All representations exhibit a rapid initial drop in loss, which gradually flattens after approximately $10^{16}$ FLOPs, demonstrating the law of diminishing returns. The fragment-based representations (FragSeq and FragLink) show faster convergence, with their loss curves descending more steeply. This indicates a more efficient use of computational resources. For a given representation, larger models (brighter colors) start at a higher initial loss but eventually converge to the same loss envelope as smaller models, albeit at a higher computational cost. The optimal performance at any given FLOPs budget is achieved by a specific model size, as captured by the black envelope.

Figure \ref{fig:isoflop_isoloss} provides an intuitive validation of the existence of a compute-optimal model size from a different perspective. Each U-shaped curve, known as an IsoFLOP curve, represents a cross-section of the loss landscape at a constant computational budget. The x-axis is model size, and the y-axis is validation loss.  The distinct U-shape confirms that for any fixed budget, there exists a unique optimal model size that minimizes loss. Being ``under-parameterized" (left side of the curve) or ``under-datasized" (right side of the curve) both lead to suboptimal performance. At any given compute budget, the IsoFLOP curves for FragSeq and FragLink consistently lie at a lower loss level than those for the SMILES family, demonstrating their systematic performance advantage.

\section{Downstream Task Benchmark Comparison}
\label{sec: appendix E}

\subsection{State-of-the-Art Benchmark Comparison}
This appendix provides a detailed comparison of our fine-tuned models' performance against existing state-of-the-art (SOTA) methods on the MoleculeNet benchmarks. Table \ref{table: MoleculeNet-cla} presents the results for the six classification tasks, measured in average AUC-ROC. Table \ref{table: MoleculeNet-reg} presents the results for the three regression tasks, measured in RMSE. Our models, denoted as ``Ours(Representation)", demonstrate competitive or superior performance across a wide range of tasks, validating the effectiveness of our pre-training and fine-tuning methodology.

\begin{table*}[]
  \caption{Molecular property prediction tasks on MoleculeNet benchmarks. We indicate the best performance in \textbf{bold}, while the second and third performance are indicated by \underline{underling}.}
  \label{table: MoleculeNet-cla}
  \begin{center}
    \begin{small}
      \begin{sc}
        \begin{tabular}{lccccccr}
          \toprule
          Dataset            & BACE  & HIV    & BBBP  & SIDER  & Tox21 & ClinTox  \\
          \midrule
          Tasks              & 1     & 1      & 1     & 27     & 12    & 2        \\
          \midrule
          Metrics            & \multicolumn{6}{c}{Average AUC-ROC($\uparrow$)}    \\
          \midrule
          GCN \cite{kipf2016semi}                        & 71.6  & 74.0   & 71.9  & 53.6   & 70.9  & 62.5     \\       
          GIN \cite{xu2018powerful}                      & 70.1  & 75.3   & 65.8  & 57.3   & 74.0  & 58.0     \\
          SchNet \cite{schutt2018schnet}                 & 76.6  & 70.2   & 84.8  & 53.9   & 77.2  & 71.5     \\
          MGCN \cite{lu2019molecular}                    & 73.4  & 73.8   & 85.0  & 55.2   & 70.7  & 63.4     \\
          D-MPNN \cite{yang2019analyzing}                & 85.3  & 75.0   & 71.2  & 63.2   & 68.9  & 90.5     \\
          AttrMask \cite{hu2019strategies}               & 79.7  & 76.8   & 79.7  & 61.2   & 74.8  & 65.0     \\
          GPT-GNN \cite{hu2020gpt}                       & 85.9  & 80.2   & 70.8  & 65.2   & 78.7  & 78.9     \\
          MolCLR-GCN \cite{wang2022molecular}            & 78.8  & 77.8   & 73.8  & 66.9   & 74.7  & 86.7     \\
          MolCLR-GIN \cite{wang2022molecular}            & 89.0  & 80.6   & 73.6  & 68.0   & 79.8  & 93.2     \\
          SimSGT \cite{liu2023rethinking}                & 84.3  & 78.0   & 72.2  & 61.7   & 76.8  & 85.7     \\
          \midrule
          MolBERT \cite{fabian2020molecular}             & 86.6  & 78.3   & 76.2  & -      & -     & -        \\
          ChemBerta \cite{chithrananda2020chemberta}     & 79.9  & 62.2   & 64.3  & -      & -     & 73.3     \\
          ChemBerta2\cite{ahmad2022chemberta}            & 85.1  & -      & 71.9  & -      & -     & 90.7     \\
          MolFormer-XL \cite{ross2022large}              & 88.2  & \underline{82.2}   & 93.7  & \underline{69.0}   & \textbf{84.7}  & 94.8    \\
          KV-PLM \cite{zeng2022deep}                     & -     & 74.0   & 74.6  & 61.5   & 72.7  & -        \\
          Galactica \cite{taylor2022galactica}           & 61.7  & 74.5   & 66.1  & 63.2   & 68.9  & 82.6     \\
          MoMu \cite{su2022molecular}                    & 77.1  & 76.2   & 70.5  & 60.5   & 75.6  & 79.9     \\
          Uni-Mol \cite{zhou2023uni}                     & 85.7  & 80.8   & 72.9  & 65.9   & 79.9  & 91.9     \\
          SELFormer \cite{yuksel2023selformer}           & 83.2  & 68.1   & 90.2  & \textbf{74.5}   & 65.3  & -        \\
          SELF-BART \cite{priyadarsini2024self}          & \underline{89.3}  & \underline{83.0}   & 95.2  & 65.0   & 76.5  & \underline{96.9}    \\
          MolXPT \cite{liu2023molxpt}                    & 88.4  & 78.1   & 80.0  & \underline{71.7}   & 77.1  & 95.3     \\
          MoleculeSTM-SMILES \cite{liu2023multi}         & 82.0  & 77.0   & 70.8  & 63.7   & 75.7  & 86.6     \\
          MolFM \cite{luo2023molfm}                      & 83.9  & 78.8   & 72.9  & 64.2   & 77.2  & 79.7     \\
          MolCA-SMILES \cite{liu2023molca}               & 79.3  & -      & 70.8  & 61.1   & 76.0  & 89.0     \\
          Atomas \cite{zhang2024atomas}                  & 83.1  & 80.6   & 73.7  & 64.4   & 77.9  & 93.2     \\
          GPT-MolBERTa \cite{balaji2023gpt}              & 73.4  & 75.5   & 74.1  & 58.5   & 65.9  & 49.7     \\
          \midrule
          Ours(DeepSMILES)   & 88.2  & 81.6   & \textbf{97.8}  & 66.0   & \underline{83.6}  & \textbf{99.8}     \\
          Ours(FragLink)     & \textbf{89.7}  & 79.2   & 95.9  & 68.0   & 82.8  & 91.9     \\
          Ours(FragSeq)      & 87.2  & 79.7   & \underline{96.3}  & 68.8   & \underline{83.7}  & 95.4     \\
          Ours(SAFE)         & 82.8  & \textbf{83.3}   & 95.6  & 66.7   & 81.3  & 93.2     \\
          Ours(SMILES)       & \underline{89.4}  & 81.9   & \underline{97.6}  & 65.8   & 82.8  & \textbf{99.8}     \\
          \bottomrule
        \end{tabular}
      \end{sc}
    \end{small}
  \end{center}
  \vskip -0.1in
\end{table*}

\begin{table*}[]
  \caption{RMSE of molecular property prediction tasks (regression) on MoleculeNet benchmarks. We indicate the best
performance in \textbf{bold}, while the second and third performance are indicated by \underline{underling}.}
  \label{table: MoleculeNet-reg}
  \begin{center}
    \begin{small}
      \begin{sc}
        \begin{tabular}{lcccr}
          \toprule
          Dataset            & ESOL  & FreeSolv & Lipophilicity       \\
          \midrule
          Tasks              & 1     & 1        & 1                   \\
          \midrule
          Metrics            & \multicolumn{3}{c}{RMSE($\downarrow$)} \\
          \midrule
          GCN \cite{kipf2016semi}                     & 1.430 & 2.870    & 0.850               \\                 
          GIN \cite{xu2018powerful}                   & 1.450 & 2.760    & 0.850               \\
          SchNet \cite{schutt2018schnet}              & 1.050 & 3.220    & 0.910               \\
          MGCN \cite{lu2019molecular}                 & 1.270 & 3.350    & 1.110               \\
          D-MPNN \cite{yang2019analyzing}             & 0.980 & 2.180    & 0.650               \\
          AttrMask \cite{hu2019strategies}            & 1.112 & -        & 0.730               \\
          GPT-GNN \cite{hu2020gpt}                    & 1.220 & 2.830    & 0.740               \\
          MolCLR-GCN \cite{wang2022molecular}         & 1.160 & 2.390    & 0.780               \\
          MolCLR-GIN \cite{wang2022molecular}         & 1.110 & 2.200    & 0.650               \\
          SimSGT \cite{liu2023rethinking}             & 0.917 & -        & 0.695               \\
          \midrule
          MolBERT\cite{fabian2020molecular}           & 0.531 & \underline{0.948}    & \underline{0.561}               \\
          ChemBerta2\cite{ahmad2022chemberta}         & -     & -        & 0.986               \\
          MolFormer-XL\cite{ross2022large}            & \textbf{0.279} & \textbf{0.231}    & \textbf{0.529}               \\
          SELFormer\cite{yuksel2023selformer}         & 0.682 & 2.797    & 0.735               \\
          GPT-MolBERTa\cite{balaji2023gpt}            & 0.477 & \underline{0.896}    & 0.758               \\
          SELF-BART\cite{priyadarsini2024self}        & 0.454 & 1.397    & 0.771               \\
          \midrule
          Ours(SMILES)       & \underline{0.368} & 1.274    & 0.602               \\
          Ours(DeepSMILES)   & \underline{0.362} & 1.210    & 0.627               \\
          Ours(SAFE)         & 0.412 & 1.113    & 0.652               \\
          Ours(FragSeq)      & 0.378 & 1.277    & 0.608               \\
          Ours(FragLink)     & 0.383 & 1.095    & \underline{0.593}               \\
          \bottomrule
        \end{tabular}
      \end{sc}
    \end{small}
  \end{center}
  \vskip -0.1in
\end{table*}

\subsection{Detailed Performance Analysis Across Pre-training Scales}

\begin{figure*}[]
  \vskip 0.2in
    \begin{center}
      \begin{subfigure}[]{0.4\linewidth}
        \centering
        \includegraphics[width=\linewidth]{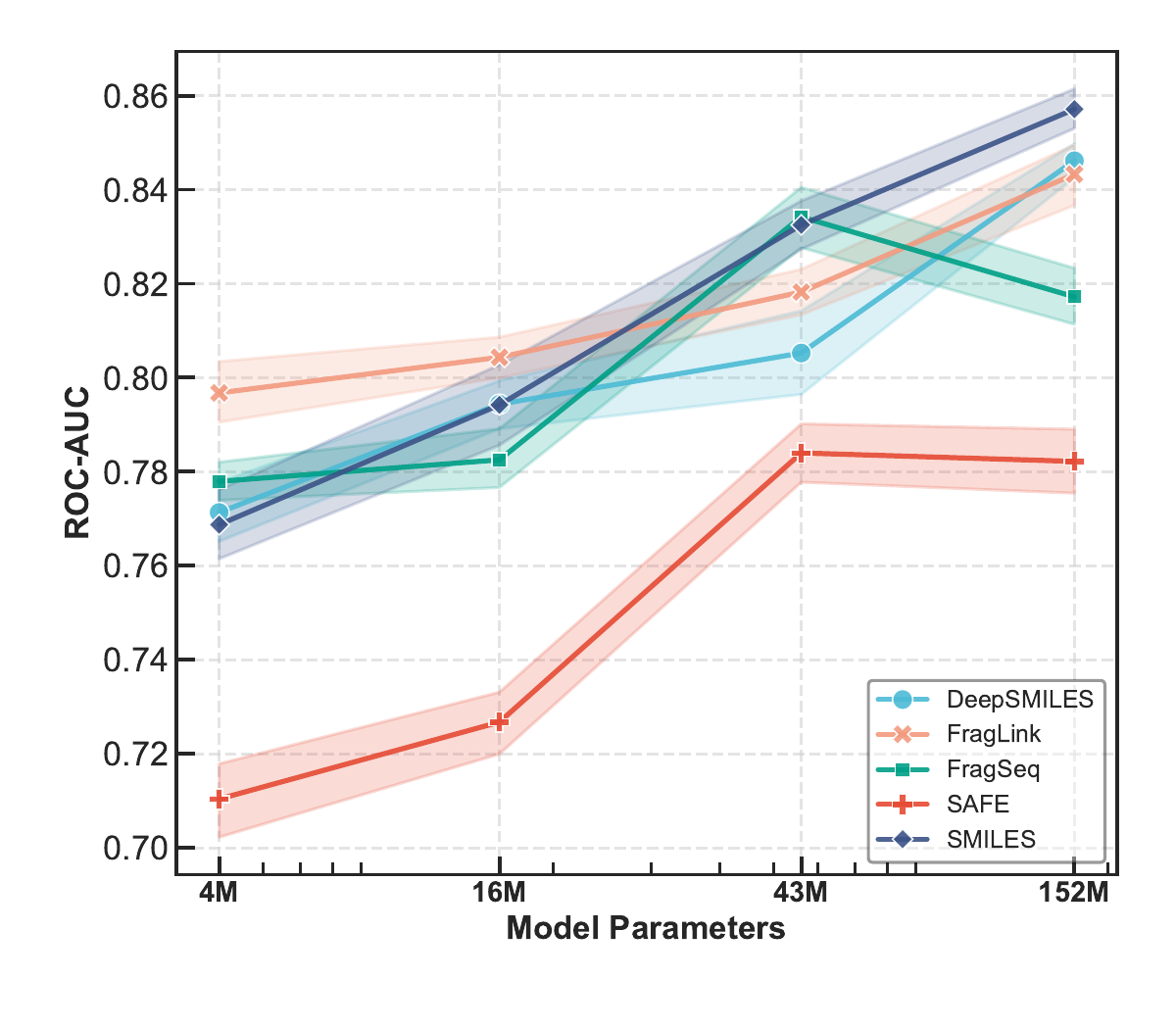}
        \caption{BACE}
      \end{subfigure}
      \begin{subfigure}[]{0.4\linewidth}
        \centering
        \includegraphics[width=\linewidth]{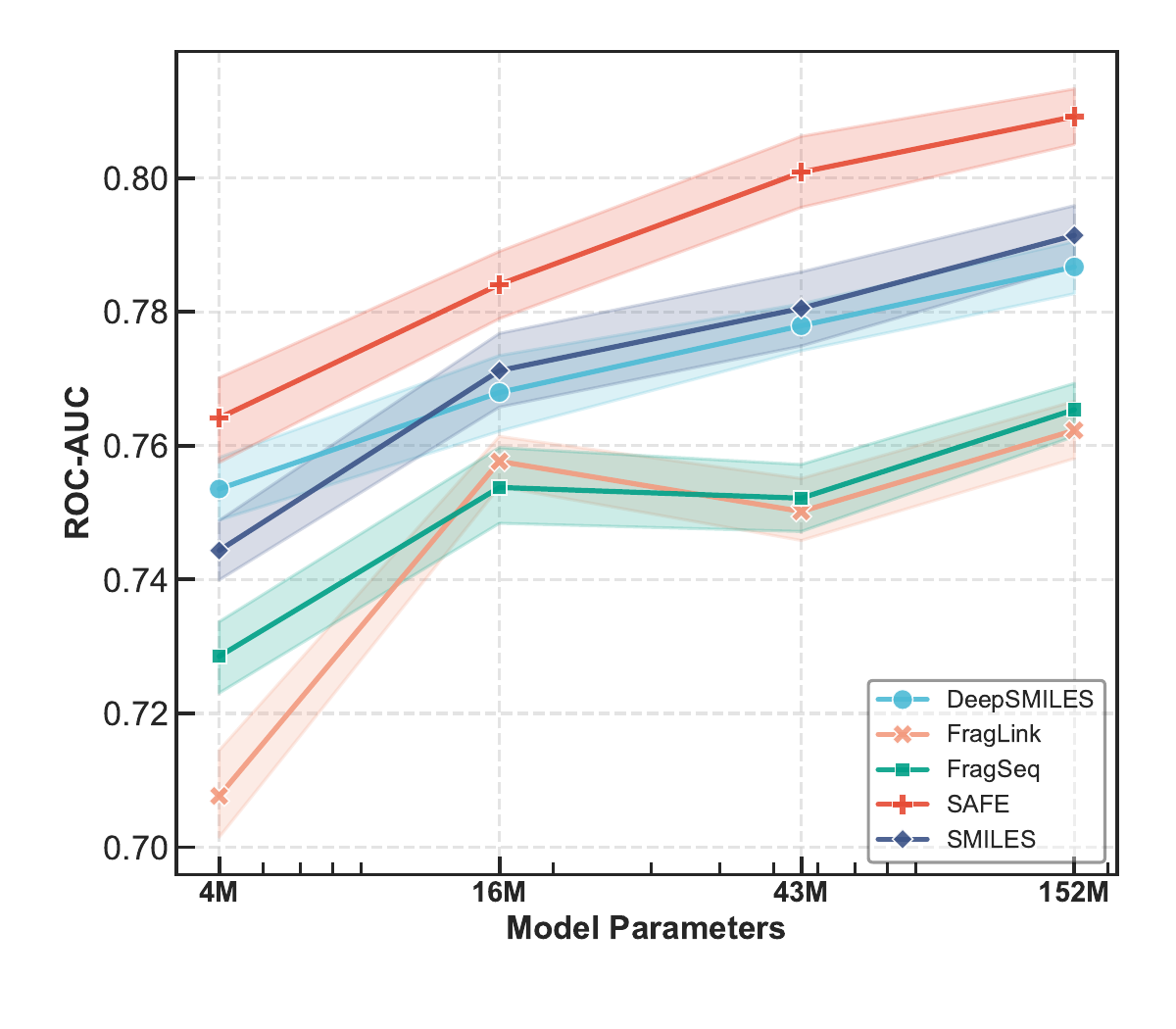}
        \caption{HIV}
      \end{subfigure}
    \caption{Performance on Biochemistry MoleculeNet benchmarks: (a) BACE, and (b) HIV. ROC-AUC is used for classification tasks (higher is better).}
    \label{figure: Biochemistry MoleculeNet Scaling}
  \end{center}
\end{figure*}

\begin{figure*}[]
  \vskip 0.2in
    \begin{center}
      \begin{subfigure}[]{0.4\linewidth}
        \centering
        \includegraphics[width=\linewidth]{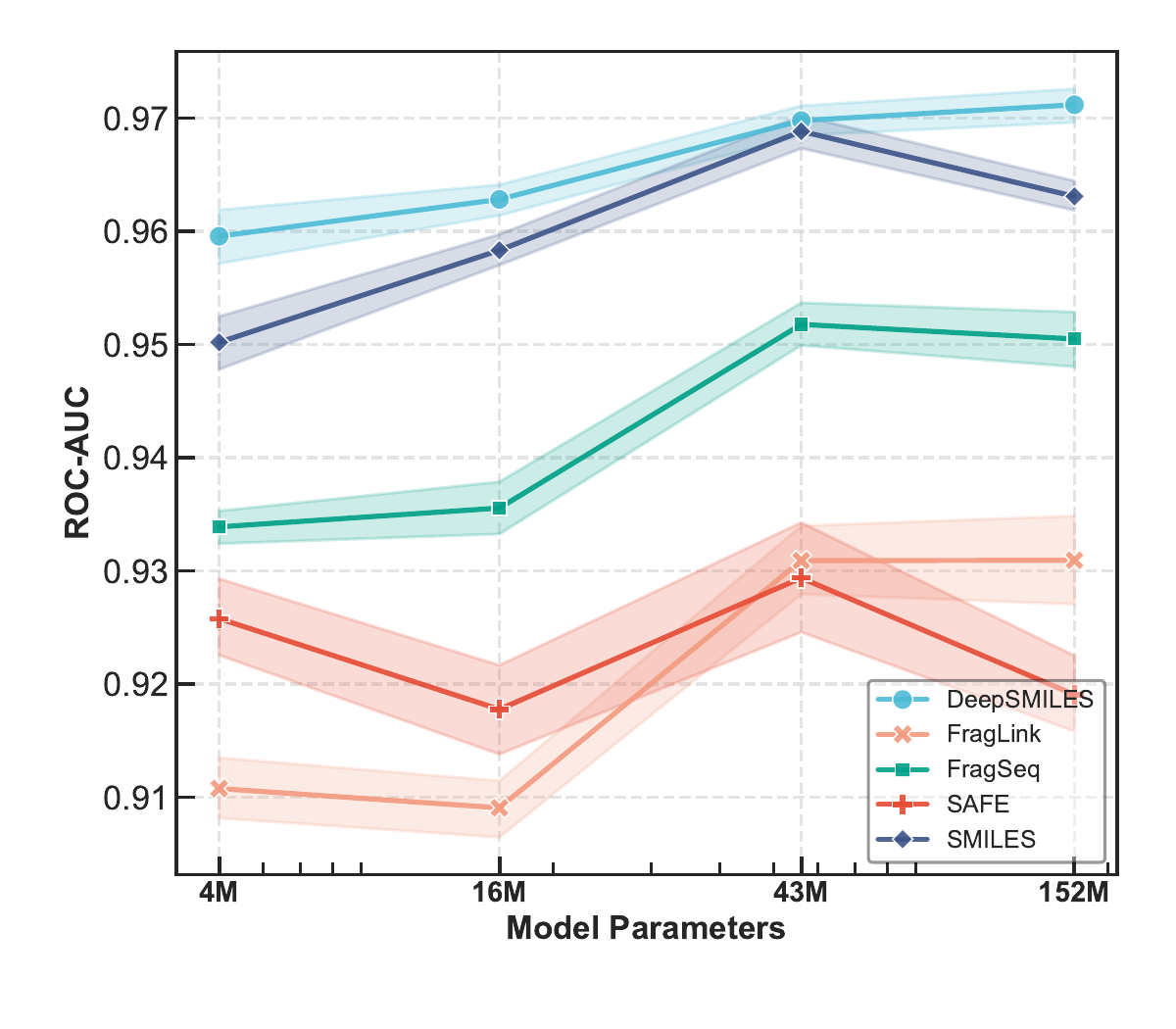}
        \caption{BBBP}
      \end{subfigure}
      \begin{subfigure}[]{0.4\linewidth}
        \centering
        \includegraphics[width=\linewidth]{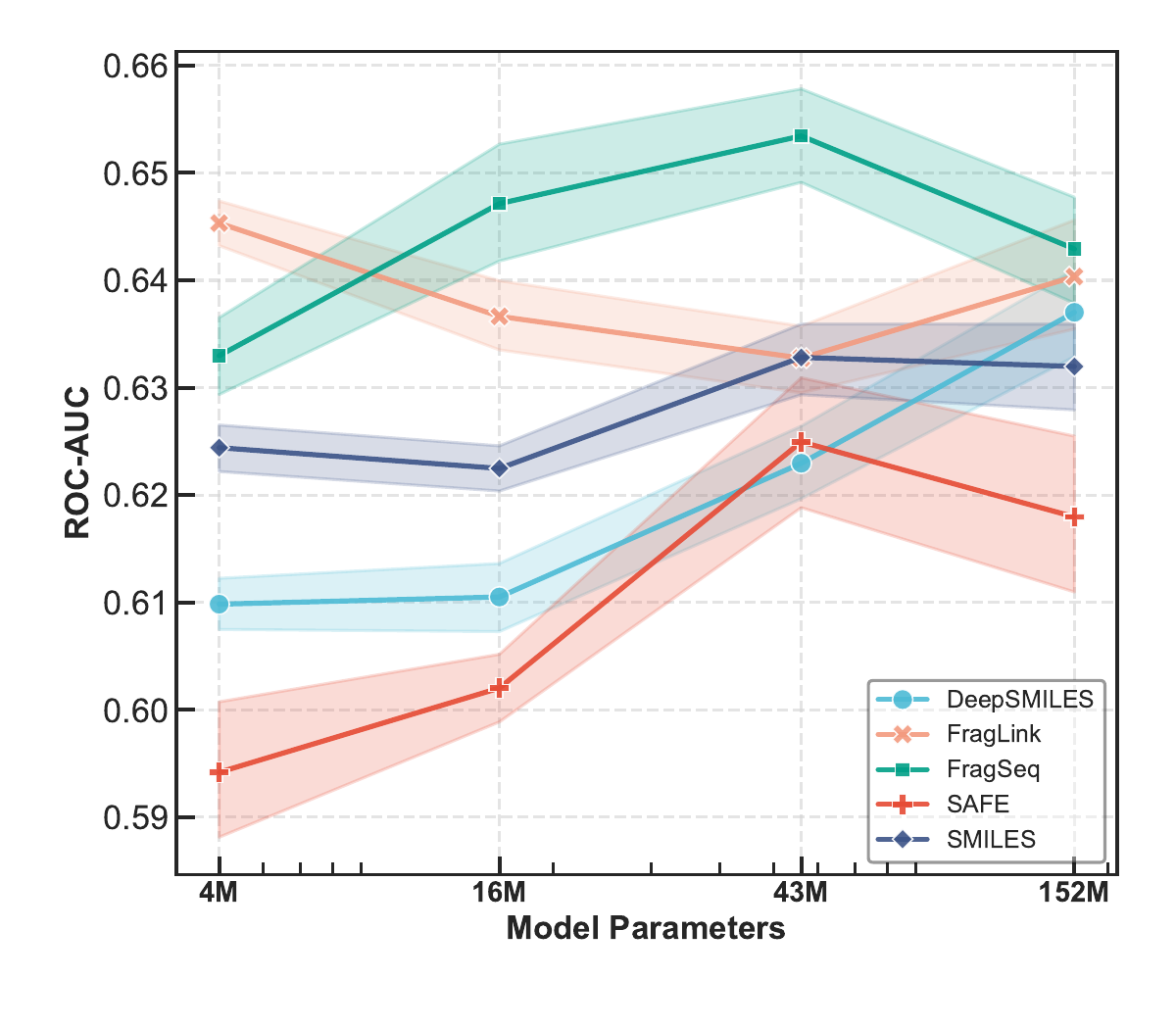}
        \caption{SIDER}
      \end{subfigure}
      \begin{subfigure}[]{0.4\linewidth}
        \centering
        \includegraphics[width=\linewidth]{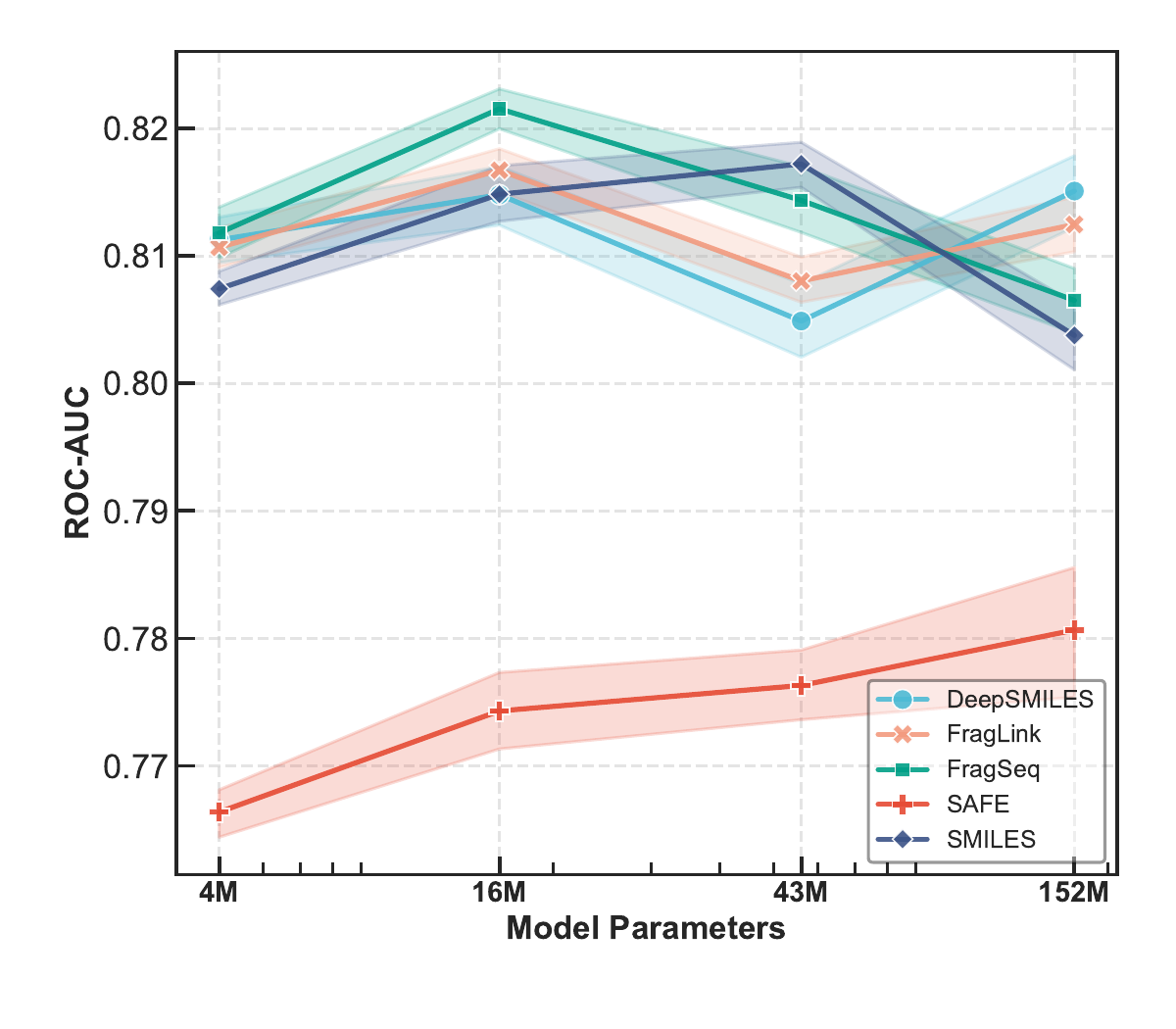}
        \caption{Tox21}
      \end{subfigure}
      \begin{subfigure}[]{0.4\linewidth}
        \centering
        \includegraphics[width=\linewidth]{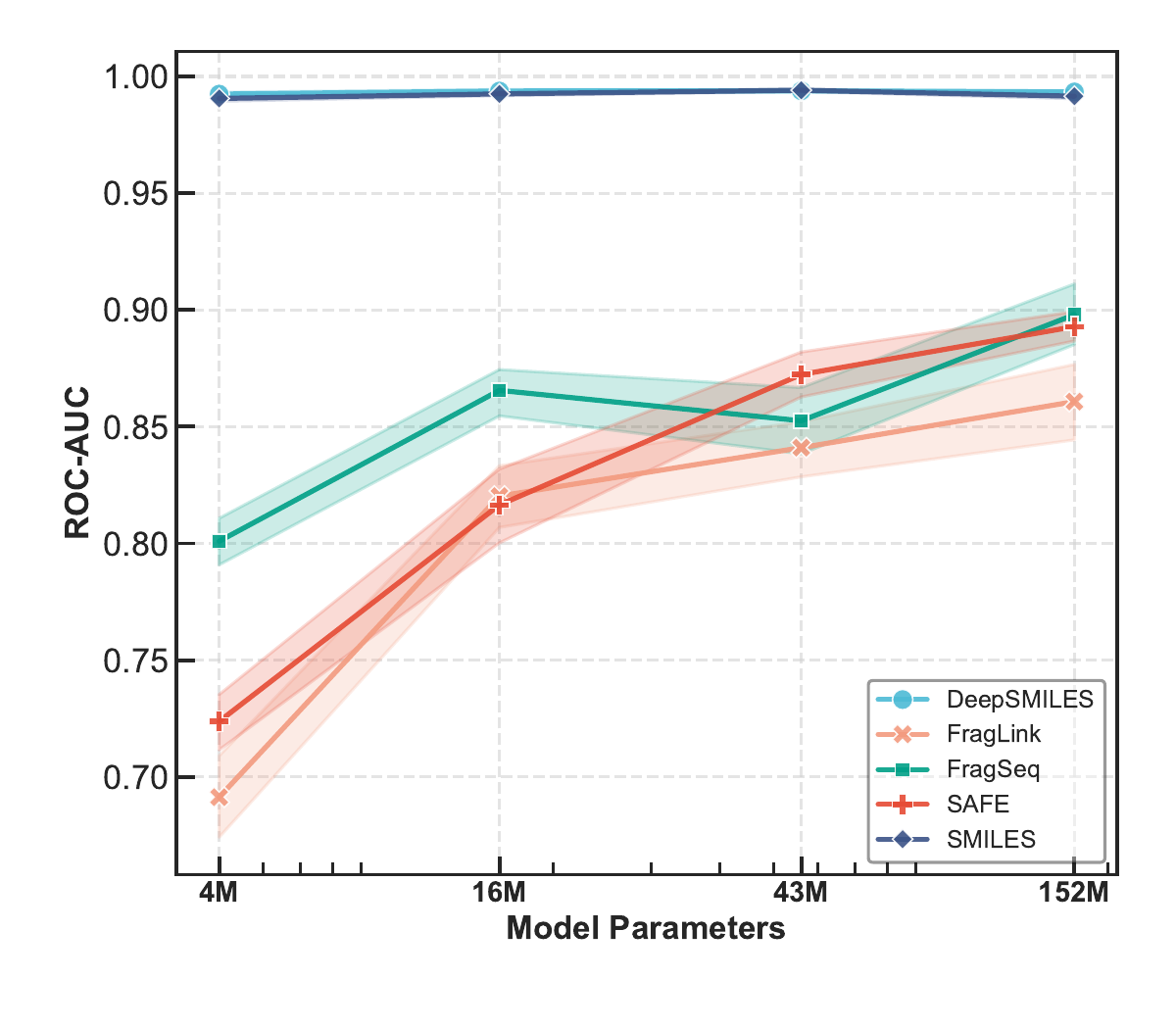}
        \caption{ClinTox}
      \end{subfigure}
    \caption{Performance on Physiology MoleculeNet benchmarks: (a) BBBP, (b) SIDER, (c) Tox21, and (d) ClinTox. ROC-AUC is used for classification tasks (higher is better).}
    \label{figure: Physiology MoleculeNet Scaling}
  \end{center}
\end{figure*}

\begin{figure*}[]
  \vskip 0.2in
    \begin{center}
      \begin{subfigure}[]{0.33\linewidth}
        \centering
        \includegraphics[width=\linewidth]{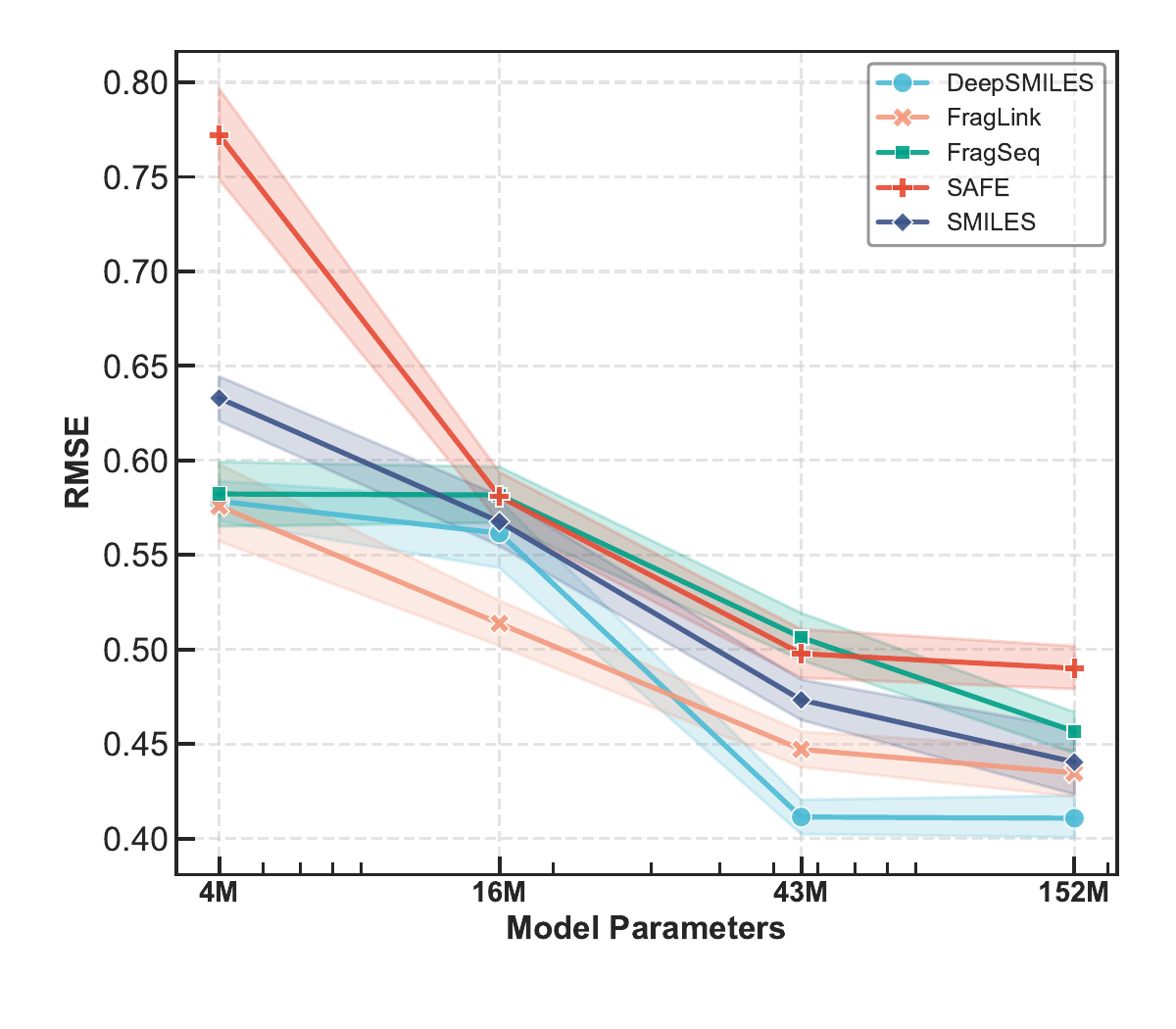}
        \caption{ESOL}
      \end{subfigure}
      \begin{subfigure}[]{0.33\linewidth}
        \centering
        \includegraphics[width=\linewidth]{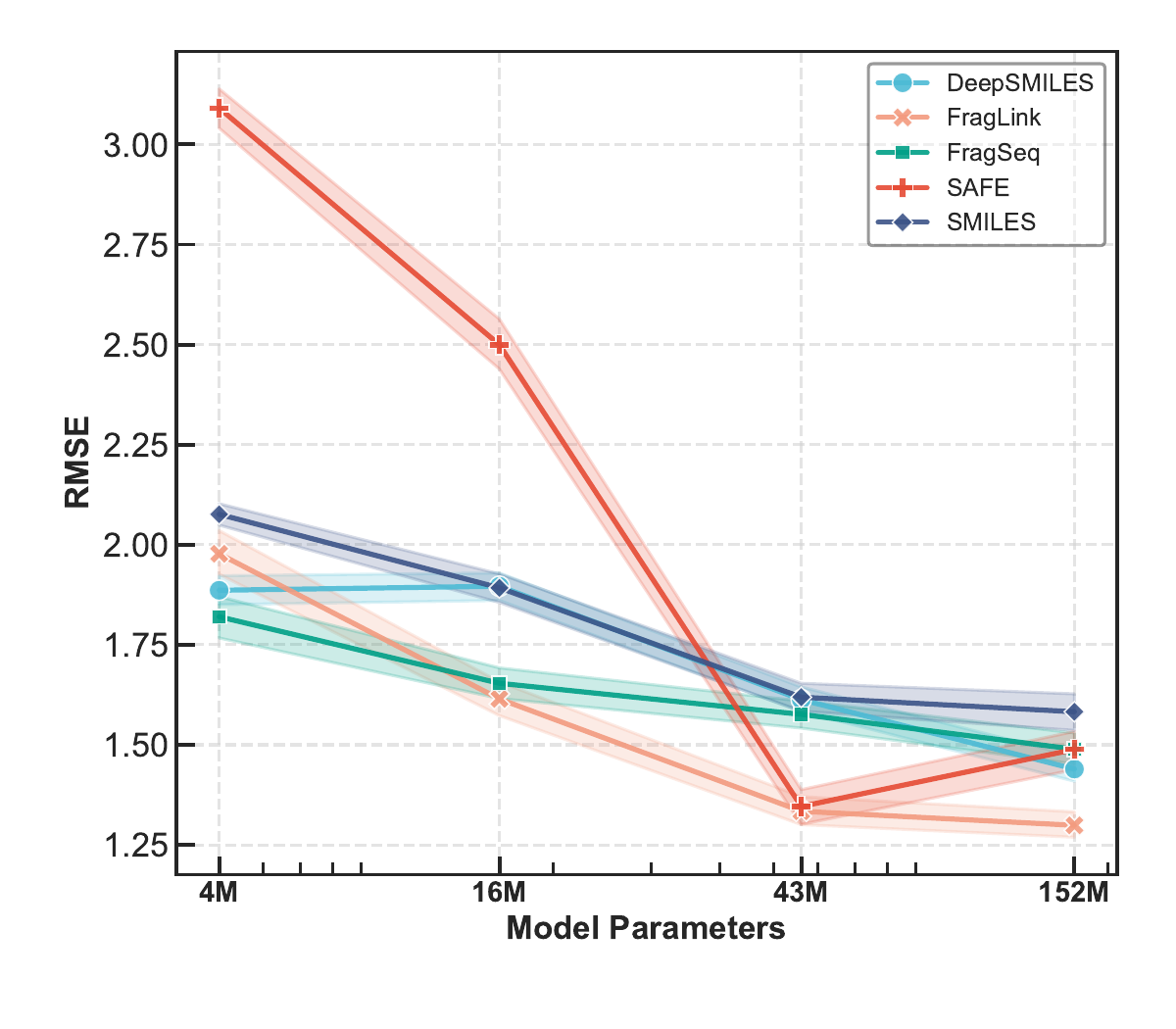}
        \caption{FreeSolv}
      \end{subfigure}
      \begin{subfigure}[]{0.33\linewidth}
        \centering
        \includegraphics[width=\linewidth]{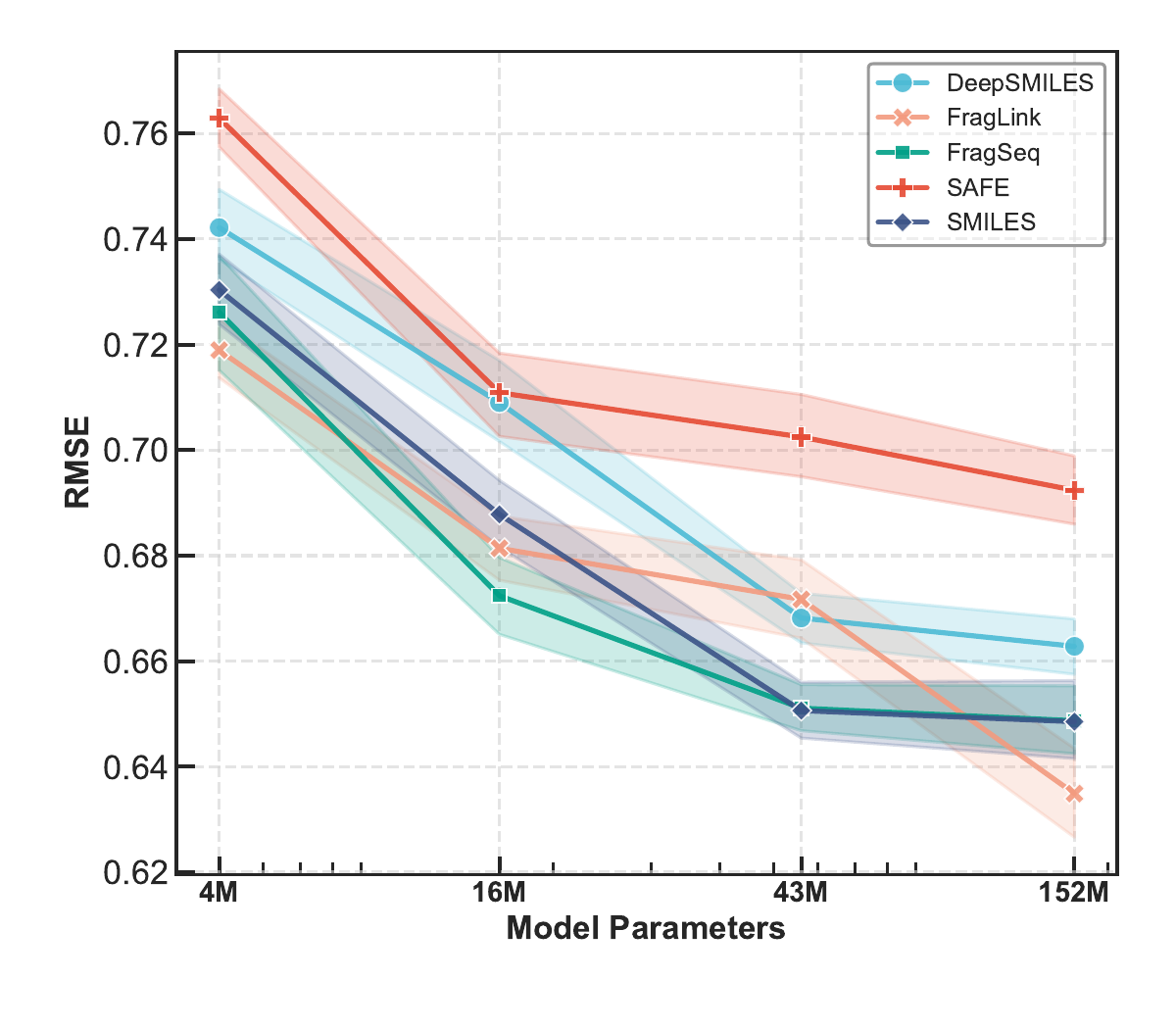}
        \caption{Lipophilicity}
      \end{subfigure}
    \caption{Performance on Biophysics MoleculeNet benchmarks: (a) ESOL, (b) FreeSolv, and (c) Lipophilicity. RMSE is used for classification tasks (lower is better).}
    \label{figure: Biophysics MoleculeNet Scaling}
  \end{center}
\end{figure*}

\begin{figure*}[]
  \vskip 0.2in
  \begin{center}
    \centerline{
        \includegraphics[width=0.98\linewidth]{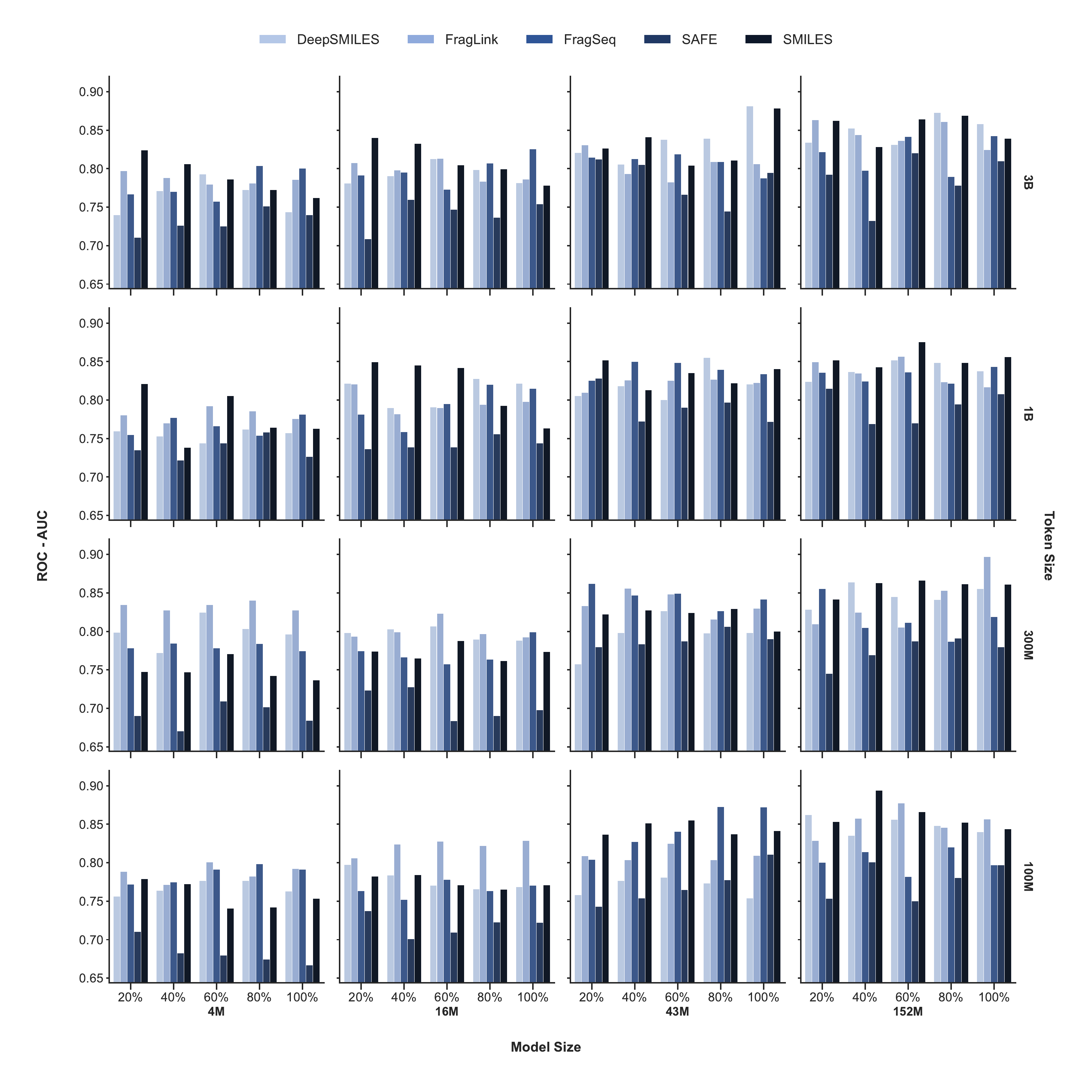}
    }
    \caption{Performance on BACE benchmark.}
    \label{figure: MoleculeNet BACE Bars}
  \end{center}
\end{figure*}

\begin{figure*}[]
  \vskip 0.2in
  \begin{center}
    \centerline{
        \includegraphics[width=0.98\linewidth]{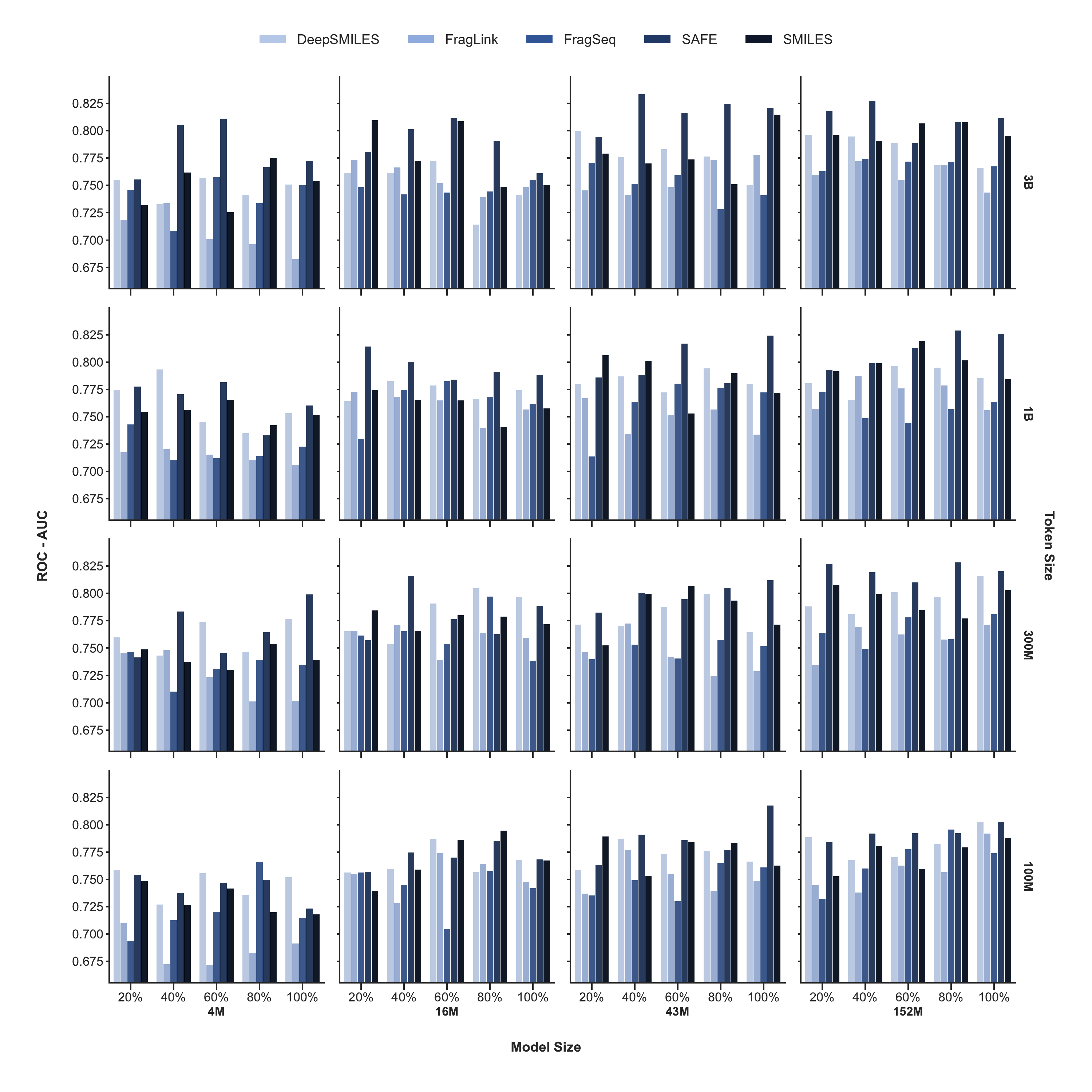}
    }
    \caption{Performance on HIV benchmark.}
    \label{figure: MoleculeNet HIV Bars}
  \end{center}
\end{figure*}

\begin{figure*}[]
  \vskip 0.2in
  \begin{center}
    \centerline{
        \includegraphics[width=0.98\linewidth]{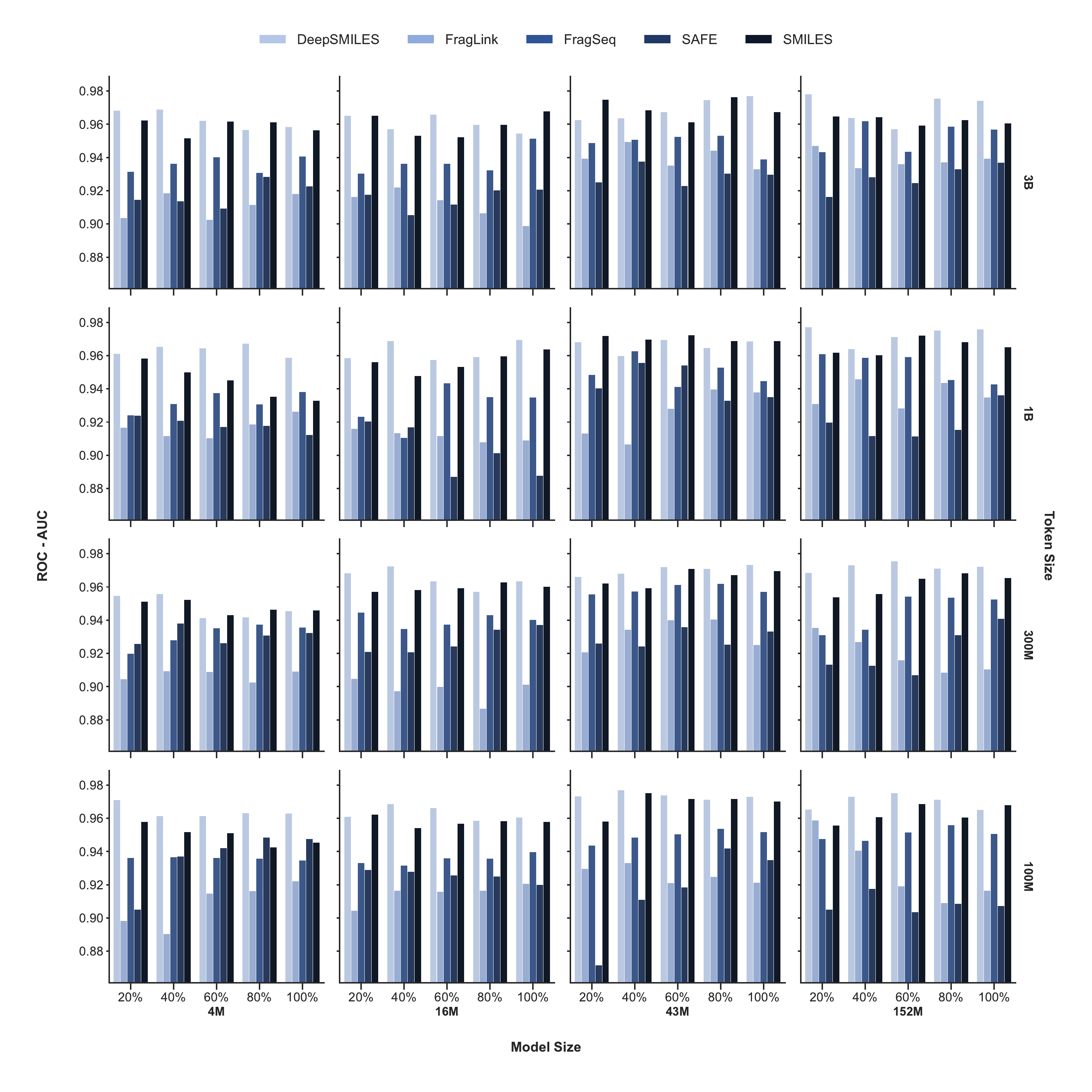}
    }
    \caption{Performance on BBBP benchmark.}
    \label{figure: MoleculeNet BBBP Bars}
  \end{center}
\end{figure*}

\begin{figure*}[]
  \vskip 0.2in
  \begin{center}
    \centerline{
        \includegraphics[width=0.98\linewidth]{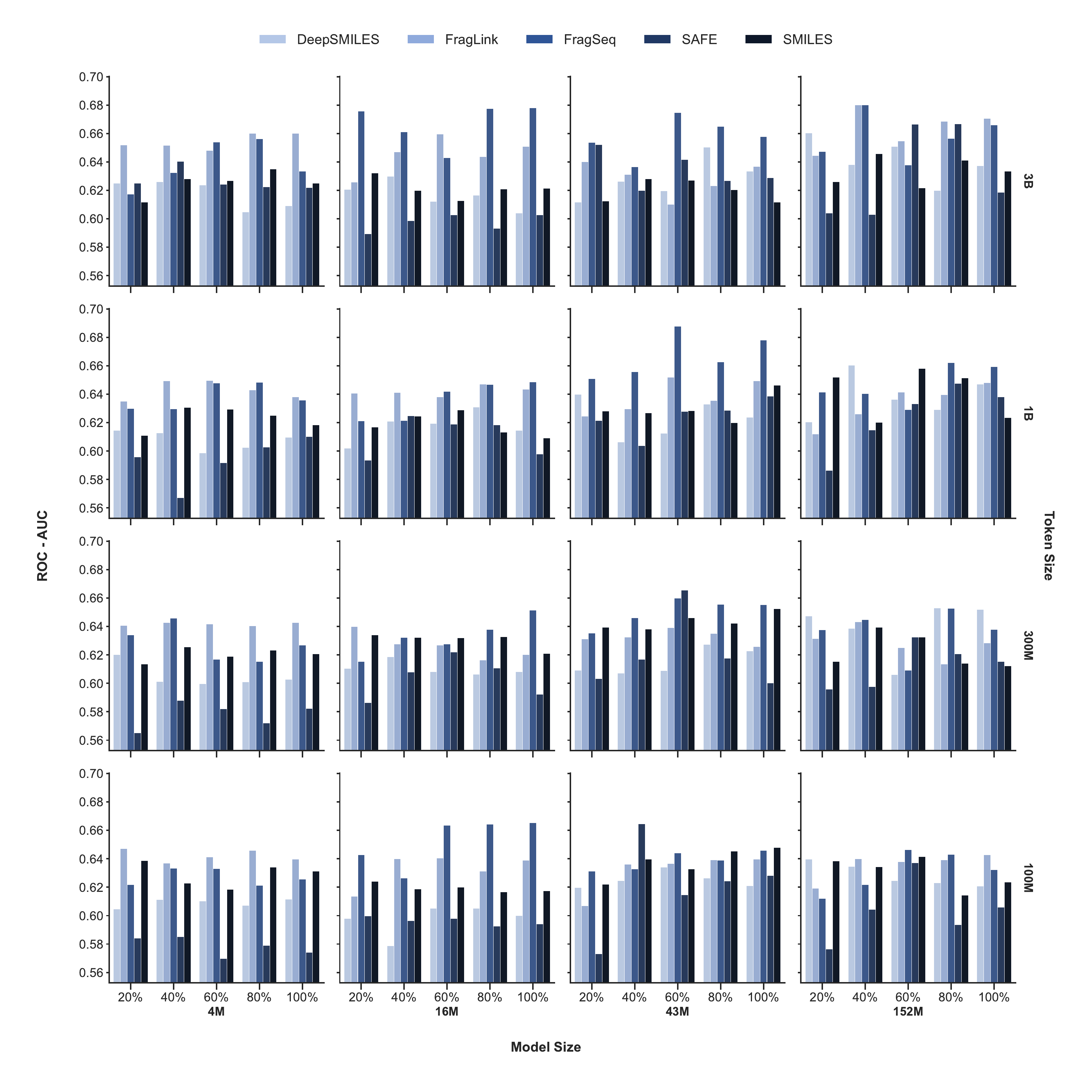}
    }
    \caption{Performance on SIDER benchmark.}
    \label{figure: MoleculeNet SIDER Bars}
  \end{center}
\end{figure*}

\begin{figure*}[]
  \vskip 0.2in
  \begin{center}
    \centerline{
        \includegraphics[width=0.98\linewidth]{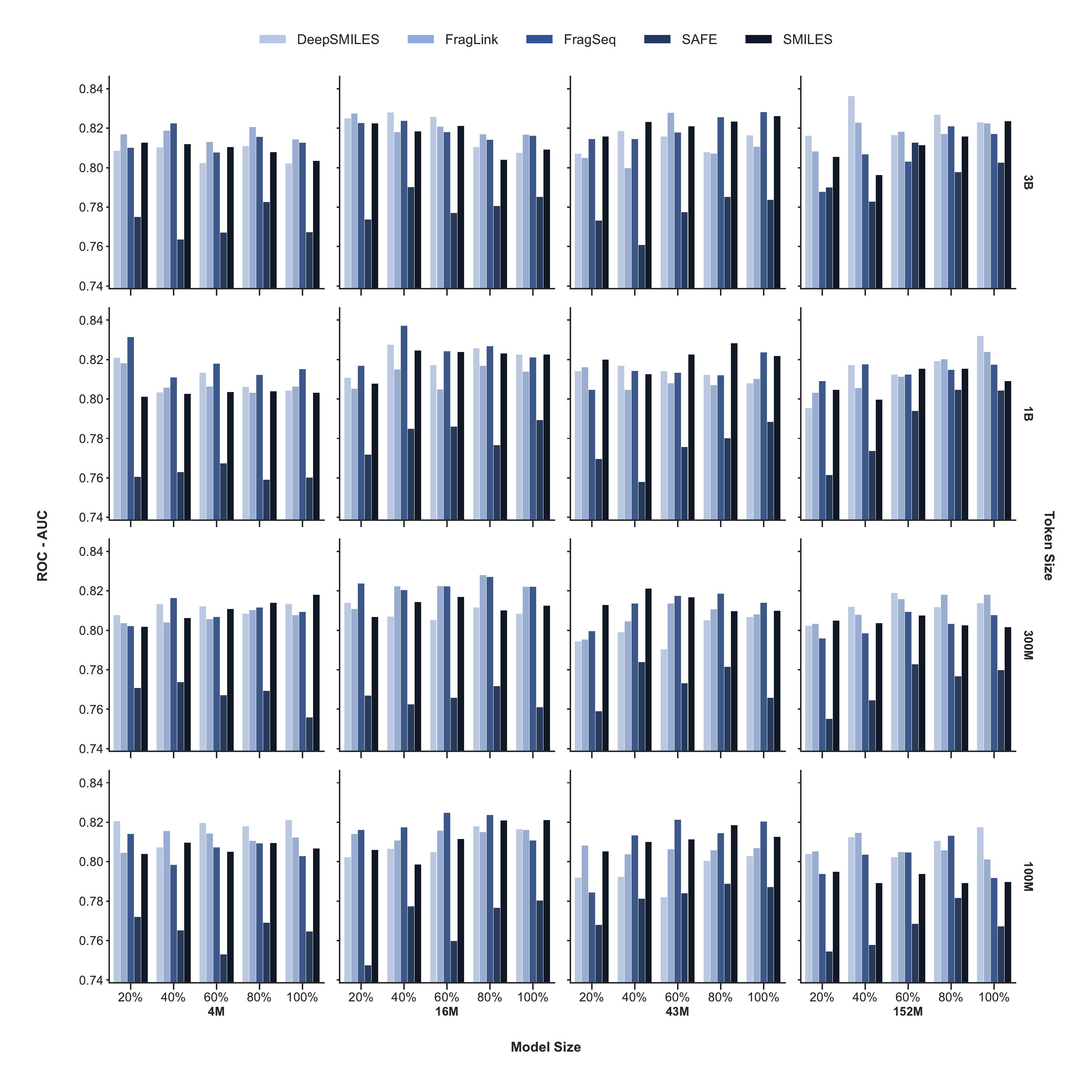}
    }
    \caption{Performance on Tox21 benchmark.}
    \label{figure: MoleculeNet Tox21 Bars}
  \end{center}
\end{figure*}

\begin{figure*}[]
  \vskip 0.2in
  \begin{center}
    \centerline{
        \includegraphics[width=0.98\linewidth]{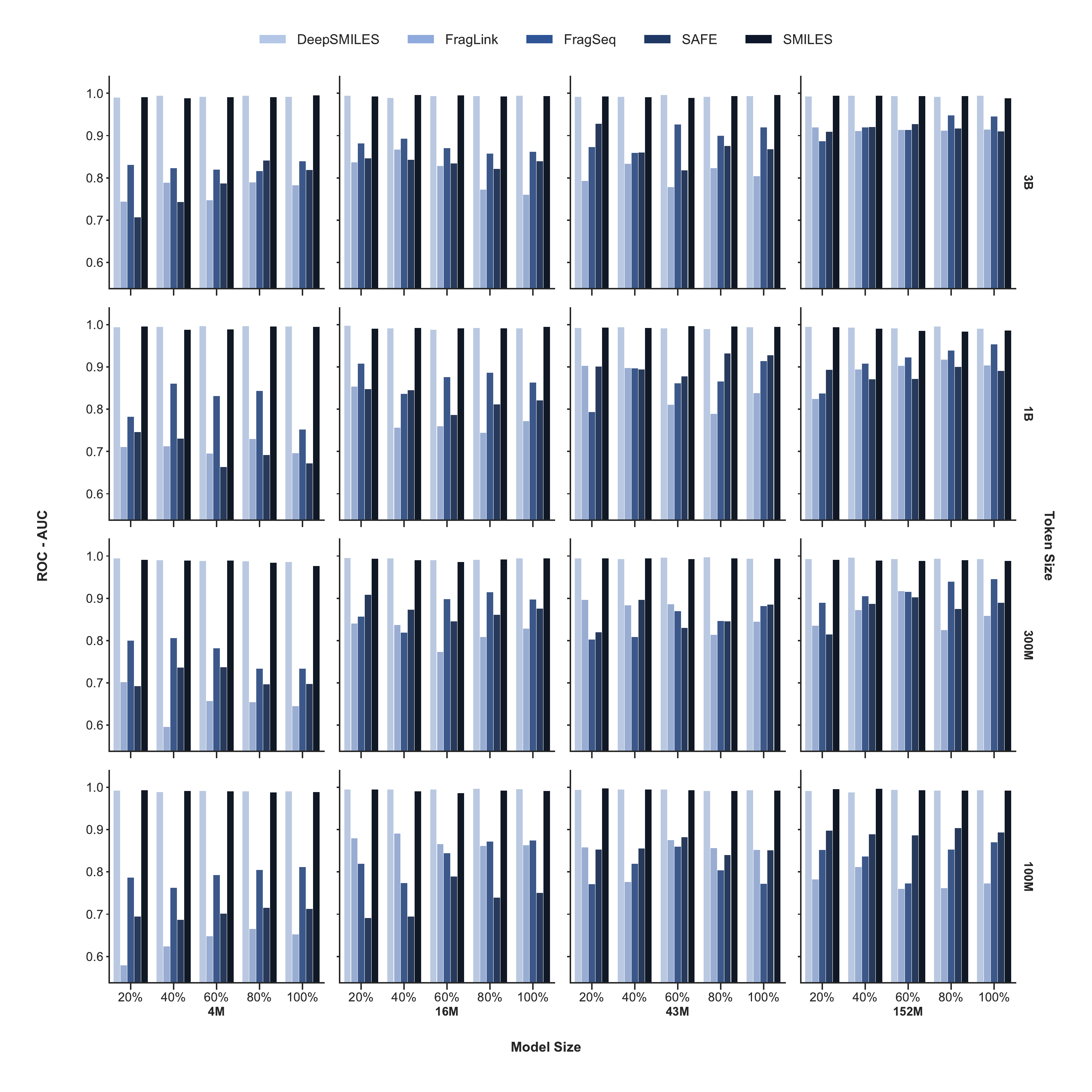}
    }
    \caption{Performance on ClinTox benchmark.}
    \label{figure: MoleculeNet ClinTox Bars}
  \end{center}
\end{figure*}

\begin{figure*}[]
  \vskip 0.2in
  \begin{center}
    \centerline{
        \includegraphics[width=0.98\linewidth]{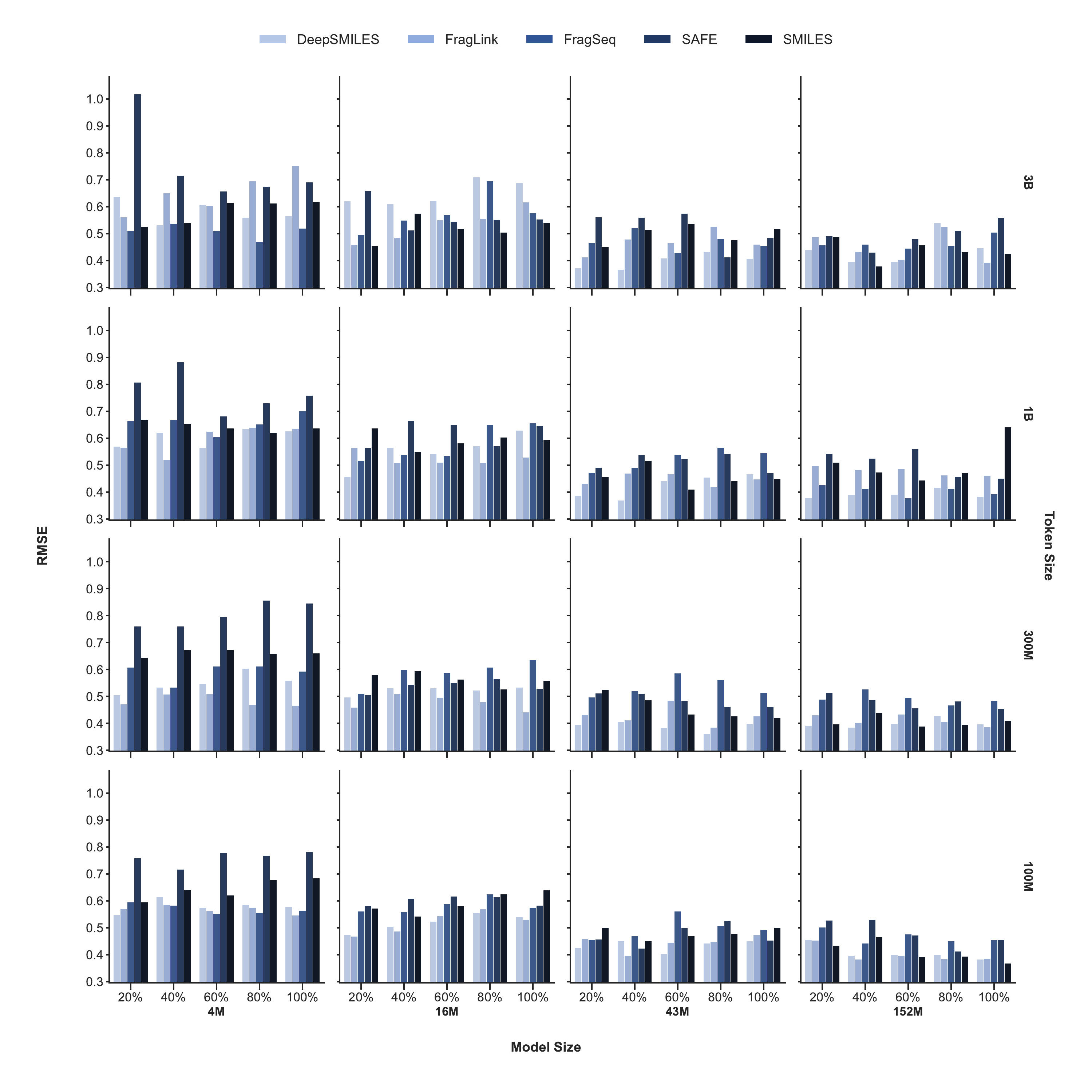}
    }
    \caption{Performance on ESOL benchmark.}
    \label{figure: MoleculeNet ESOL Bars}
  \end{center}
\end{figure*}

\begin{figure*}[]
  \vskip 0.2in
  \begin{center}
    \centerline{
        \includegraphics[width=0.98\linewidth]{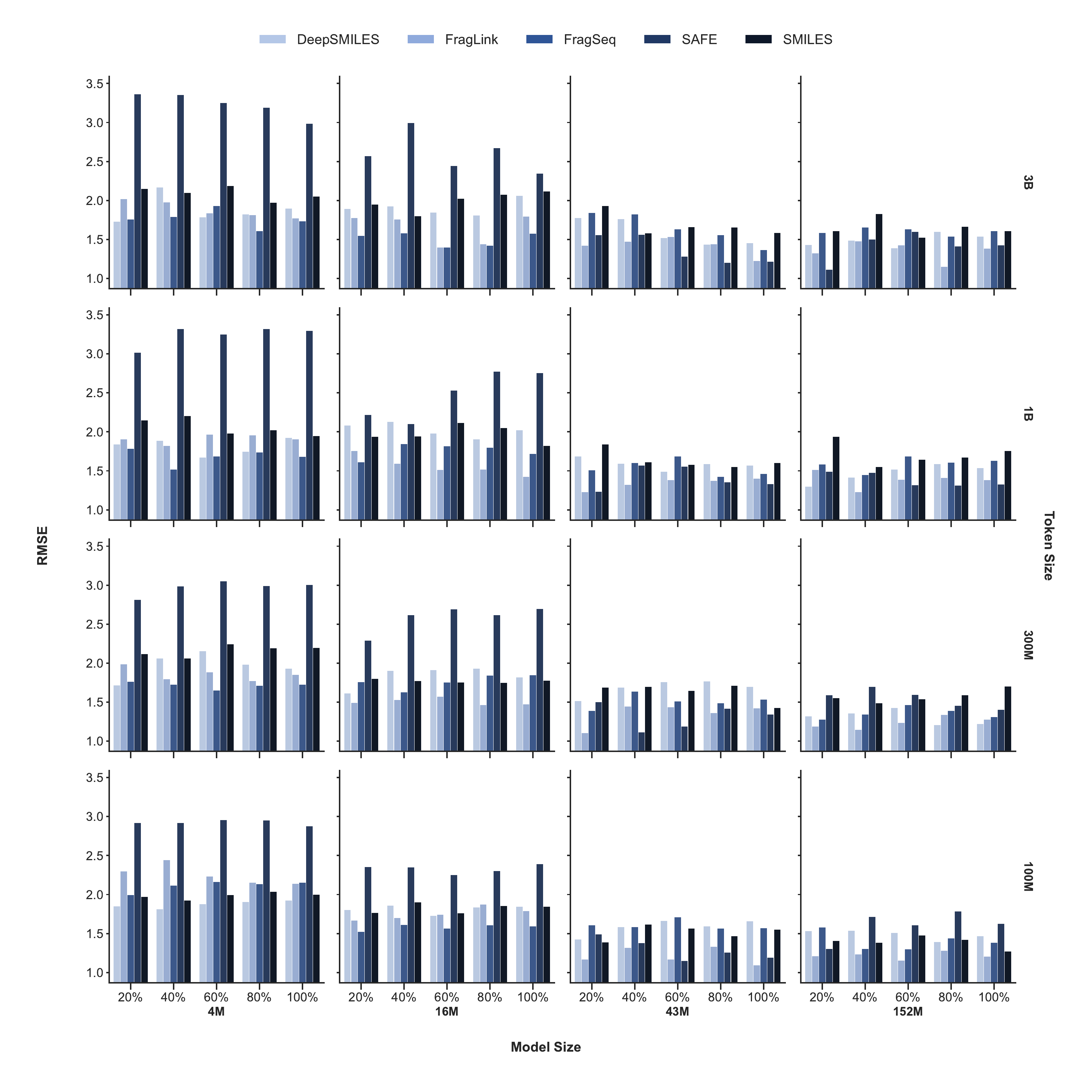}
    }
    \caption{Performance on FreeSolv benchmark.}
    \label{figure: MoleculeNet FreeSolv Bars}
  \end{center}
\end{figure*}

\begin{figure*}[]
  \vskip 0.2in
  \begin{center}
    \centerline{
        \includegraphics[width=0.98\linewidth]{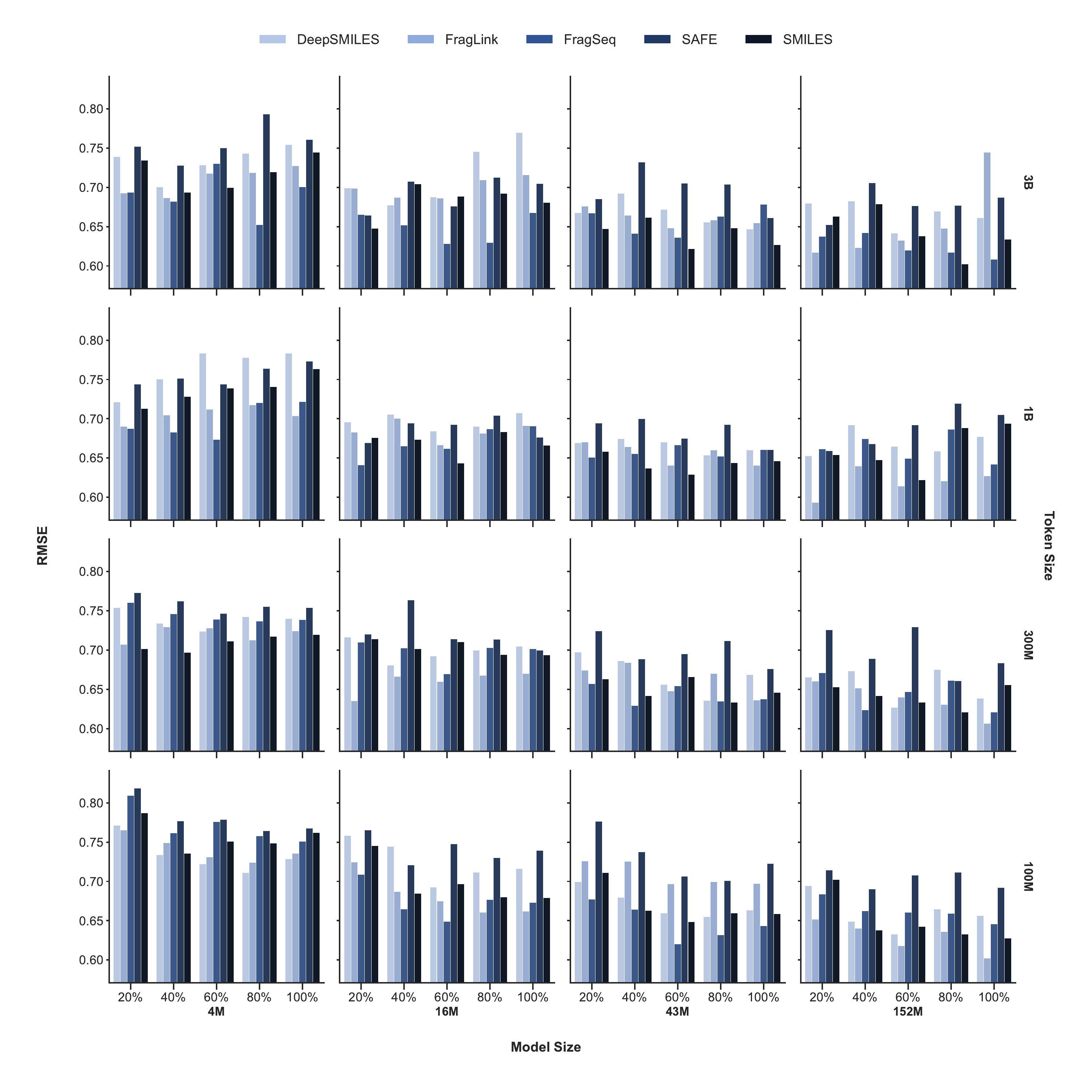}
    }
    \caption{Performance on Lipophilicity benchmark.}
    \label{figure: MoleculeNet Lipo Bars}
  \end{center}
\end{figure*}

Figures \ref{figure: MoleculeNet BACE Bars} through \ref{figure: MoleculeNet Lipo Bars} provide a granular view of model performance on each of the nine MoleculeNet benchmarks. Each figure is a grid of bar plots, where each subplot corresponds to a specific combination of model size (4M, 16M, 43M, 152M) and pre-training data volume (100M, 300M, 1B, 3B tokens). The y-axis represents the performance metric (ROC-AUC for classification, RMSE for regression), and each bar within a group corresponds to one of the five molecular representations. The x-axis within each subplot group shows the percentage of the total pre-training steps (20\% to 100\%), allowing us to analyze how performance evolves as pre-training progresses.

For BACE (Figure \ref{figure: MoleculeNet BACE Bars}), a consistent trend is visible where performance generally improves with more pre-training data (rows moving from bottom to top) and larger model sizes (columns moving from left to right). FragLink and FragSeq often show a strong advantage, particularly with more extensive pre-training. For HIV (Figure \ref{figure: MoleculeNet HIV Bars}), the performance gains are also evident with increased pre-training, but the differences between representations are more nuanced, with atom-level representations like SMILES and DeepSMILES performing competitively, especially at larger model scales.

On BBBP (Figure \ref{figure: MoleculeNet BBBP Bars}), DeepSMILES and SMILES show remarkable performance, benefiting significantly from larger pre-training datasets. This suggests that atom-level detail is critical for predicting membrane permeability. The SIDER (Figure \ref{figure: MoleculeNet SIDER Bars}) and Tox21 (Figure \ref{figure: MoleculeNet Tox21 Bars}) tasks exhibit more complex patterns. For these multi-task classification problems, performance does not always improve monotonically with model size or data, and certain representations like FragSeq show robustness in these challenging scenarios. For ClinTox (Figure \ref{figure: MoleculeNet ClinTox Bars}), most models achieve very high performance, indicating a potential ceiling effect, but the plots show that even a small amount of pre-training (20\%) on a large dataset (3B tokens) can lead to excellent results.

A clear and consistent trend is observed across ESOL (Figure \ref{figure: MoleculeNet ESOL Bars}), FreeSolv (Figure \ref{figure: MoleculeNet FreeSolv Bars}), and Lipophilicity (Figure \ref{figure: MoleculeNet Lipo Bars}): performance (lower error) consistently improves with both larger model sizes and greater pre-training data volume. The benefits of extensive pre-training are particularly pronounced here. For instance, models pre-trained on 3B tokens consistently outperform those trained on 100M tokens by a significant margin across all model sizes. While no single representation is universally dominant, FragLink often achieves or is competitive with the best-performing models, highlighting its strong capability for predicting these fundamental physicochemical properties.

\end{document}